\documentclass[dvipsnames]{bmvc2k}

\usepackage{xcolor}
\usepackage{graphicx}
\usepackage{comment}
\usepackage{amsmath,amssymb} 
\usepackage{mathtools}
\usepackage[capitalize, noabbrev]{cleveref}
\usepackage{multirow}
\usepackage{pifont}
\usepackage[labelfont={color=bmv@captioncolor,bf},font={color=bmv@captioncolor,small}]{caption}
\usepackage[export]{adjustbox}
\usepackage{wrapfig}
\usepackage[linesnumbered, commentsnumbered, ruled, vlined]{algorithm2e}

\newcommand{\vectorname}[1]{{\mathrm{\mathbf{#1}}}}
\newcommand{\myparagraph}[1]{\vspace{4pt}\noindent{\bf \color{bmv@sectioncolor} #1}}
\newcommand{\etal}{\textit{et al.}}
\newcommand{\ie}{\textit{i.e.}}
\graphicspath{{figures/}}
\DeclareGraphicsExtensions{.pdf,.png}

\Crefname{algocf}{Algorithm}{Algorithms}

\title{Cross-Modal Fusion Distillation for Fine-Grained Sketch-Based Image Retrieval}

\addauthor{Abhra Chaudhuri}{ac1151@exeter.ac.uk}{1}
\addauthor{$\text{Massimiliano Mancini}^\text{2}\\
\text{Yanbei Chen}^\text{2}\\
\text{Zeynep Akata}^\text{2,3,4}\\
\text{Anjan Dutta}$}{}{5}

\addinstitution{University of Exeter, UK}
\addinstitution{University of T\"{u}bingen, Germany}
\addinstitution{MPI for Informatics, Germany}
\addinstitution{MPI for Intelligent Systems, Germany}
\addinstitution{University of Surrey, UK}

\runninghead{Chaudhuri \etal}{Cross-Modal Fusion Distillation for FG-SBIR}

\def\etal{\emph{et al}\bmvaOneDot}

\begin{document}

\maketitle

\begin{abstract}
Representation learning for 
sketch-based image retrieval has mostly been 
tackled by 
learning embeddings that discard modality-specific information. As instances from different modalities can often provide complementary information describing the underlying concept, 
we 
propose a cross-attention framework 
for Vision Transformers (XModalViT) that fuses modality-specific information instead of discarding them. Our framework 
first maps paired datapoints from the individual photo and sketch modalities to fused representations that unify information from both modalities. We then decouple the input space of the aforementioned modality fusion network into independent encoders of the individual modalities via contrastive and relational cross-modal knowledge distillation. Such encoders can then be applied to downstream tasks like cross-modal retrieval. We demonstrate the expressive capacity of the learned representations 
by performing a wide range of experiments and achieving state-of-the-art results on three fine-grained sketch-based image retrieval benchmarks: Shoe-V2, Chair-V2 and Sketchy.
Implementation is available at \url{https://github.com/abhrac/xmodal-vit}.
\end{abstract}


\section{Introduction}
\label{sec:intro}
\begin{wrapfigure}[15]{r}{0.47\textwidth}
\vspace{-5pt}
        \includegraphics[width=0.47\textwidth]{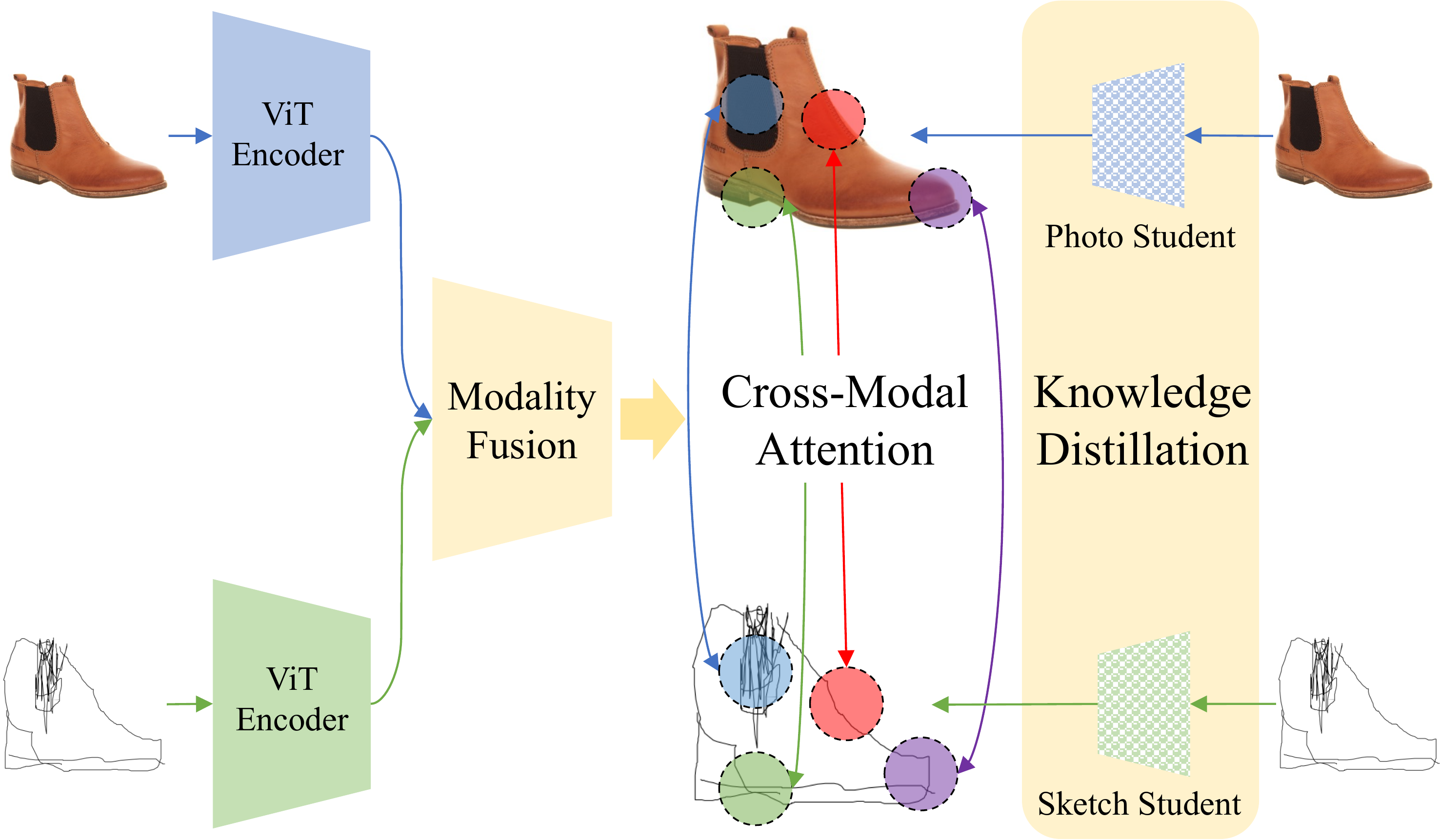}
        \caption{Our XModalViT retains semantically relevant, modality-specific features by learning a fused representation space, while bypassing the expensive cross-attention computation at runtime via cross-modal knowledge distillation.}
    \label{fig:model_teaser}
\end{wrapfigure}

Fine-grained sketch-based image retrieval (FG-SBIR) \cite{Bhunia2020SketchLF,Bhunia_2021_CVPR_MPaA} is a particular setting of SBIR \cite{Radenovic2018DSM,Dey2019Doodle,Dutta2019CVPR,Dutta2020SEMPCYC,PangKaiyue2020SMJP,Bhunia2020SketchLF,Bhunia_2021_CVPR_MPaA} that aims to retrieve a \emph{specific} photo based on a query sketch. Classical metric-learning based literature on FG-SBIR directly employs a contrastive learning strategy to estimate the modality-invariant component in a sketch-photo pair, optimizing an objective that aligns embeddings of similar photo-sketch pairs closer to and dissimilar pairs away from each other \cite{Yu2016SketchMe,Sangkloy2016Sketchy,FgsbirSpatialAttention,PangKaiyue2020SMJP}.
While such an approach only takes into account the shared mutual semantics between the two modalities, it does not consider how information specific to a modality might manifest itself upon being translated to the other modality. 
This is particularly important for free-hand sketches where the modality-gap is large due to imprecise 
depiction of object attributes. For example, as depicted in \cref{fig:model_teaser}, a region with a dark shade, may be distinguished by the sketcher, from one with a lighter shade, using a sparse collection of zigzag lines spanning the dark region (highlighted with a blue circle). 
Such nuances get eliminated during the modality filtering stage, but should ideally be preserved in an optimal fine-grained representation.

%
To model the large sketch-photo modality gap, we start with viewing sketches and photos as instantiations of an abstract conceptual representation derived from the interrelationships between local, spatial regions of the underlying object and their higher order interactions. The goal then is to derive the aforementioned representation by modeling the interrelationships and higher order interactions between the different localities across the two modalities.
%
%
Hence, we design our learning objective so as to unify the instance-discriminative modality-specific information into the encoded representations rather than discard them. By accounting for such variations across modalities, a representation would correspond to the complete abstract higher-order concept of which photos and sketches are different manifestations.
%
%
%
%
As elaborated above through the example in \Cref{fig:model_teaser}, in the cross-modal setting, a sketch-photo pair has the following three kinds of features 
-- (1) Modality-shared (useful for retrieval), (2) Modality-specific, that represents a \emph{shared underlying concept}, but manifests differently in the two modalities (useful for retrieval), (3) Modality-specific, that does not represent a shared concept (not useful for retrieval). Different from existing methods, we simultaneously preserve (1) and (2) while discarding (3) via a Modality Fusion Network and thus, are able to substitute modality alignment with a \emph{modality-fused instance alignment} by minimizing a cross-modal contrastive loss.
We decouple the fused space into independent, modality-specific encoders via a novel cross-modal knowledge distillation strategy, thereby avoiding the need for performing the expensive cross-modal fusion operation at runtime.

We encapsulate all of the above steps by designing the XModalViT framework centered on our novel cross-modal attention operation for Vision Transformers. In this paper, we make the following contributions: (1) A novel approach to the cross-modal visual representation learning task for FG-SBIR by designing a modality fusion operator for ViT based on the cross-attention mechanism, which unifies complementary information across modalities while being instance-discriminative at the same time.
(2) A cross-modal distillation technique to train independent encoders that can leverage a modality-fused representation space, without having to perform a computationally expensive cross-attention.
(3) State-of-the-art results from a wide range of experiments conducted on three benchmark datasets for the task of fine-grained SBIR, which further strengthens our claim in favor of preserving instance-discriminative, modality-specific information in the learned representations. 


\section{Related Work}
\label{sec:related}
Below we review the literature on FG-SBIR, Cross-Attention and Knowledge Distillation. 

%
\myparagraph{Fine-Grained Sketch-Based Image Retrieval:}
Early works on (FG-)SBIR \cite{Wei2021FGIA} were mainly focused on hand-crafted features, such as gradient field HOG \cite{Hu2013GFHOG}, deformable parts model \cite{DeformableParts}, histogram of edge local orientations \cite{Saavedra2014SHELO}, learned key shapes \cite{Saavedra2015LKS}, which were limited by the domain gap between sketches and photos. To address this issue, deep FG-SBIR models employing a deep triplet network to learn a common embedding space were proposed in \cite{Yu2016SketchMe,Sangkloy2016Sketchy},
an idea that was extended by the introduction of spatial attention \cite{FgsbirSpatialAttention}, self-supervised pre-training tasks \cite{PangKaiyue2020SMJP, bhunia2021sketch2vec}, or mining fine-grained local-features \cite{Xu2021DLANet}.
%
Generative models have also shown promising results \cite{Sain_2021_CVPR_StyleMeUp,BMVC2017_46,Bhunia_2021_CVPR_MPaA}, employing ideas like style-semantic disentanglement, cross-domain image synthesis, and reinforcement learning.
%
To the best of our knowledge, we are the first to study the effects of fusing information across modalities on 
fine-grained representations for FG-SBIR.

\myparagraph{Cross-Attention:} Dosovitskiy \etal \cite{dosovitskiy2020vit} introduced the idea of visual self-attention to be an effective strategy to learn local and global representations of an image.
Cross-attention extends this idea to incorporate multiple embedding spaces, which can arise from different modalities \cite{KrishnaD2020MultimodalER,Miech2021CVPR}, and can be used to perform tasks like capturing relationships between sentence words and image regions \cite{Wei_2020_CVPR}, unified, modality-agnostic classification \cite{Mohla_2020_CVPR_Workshops}, or multi-scale feature learning \cite{chen2021crossvit}.
%
%
%
In contrast, we design a cross-modal attention mechanism that facilitates the retention of semantically relevant, modality-specific features.
%

\begin{figure*}[t]
\centering
\includegraphics[width=\textwidth]{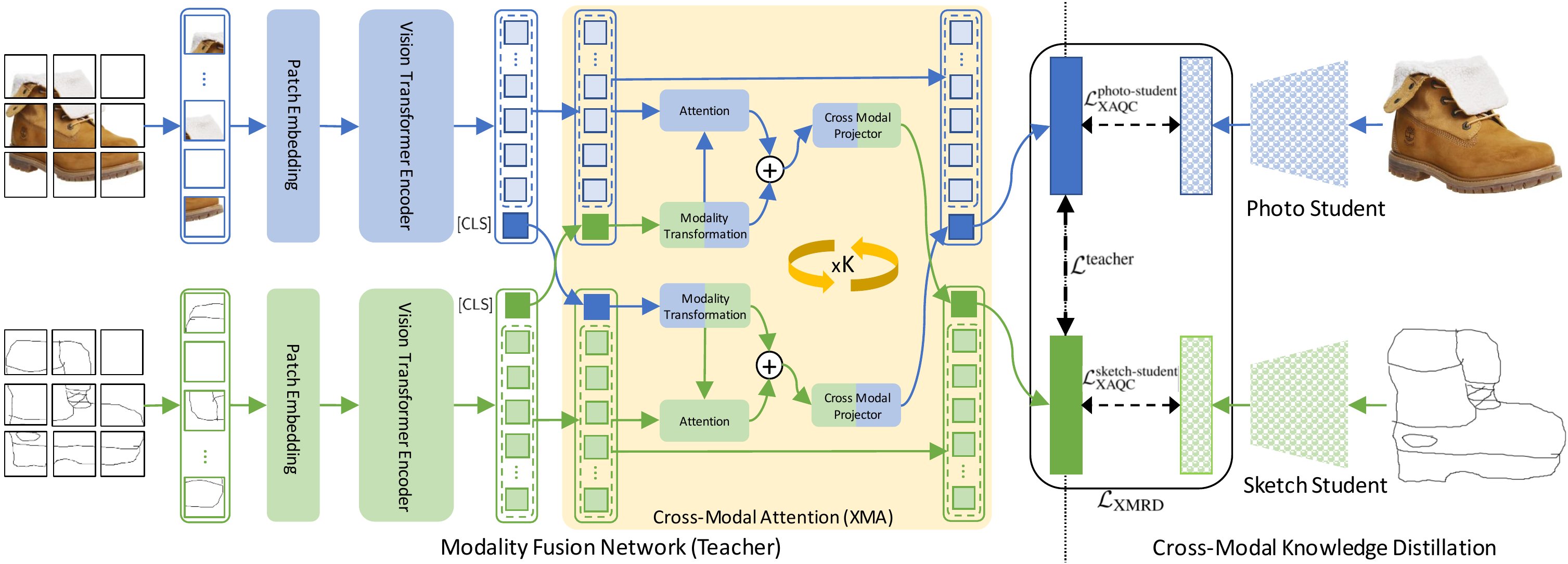}
\caption{Our XModalViT framework: The modality fusion network (teacher) takes as input, a positive photo-sketch pair and computes XMA embeddings for the individual modalities. During the cross-modal knowledge distillation phase, the weights of the teacher are frozen and the independent student networks are tasked with mapping the same pair of datapoints to the corresponding XMA branch of the teacher, thereby decoupling the domain of the XMA operator.}
\label{fig:model_arch}
\end{figure*}
\myparagraph{Knowledge Distillation:}
Knowledge Distillation \cite{hinton2015distilling} is the technique for transferring representations learned by one model (\emph{teacher}) to another (\emph{student}).
Originally formulated as a KL-Divergence minimization problem between the teacher-student outputs \cite{hinton2015distilling},
making the student equal to, or greater in size than the teacher \cite{Xie_2020_CVPR}, or maintaining geometric and relational invariants between the representation spaces of the two networks \cite{Park2019RelationalKD, Roth2021S2SD, Yang2022CVPR} provide improved approximation performance.
%
%
The idea of knowledge distillation was adapted in domains like image-to-video person ReID \cite{Gu2019TemporalKP,porrello2020RobustReIDKD}, as well as for retrieval tasks by minimizing various distance metrics between the teacher and the student embeddings \cite{Yu_2019_CVPR}, or by formulating the problem in contrastive learning terms \cite{Tian2020Contrastive, Oord2018RepresentationLW,Wu2018CVPR,He2020MomentumCF}.
%
Our work presents a novel and significantly more challenging application domain for knowledge distillation, with the requirement of transferring fused cross-modal representations to independent, modality specific encoders.

\section{The XModalViT Framework 
}

Let $\mathcal{P}$ and $\mathcal{S}$ denote the sets of photos and sketches respectively. Consider a function $\mathcal{P} \xrightarrow{} \mathcal{S}$, that takes a photo as input and returns the set of corresponding sketches, defined as
$p_i \in \mathcal{P} \longmapsto \{s_{ij} \in \mathcal{S} \mid p_i \xleftrightarrow{} s_{ij}\}$,
%
%
where $\xleftrightarrow{}$ denotes the correspondence relation between a photo-sketch pair. The FG-SBIR task is then to derive the function 
%
    $g(s) = p_i, \; \forall s \in \{s_{ij}\}$
%
This can be achieved by projecting the photo-sketch pairs into a common dot-product space $\mathbb{X}$, \ie, a cross-modal representation space, such that
$\forall p, p' \in \mathcal{P}$, where $p \xleftrightarrow{} s$ and $p \ne p'$, we have, $\xi_\text{photo}(p) \cdot \xi_\text{sketch}(s) > \xi_\text{photo}(p') \cdot \xi_\text{sketch}(s)$.
The encoders $\xi_\text{photo}$ and $\xi_\text{sketch}$ represent the respective photo and sketch projection functions. 
Thus, given a query sketch $s_q$, the retrieval result would be:
$\operatorname{arg}\displaystyle\max_{p \in \mathcal{P}} \xi_\text{photo}(p) \cdot \xi_\text{sketch}(s_q)$,
%
%
\ie, the photo $p \in \mathcal{P}$ that has the highest similarity with $s_q$ in $\mathbb{X}$.


We organize the cross-modal representation learning into a 2-step process, which we term XModalViT. First, we train a Vision Transformer based modality fusion network as a \emph{teacher} by taking photo-sketch pairs as inputs, to learn unified representations by fusing information from both the modalities. 
Next, we decouple the input space of this teacher network into independent, modality-specific encoders (\emph{students}) via knowledge distillation \cite{hinton2015distilling}. Our end-to-end framework is graphically depicted in \Cref{fig:model_arch}. Pseudocodes for training our models are provided in the supplementary material.

\subsection{Modality Fusion Network}
\label{sec:xmodalvit}
With the objective of fusing instance-discriminative features from multiple modalities, we propose a Vision Transformer based modality fusion network.
We obtain patch embeddings for an input image by dividing it into patches of size $16 \times 16$ and propagating their flattened versions through a linear layer.
%
A learnable vector, that we call the instance token, of the same size is prepended to the patch embeddings; this serves as the output representation of the ViT.
%
Abstractly, the teacher network can be defined as a binary function $\Gamma(p, s)$, that takes a photo $p$, and a sketch $s$ (of the same instance), as inputs and returns two vectors $\vectorname{x}_p$ and $\vectorname{x}_s$, the sketch-to-photo and photo-to-sketch cross-attention representations, respectively.

We now proceed to a more concrete definition of the sketch-to-photo cross-attention fusion operation, and the photo-to-sketch version will be analogous and symmetric. Let the instance tokens of the photo and the sketch branches be represented by $\mathrm{\mathbf{p}}_c$ and $\mathrm{\mathbf{s}}_c$ respectively, and the set of patch tokens for the sketch branch by $\mathrm{\mathbf{s}}_{patch}$.
We first propagate $\mathrm{\mathbf{p}}_c$ through a modality transformation layer $t_{\mathcal{P}\xrightarrow{}\mathcal{S}}$ that maps it from the photo-modality to the sketch-modality, producing $\mathrm{\mathbf{\overline{p}}}_c = t_{\mathcal{P}\xrightarrow{}\mathcal{S}}\left( \mathrm{\mathbf{p}}_c \right)$. We then concatenate $\mathrm{\mathbf{\overline{p}}}_c$ to $\mathrm{\mathbf{s}}_\text{patch}$ to obtain $\mathrm{\mathbf{\overline{s}}} = [\mathrm{\mathbf{\overline{p}}}_c \,,\, \mathrm{\mathbf{s}}_\text{patch}]$.
Learnable matrices $\textbf{W}_q$ and $\textbf{W}_k$ are used to project $\mathrm{\mathbf{\overline{p}}}_c$ and $\mathrm{\mathbf{\overline{s}}}$ respectively, onto the same dot-product space with $D$ dimensions where the attention scores for $\mathrm{\mathbf{\overline{p}}}_c$ are computed, followed by a subsequent \emph{softmax} on their product. The cross-modal-attention embedding for the sketch-to-photo branch can then be computed as:
%
%
%
\begin{align*}
    \mathrm{\mathbf{a}} = \sigma\left( (\mathrm{\mathbf{\overline{p}}}_c\textbf{W}_q) \boldsymbol{\cdot} (\mathrm{\mathbf{\bar{s}}}\textbf{W}_k)^\text{T}) / \sqrt{D} \right), \;
    \mathrm{\mathbf{x}}_p = \Phi^{\mathcal{S}\xrightarrow{}\mathcal{P}}_\mathbb{X} \left( \mathrm{\mathbf{\overline{p}}}_c + l_{norm}(\mathrm{\mathbf{a}} \boldsymbol{\cdot} \mathrm{\mathbf{\bar{s}}}\textbf{W}_k) \right)
\end{align*}
%
%
where $\Phi^{*}_\mathbb{X}$ is a fully-connected projection head, mapping the sketch-to-photo and photo-to-sketch embeddings to the same representation space $\mathbb{X}$, which we term as the cross-modal attention space, and $l_{norm}(\cdot)$ is the layer normalization operation \cite{ba2016layer}.
Since cross-modal attention follows the same operational semantics as that of self-attention, it can also be computed for multiple attention heads \cite{dosovitskiy2020vit,chen2021crossvit}. For $m$ heads, $m$ cross-modal attention operations are performed in parallel followed by a concatenation and projection of their outputs, and $D$ is set to $D/m$ to keep compute and number of parameters constant.

\myparagraph{Learning Objective:} We design an objective
termed Cross-Attention Queue Contrast (XAQC), to learn the cross-modal attention space $\mathbb{X}$. XAQC aims to bring cross-modal attention (XMA) representations $\mathrm{\mathbf{x}}_p$ (XMA-photo) and $\mathrm{\mathbf{x}}_s$ (XMA-sketch) for the positive sketch-photo pairs close together and push those of different instances away from each other. This alignment is achieved by making the XMA-sketch representations act as soft targets for the XMA-photo representations for the same instance. XMA-sketch representations for other instances are treated as negatives and a $(k+1)$-way softmax-based binary cross-entropy loss is minimized under this setting, where $k$ is the number of negatives.

Consider a photo $p$, and two of its corresponding sketches $s_1$ and $s_2$. The cross-attention representations of the sketch-photo pairs $(p, s_1)$ and $(p, s_2)$ are $(\vectorname{x}^1_p, \vectorname{x}^1_s)$ and $(\vectorname{x}^2_p, \vectorname{x}^2_s)$. The loss $\textbf{\texttt{XAQC}}(\vectorname{x}^1_p, \vectorname{x}^2_s, \mathcal{Q})$ is then formulated as:
%
\begin{align}
\label{teacherXAQC}
    \mathcal{L}^{\text{teacher}} = \textbf{\texttt{XAQC}}(\vectorname{x}^1_p, \vectorname{x}^2_s, \mathcal{Q}) = - \log\frac{
        \exp{(
        \vectorname{x}^1_p \boldsymbol{\cdot} \vectorname{x}^2_s} / 
        \tau)
    }{\sum\limits_{\forall \vectorname{h}_s \in \mathcal{X}} \exp{\left((\vectorname{x}^1_p \boldsymbol{\cdot} \vectorname{h}_s) / \tau \right)}}%
\end{align}
where $\mathcal{X} = \mathcal{Q} \cup \{\vectorname{x}^2_s\}$, $\mathcal{Q}$ is a fixed-size dynamic queue of XMA-sketch representations from previous mini-batches,
and $\tau$ is a hyperparameter controlling the concentration of the distribution
(higher values produce softer distributions).
For each new sample, we enqueue its photo-to-sketch representation $\vectorname{x}^2_s$ into $\mathcal{Q}$ after computing \cref{teacherXAQC}\footnote{In implementation, the queue update is done at the level of a mini-batch rather than a sample.}. We freeze the weights of the network while computing $x^2_s$. While training, we randomly sample the sketch pairs, so $\textbf{\texttt{XAQC}}(\vectorname{x}^2_p, \vectorname{x}^1_s, \mathcal{Q})$ would also be invoked at some point during the training. Since the dot-product is a symmetric similarity metric, the corresponding symmetric property of a sketch representation being closer to its photo representation rather than to the representations of other photos is also satisfied by minimizing the $\textbf{\texttt{XAQC}}$ loss.

\subsection{Cross-Modal Fusion Distillation}
\label{sec:kd}

To bypass the expensive cross-modal attention computation at test-time, we propose a strategy to decouple the joint photo-sketch input space of the modality fusion \emph{teacher} network, $\Gamma$, by training simple uni-modal CNNs or ViTs (\emph{students}) to align their output with that of the corresponding branches of the teacher. We formulate this process as a composition of contrastive and relational cross-modal distillation.

Since the uni-modal students only ever encounter information from a single modality, the modality-fused XMA representations obtained from the teacher
act as oracles, directing the optimizers of the uni-modal students to converge to a locality in the representation space that captures information from both the modalities, while being also instance-discriminative. 

\myparagraph{Contrastive Cross-Modal Distillation:}
We introduce an objective that treats cross-modal distillation as a contrastive learning problem,
aiming to pull teacher and student representations for the same input close together, while pushing those for different inputs farther apart. {More specifically, we leverage the contrastive nature of the proposed XAQC loss (Eq. \eqref{teacherXAQC}) 
and use it as a representation alignment criterion for training the students, as we detail below.}

Given a corresponding photo-sketch pair $(p_i, s_i)$, we formulate our contrastive cross-modal distillation objective so as to learn photo and sketch encoders $\xi_\text{photo}$ and $\xi_\text{sketch}$ that bring $\xi_\text{photo}(p_i)$ and $\xi_\text{sketch}(s_i)$ close to $\vectorname{x}_{p_i}$ and $\vectorname{x}_{s_i}$ (teacher embeddings obtained from $\Gamma$) respectively, and push apart the representations $\vectorname{x}_{p_j}$ and $\vectorname{x}_{s_j}$, where $i \ne j$. For this purpose, we {use the \textbf{\texttt{XAQC}} loss as}: 
%
\begin{align}
    \label{SketchStudentXAQC}
    \mathcal{L}^\text{sketch-student}_\text{XAQC} = 
    \textbf{\texttt{XAQC}} (\xi_\text{sketch}(s_i), \vectorname{x}_{s_i}, \mathcal{Q}_\mathcal{S})
\end{align}
where $\xi_\text{sketch}: \mathcal{S} \xrightarrow{} \mathbb{X}$ is an encoder mapping sketches to the cross-modal-attention space, $s_i$ is a sketch of the instance $i$, and $\vectorname{x}^i_s$, the XMA-sketch with its corresponding photo. $\mathcal{Q}_\mathcal{S}$ is a fixed size dynamic queue containing XMA-sketch representations of samples from previous mini-batches, and which is enqueued with $\vectorname{x}^i_s$ after computing \cref{SketchStudentXAQC}.
The learning objective for the photo student, \ie, $\mathcal{L}^\text{photo-student}_\text{XAQC}$, is also symmetrically formulated.

\myparagraph{Transfer of Cross-Modal Attention Geometry:} We also introduce the additional constraint of preserving certain semantically meaningful geometric properties of the teacher's cross-modal attention space, in the students' representation space.
Such constraints have been found to benefit knowledge distillation in the uni-modal scenario \cite{Park2019RelationalKD}.
To fulfill this goal, we model 
the relationships between the embeddings of arbitrary $k$-tuples of datapoints in terms of distance and angular relationships.

Consider photo-sketch pairs $(p_1, s_1), (p_2, s_2)$ and $(p_3, s_3)$.
%
Let the XMA and the student embeddings be given by $\vectorname{x}_{p_i}, \vectorname{x}_{s_i} = \Gamma(p_i, s_i)$ and $\vectorname{z}_{p_i} = \xi_\text{photo}(p_i),\;\vectorname{z}_{s_i} = \xi_\text{sketch}(s_i)$, respectively.
%
%
%
%
Let $m$ be an abstract notation for a modality, i.e., $m \in \{p, s\}$, generically standing for both the photo and the sketch modalities.
The computation of the mutual relational potentials among the teacher and the student embeddings can then be expressed as
$\pi^\text{teacher}_m = \psi_1(\vectorname{x}_{m_1}, \vectorname{x}_{m_2}) + \psi_2(\vectorname{x}_{m_1}, \vectorname{x}_{m_2}, \vectorname{x}_{m_3})$ and $\pi^\text{student}_m = \psi_1(\vectorname{z}_{m_1}, \vectorname{z}_{m_2}) + \psi_2(\vectorname{z}_{m_1}, \vectorname{z}_{m_2}, \vectorname{z}_{m_3})$, respectively.
%
%
%
$\psi_1$ and $\psi_2$ are distance and angle based relation potential functions respectively, defined as:
%
\vspace{-5pt}
\begin{align*}
    \psi_1(x, y) =
    \frac{1}{\mu}||x - y||_2; \; \psi_2(x, y, z) =
    \frac{x - y}{||x - y||_2} \cdot \frac{z - y}{||z - y||_2}
\end{align*}
\vspace{-2pt}
%
%
where $\mu$ is a normalization factor equal to the average distance among all $(x, y)$ pairs in a mini-batch. Finally, the {Cross-Modal Relational Distillation (XMRD)} loss between the teacher and the student is obtained as follows:
\begin{align*}
    \mathcal{L}_\text{XMRD} =
    \delta(\pi^\text{teacher}_p, \pi^\text{student}_p) + 
    \delta(\pi^\text{teacher}_s, \pi^\text{student}_s),
\end{align*}
where $\delta(\cdot)$ is the Huber loss, given by
$
    \delta(a, b) = 
    \begin{cases}
        \frac{1}{2}(a - b)^2 \qquad \text{for} \; |a-b| \leq 1 \\
        |a - b| - \frac{1}{2}, \quad \text{otherwise.}
    \end{cases}
$
%

\myparagraph{Learning Objective:} The final objective for training the photo and the sketch students after combining the contrastive and the cross-modal relational distillation losses is as follows:
\begin{align*}
    \mathcal{L}^\text{student} =
    \mathcal{L}^\text{sketch-student}_\text{XAQC} + \mathcal{L}^\text{photo-student}_\text{XAQC} + \lambda \cdot \mathcal{L}_\text{XMRD},
\end{align*}
where $\lambda$ is a hyperparameter for balancing the contrastive and the relational components.


\section{Experiments}
\label{sec:expt}
In this section, we perform an extensive experimental evaluation of our model on several FG-SBIR benchmarks and ablate our model components.

\myparagraph{Datasets:} 
We consider both single and multi-class standard FG-SBIR datasets for experimental evaluation, where the former contain photos and sketches of a large number of instances of a single class, and the latter contain a large number of classes, with relatively fewer instances per class. The single-class datasets comprise of the QMUL-Shoe-V2 and QMUL-Chair-V2 \cite{Yu2016SketchMe}, while for multi-class evaluation, we use the Sketchy database \cite{Sangkloy2016Sketchy}.

The total number of photos and corresponding sketches present in each of the datasets along with their fraction used for testing are listed in the supplementary. 
We follow the dataset splitting convention as well as the performance evaluation metric of recent works \cite{Song2018LearningTS,SainAneeshan2020CHMf,Bhunia_2021_CVPR_MPaA,Sain_2021_CVPR_StyleMeUp,Bhunia2020SketchLF} for the QMUL FG-SBIR datasets and \cite{Sangkloy2016Sketchy,BMVC2017_46} for Sketchy.
We use the $\text{acc}@K$ evaluation metric \cite{Su2015PerfEvalIR} (with $K = $ 1 and 10)
which computes the proportion of query sketches for which the correct photo was present in the top-$K$ returned results \cite{Bhunia_2021_CVPR_MPaA,Sain_2021_CVPR_StyleMeUp,Yang2021SketchAA}.


\myparagraph{Implementation Details:} 
%
We use ViT base (ViT-B) models pre-trained on ImageNet \cite{ImageNet21k} from \cite{rw2019timm} as homogeneous encoders of the XMA teacher. We use 12 layers of cross-modal attention, which produce an output embedding of 768 dimensions. For the students, we primarily report and suggest the usage of ViT small (ViT-S) networks as backbones, since they provide the most optimal size-to-accuracy ratio. However, depending on requirements, one may choose to use other ViT architectures or even CNNs for the purpose of knowledge distillation. 
The size of the XAQC queues were computed as $2^{\lfloor \log N \rfloor}$, where $N$ is the number of datapoints.
The teacher network was optimized with a learning rate of $3\times10^{-6}$ under a cosine-annealed schedule and a weight decay of $10^{-4}$ using the Adam optimizer \cite{Diederik2015Adam}. The student networks were optimized using an initial learning rate of $10^{-5}$ with an exponential decay and a weight decay of $10^{-5}$. 
The details of our hardware and software platforms are provided in the supplementary.

\subsection{Ablation Studies}

\myparagraph{Model Components and Losses:}
\cref{tab:ablation} reports our ablation studies, where XMA stands 
\begin{wraptable}[14]{r}{0.5\textwidth}
    \vspace{-5pt}
    \centering
    \resizebox{0.48\textwidth}{!}{
        \begin{tabular}{c|c|c|c|c|c|c}
        \multirow{2}{*}{ID} & \multicolumn{2}{c|}{Teacher} & \multicolumn{2}{c|}{Student} & Shoe-V2 & Chair-V2 \\
        & XMA & XAQC-R & XAQC-D & XMRD & acc@1 & acc@1\\
        \hline
        1 & -- & -- & -- & -- & 30.83 & 49.66 \\
        \hline
        \multirow{4}{*}{2} & \ding{51}  &  &  &  & 34.31 & 52.35 \\
        & \ding{51} & \ding{51} & & & 41.33 & 57.65 \\
        & \ding{51} & & \ding{51} & & 35.62 & 53.40 \\
        & \ding{51} & & & \ding{51} & 35.45 & 52.40 \\
        \hline
        \multirow{4}{*}{3} & \ding{51}  & \ding{51} & \ding{51} & & 43.21 & 62.70 \\
        & \ding{51} & \ding{51} & &\ding{51} & 42.73 & 60.55 \\
        & \ding{51} & & \ding{51} & \ding{51} & 35.70 & 54.98 \\
        & & \ding{51} & -- & -- & 36.50 & 55.22 \\
        \hline
        \textbf{4} & \ding{51} & \ding{51} & \ding{51} & \ding{51} & \textbf{45.05} & \textbf{63.48}
        \end{tabular}}
    \caption{Results of ablating the core components of our model; grouped so as to provide a view of how each of the components contribute individually and in conjunction with each other. 
    `--' indicates that the component is irrelevant for that particular setting.}
\label{tab:ablation}
\end{wraptable}
for cross-modal attention, and its presence means that the XMA operator is applied to the outputs of the homogeneous encoders. Otherwise, the outputs of the homogeneous encoders serve as the final representation. XAQC-R and XAQC-D refer to the usage of the XAQC loss for ranking the teacher's representation space and as a contrastive knowledge distillation criterion for the students respectively. Upon ablation, the triplet loss serves as the alternative in both cases.
XMRD indicates the presence or absence of the cross-modal relational distillation criterion while training the students.

We group our results into 4 categories (indicated by ID). ID-1 establishes a baseline performance by following a classic deep learning based SBIR framework \cite{Sangkloy2016Sketchy} with ViTs as encoders for the individual modalities, with the triplet loss as the only optimization objective. The improvements over this model could be used to demonstrate the algorithmic contributions of our work. ID-2 illustrates the contribution of the cross-modal attention operation, and how each of the other components of the network enhances the overall performance by appropriately utilizing the representation learned by it.
ID-3 provides an estimate of how individually suppressing each of the components of the framework affects the overall performance. ID-4 reports the performance of the complete model.
%
%
%
\begin{figure}
    \centering
    \resizebox{\textwidth}{!}{
    \begin{tabular}{cc}
        \includegraphics[width=0.5\textwidth]{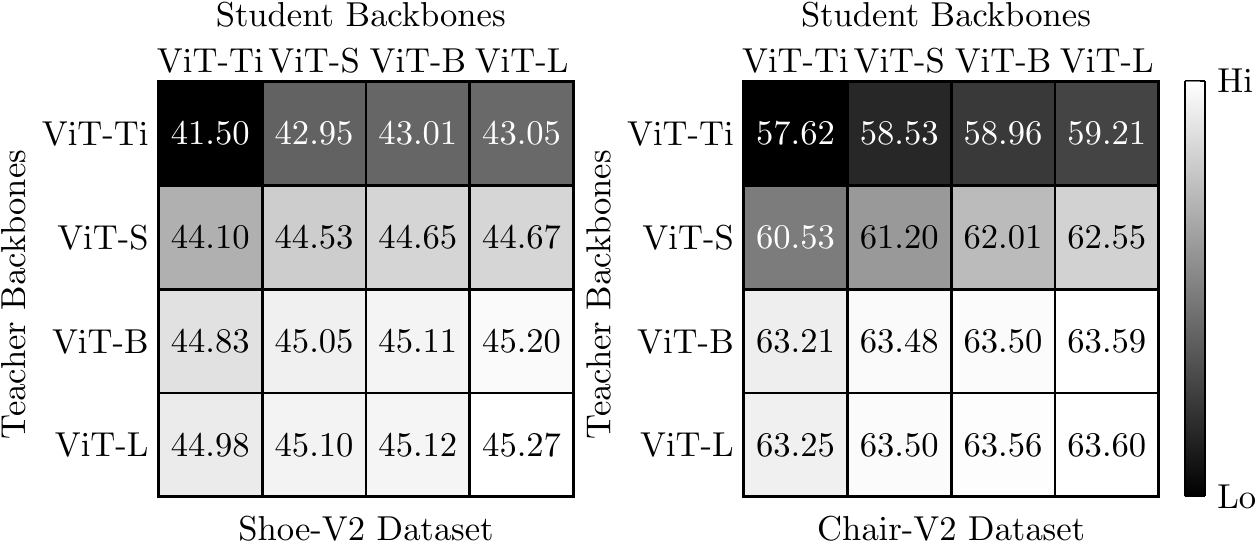} & \includegraphics[width=0.45\textwidth]{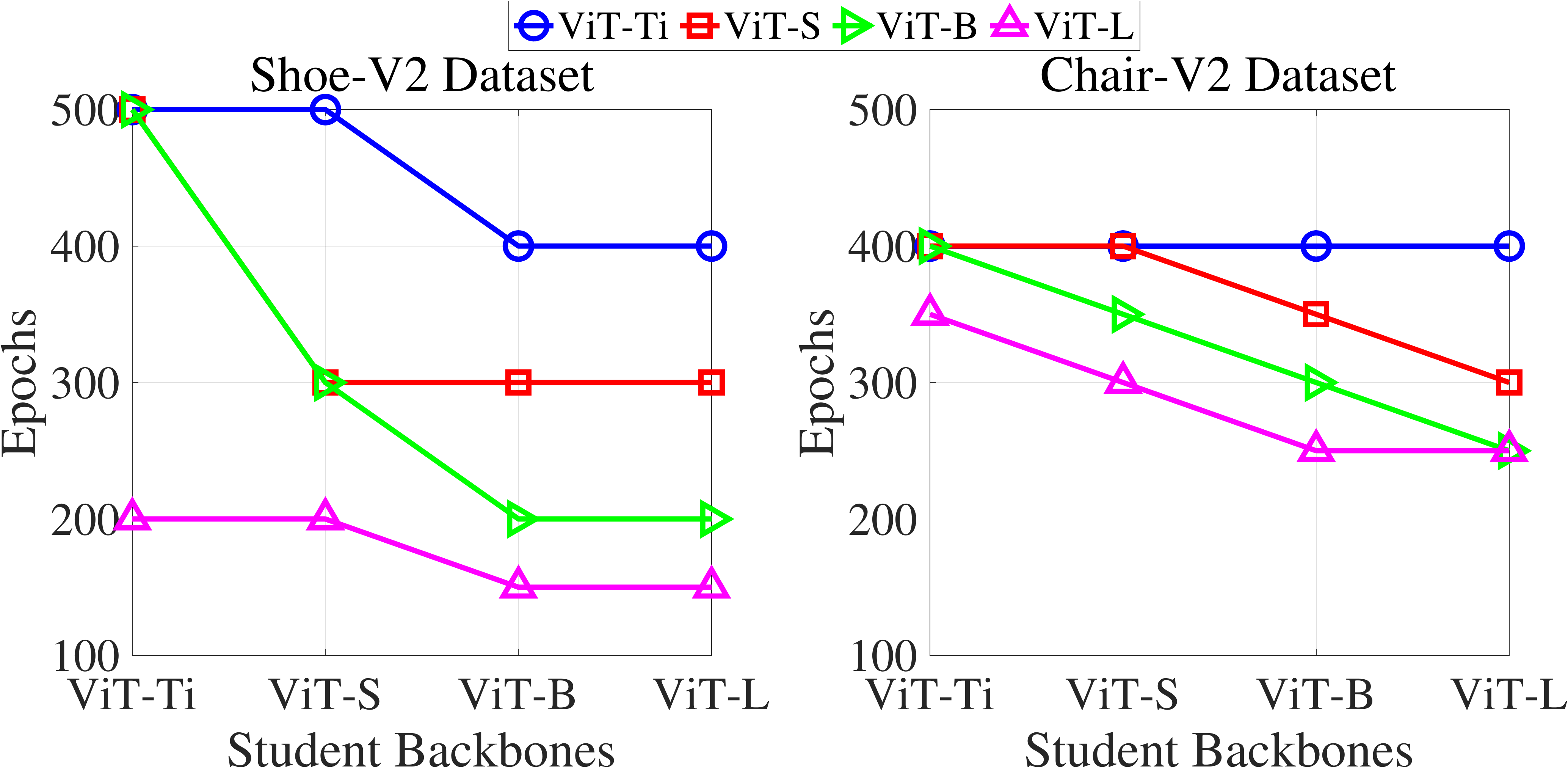}
    \end{tabular}}
    \caption{(Left) Variation in acc@1 and (right) student convergence time (in number of epochs) with teacher/student encoder size.}
    \label{fig:ablation_backbones}
\end{figure}
The first row in ID-2 indicates that the XMA in isolation provides over 2.6\% gain in acc@1. The results also indicate that the contrastive loss used for ranking the teacher's representation space is of pivotal importance. Using the XAQC loss to train the teacher instead of triplet can improve the performance by up to 7\%. Thus, it is only when the teacher learns robust enough representations that the students are able to distill out the relevant knowledge.
The results in ID-3 also echo the findings of ID-2; suppressing the cross-modal attention and the XAQC loss in the teacher have the most degrading effects. The last row in ID-3 is equivalent to the setup in ID-1 trained with the XAQC-R loss. 
%
With no modality fusion, the encoders are already independent, and hence, do not require the distillation step.

\myparagraph{Encoder Backbones}: We considered the Tiny (Ti), Small (S), Base (B) and Large (L) versions of the ViT and used them as backbones in both the teacher and the students. \Cref{fig:ablation_backbones} summarizes the effects of pairing backbones of different sizes. 
We observe that, with the Tiny teacher, the performance significantly improves as we increase the size of the student. However, this trend 
stabilizes as we increase the size of the teacher. However, with the student fixed, as we increase the size of the teacher, the convergence times 
reduce.
This drop in convergence time continues as we increase the size of the student.
Based on the above observations, we postulate that, larger teachers, by the virtue of their higher entropic capacity and output dimensionality, are able to learn more expressive representations,
thereby presenting a simpler target hypothesis for the downstream students to approximate. Smaller teachers are forced to obtain a more compressed representation, thereby limiting their amount of expressivity. This results in a more complex target hypothesis and thus, 
can be better learned by larger students (searching through a larger hypothesis space).
We also experimented with CNNs as student backbones and achieved similar results (details in the supplementary).

\subsection{Comparison with State-of-the-Art}
\myparagraph{QMUL FG-SBIR:}
The quantitative comparison of our method with the QMUL FG-SBIR datasets (Shoe-V2 and Chair-V2) is given in \Cref{tab:SOTAComparisonQMUL}.
\emph{Triplet-SN} \cite{Yu2016SketchMe} uses triplet loss to train a Sketch-a-Net \cite{Yu2017SketchANet} baseline. \emph{Triplet-Attn} \cite{FgsbirSpatialAttention} is a spatial attention based extension of \cite{Yu2016SketchMe}. \emph{Triplet-RL} performs on-the-fly FG-SBIR by a reinforcement learning based fine-tuning. \emph{CC-Gen} \cite{Pang2019CVPR} models a universal manifold of prototypical cross-category sketch traits. \emph{TVAE} \cite{Ishfaq2018tvae} employs a VAE with single modality translation, while \emph{DVML} \cite{Lin2018ECCV} disentangles sketch features into variant and invariant components. Strong performance on Shoe-V2 was achieved
\begin{wraptable}[14]{r}{0.5\textwidth}
\vspace{-8pt}
\resizebox{0.5\textwidth}{!}{
\begin{tabular}{l|c|c|c|c}
\multirow{2}{*}{Method} & \multicolumn{2}{c|}{Shoe-V2} & \multicolumn{2}{c}{Chair-V2} \\
& acc@1 & acc@10 & acc@1 & acc@10 \\
\hline
Triplet-SN \cite{Yu2016SketchMe} & 28.71 & 71.56 & 47.65 & 84.24 \\
Triplet-Attn \cite{FgsbirSpatialAttention} & 31.74 & 75.78 & 53.41 & 87.56 \\
Triplet-RL \cite{Bhunia2020SketchLF} & 34.10 & 78.82 & 56.54 & 89.61 \\
CC-Gen \cite{Pang2019CVPR} & 33.80 & 77.86 & 54.21 & 88.23 \\
\hline
TVAE \cite{Ishfaq2018tvae} & 27.62 & 70.32 & 49.37 & 81.63 \\
DVML \cite{Lin2018ECCV} & 32.07 & 76.23 & 52.78 & 85.24 \\
\hline
SketchAA \cite{Yang2021SketchAA} & 32.33 & 79.63 & 52.89 & 94.88 \\
StyleVAE \cite{Sain_2021_CVPR_StyleMeUp} & 36.47 & 81.83 & 62.86 & 91.14 \\
ReinfGen \cite{Bhunia_2021_CVPR_MPaA} & 39.10 & 87.50 & 62.20 & 90.80 \\
Partial-SBIR \cite{Chowdhury2022PartialSBIR} & 39.90 & 82.90 & - & - \\
NT-SBIR \cite{Bhunia2022NTSBIR} & 43.70 & - & \textbf{64.80} & - \\
\hline
\textbf{Ours (XModalViT)} & \textbf{45.05} & \textbf{90.23} & 63.48 & \textbf{95.02}
\end{tabular}}
\caption{Quantitative comparison (in \%) with state-of-the-art for fine-grained SBIR on the QMUL datasets.}
\label{tab:SOTAComparisonQMUL}
\end{wraptable}
by \emph{ReinfGen} \cite{Bhunia_2021_CVPR_MPaA} via a sketch generative framework based on reinforcement learning using additional unpaired training photos. 
Promising results were reported by StyleVAE \cite{Sain_2021_CVPR_StyleMeUp} and \emph{SketchAA} \cite{Yang2021SketchAA} via sketch disentanglement into independent style and content components, and sketch abstraction through a graph convolutional network respectively.
\emph{Partial-SBIR} \cite{Chowdhury2022PartialSBIR} and \emph{NT-SBIR} \cite{Bhunia2022NTSBIR} respectively proposed dealing with partial information and noise in sketches via optimal transport and reinforcement learning based approaches. \emph{NT-SBIR} is currently the SOTA on acc@1 for Chair-V2.
%
With our novel XModalViT framework, we were able to beat the acc@1 and acc@10 SOTA on Shoe-V2 and acc@10 SOTA on Chair-V2 by 1.35\%, 2.73\%, and 0.14\% respectively.

\myparagraph{Sketchy:} The quantitative comparison of our method on the Sketchy dataset is given in
\begin{wraptable}[12]{r}{0.4\textwidth}
\vspace{-6pt}
\resizebox{0.4\textwidth}{!}{
\begin{tabular}{l|c|c}
\multirow{2}{*}{Method} & \multicolumn{2}{c}{Sketchy} \\
& acc@1 & acc@10 \\
\hline
Human \cite{Sangkloy2016Sketchy} & 54.27 & -- \\
\hline
SAN-Triplet \cite{Yu2016SketchMe} & 25.87 & -- \\
GN-Siamese \cite{Sangkloy2016Sketchy} & 27.36 & -- \\
GN-Triplet \cite{Sangkloy2016Sketchy} & 37.10 & -- \\
\hline
XDGen \cite{BMVC2017_46} & 50.14 & -- \\
DCCRM (S+I) \cite{Wang2020PR} & 40.16 & 92.00 \\
DCCRM (S+I+D) \cite{Wang2020PR} & 46.20 & 96.49 \\
\hline
\textbf{Ours (XModalViT)} & \textbf{56.15} & \textbf{96.86}
\end{tabular}}
\caption{Quantitative comparison (in \%) with state-of-the-art for fine-grained SBIR on Sketchy dataset.}
\label{tab:SOTAComparisonSketchy}
\end{wraptable}
\cref{tab:SOTAComparisonSketchy}.~\emph{GN-Siamese} \cite{Sangkloy2016Sketchy} and \emph{GN-Triplet} \cite{Sangkloy2016Sketchy} train a GoogleNet \cite{Szegedy2015GoogleNet} with the siamese and triplet losses respectively in a contrastive manner, while \emph{SAN-Triplet} \cite{Yu2016SketchMe} applies the triplet loss on Sketch-a-Net \cite{Yu2017SketchANet}.~\emph{XDGen} \cite{BMVC2017_46} introduces the task of cross-domain image synthesis, achieving the current SOTA on acc@1.
By additionally leveraging textual descriptions and cascaded coarse-to-fine instance-level features, \emph{DCCRM} (S+I+D) \cite{Wang2020PR} is the current SOTA on acc@10.
With our novel XModalViT framework, we were able to beat both the acc@1 and acc@10 SOTA on Sketchy by 6.01\% and 0.37\% respectively, without requiring hard to obtain textual annotations as in DCCRM (S+I+D). We were also able to surpass the average human-level acc@1 performance on Sketchy, as reported in \cite{Sangkloy2016Sketchy}.

{
\setlength{\tabcolsep}{6pt}
\renewcommand{\arraystretch}{0.7}
\begin{figure*}[!t]
\begin{minipage}{\textwidth}
\resizebox{\textwidth}{!}{
\begin{tabular}{@{}c@{}c@{}c@{}c@{}c@{}c@{}c@{}c@{}c@{}c@{}c}
\includegraphics[width=2cm, height=2cm]{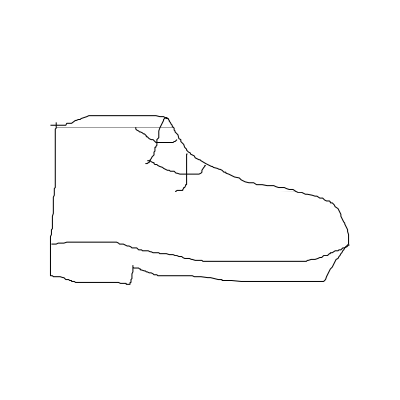} & \includegraphics[width=2cm, height=2cm, cfbox=ForestGreen 3pt 1pt]{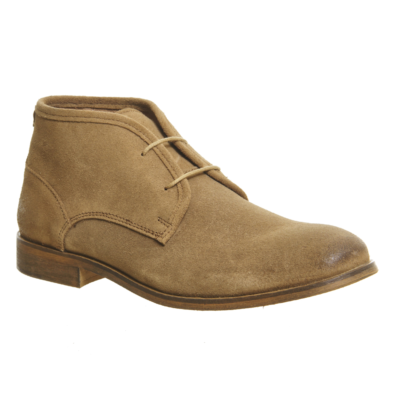} & \includegraphics[width=2cm, height=2cm]{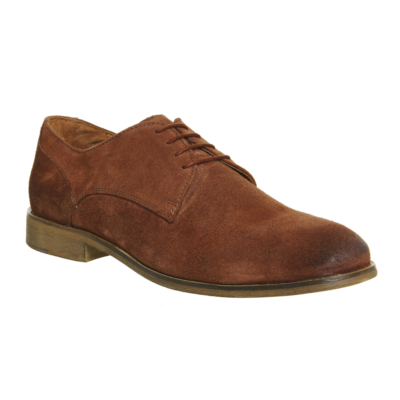} & \includegraphics[width=2cm, height=2cm]{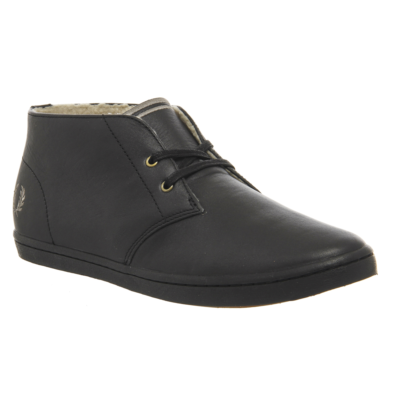} & \includegraphics[width=2cm, height=2cm]{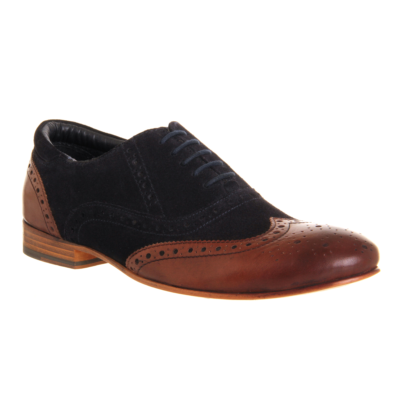} & \includegraphics[width=2cm, height=2cm]{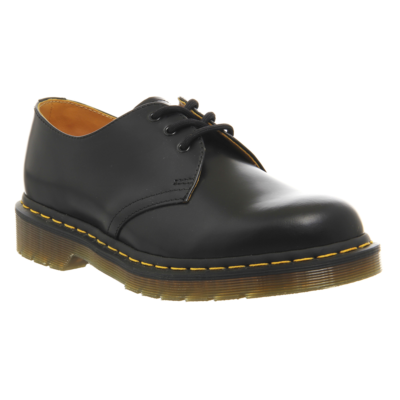} & \includegraphics[width=2cm, height=2cm]{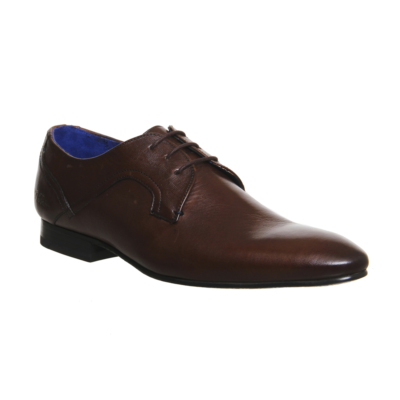} & \includegraphics[width=2cm, height=2cm]{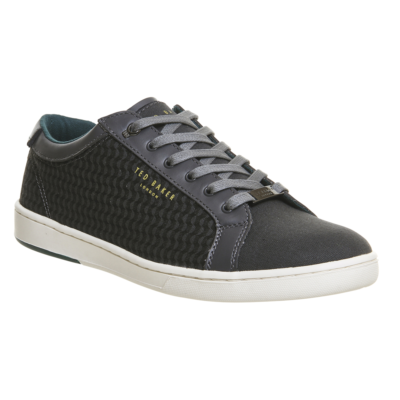} & \includegraphics[width=2cm, height=2cm]{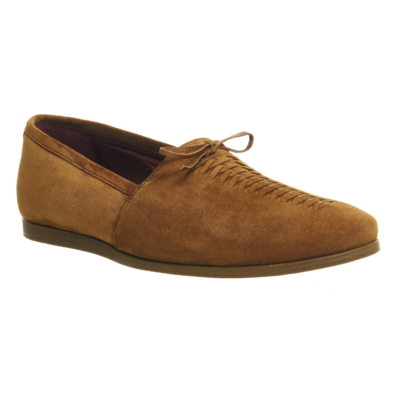} & \includegraphics[width=2cm, height=2cm]{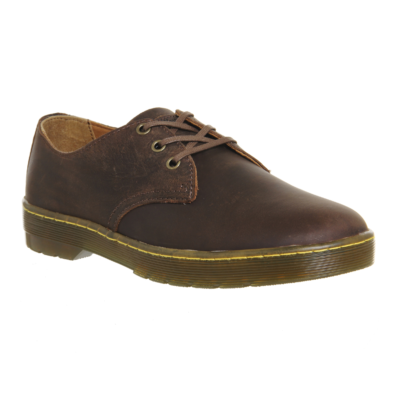} & \includegraphics[width=2cm, height=2cm]{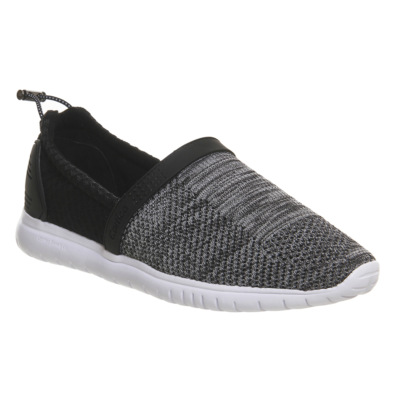} \\ 
\includegraphics[width=2cm, height=2cm]{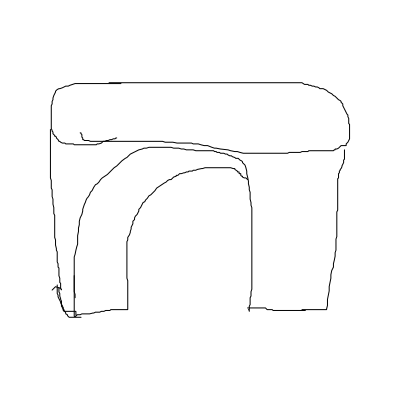} & \includegraphics[width=2cm, height=2cm, cfbox=ForestGreen 3pt 1pt]{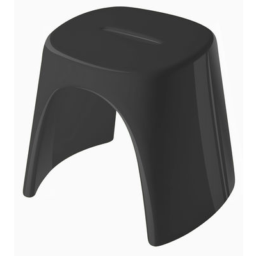} & \includegraphics[width=2cm, height=2cm]{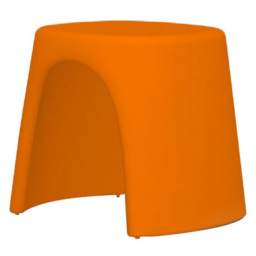} & \includegraphics[width=2cm, height=2cm]{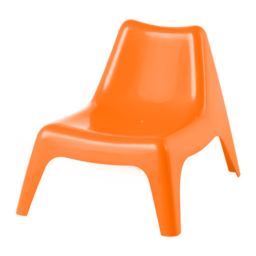} & \includegraphics[width=2cm, height=2cm]{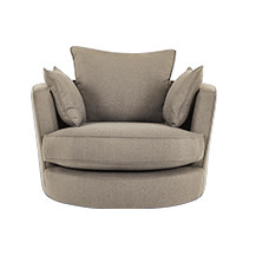} & \includegraphics[width=2cm, height=2cm]{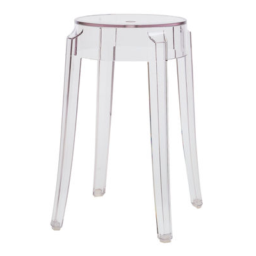} & \includegraphics[width=2cm, height=2cm]{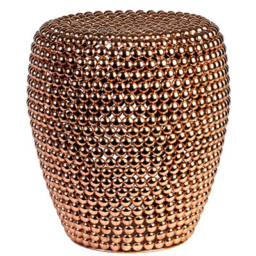} & \includegraphics[width=2cm, height=2cm]{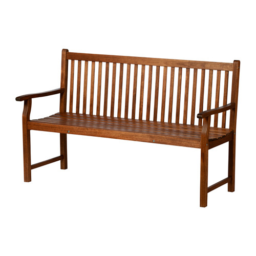} & \includegraphics[width=2cm, height=2cm]{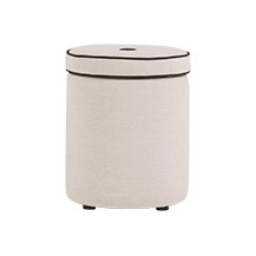} & \includegraphics[width=2cm, height=2cm]{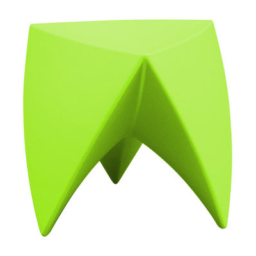} & \includegraphics[width=2cm, height=2cm]{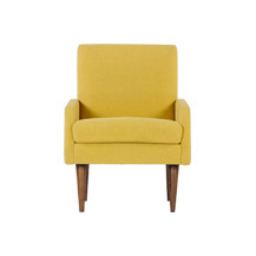} \\ 
\includegraphics[width=2cm, height=2cm]{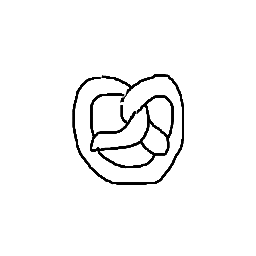} & \includegraphics[width=2cm, height=2cm]{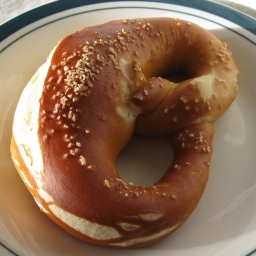} & \includegraphics[width=2cm, height=2cm, cfbox=ForestGreen 4pt 1pt]{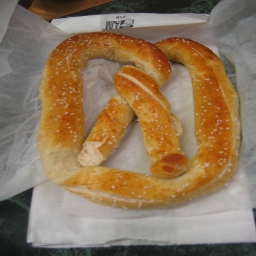} & \includegraphics[width=2cm, height=2cm]{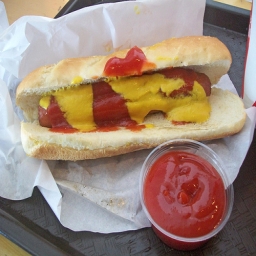} & \includegraphics[width=2cm, height=2cm]{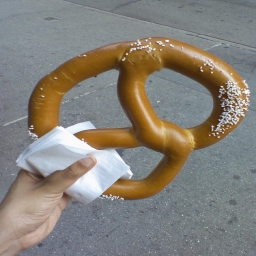} & \includegraphics[width=2cm, height=2cm]{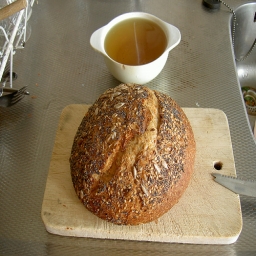} & \includegraphics[width=2cm, height=2cm]{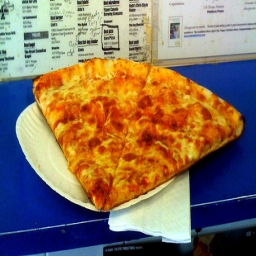} & \includegraphics[width=2cm, height=2cm]{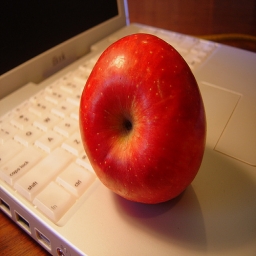} & \includegraphics[width=2cm, height=2cm]{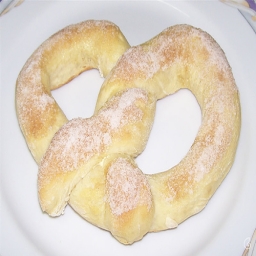} & \includegraphics[width=2cm, height=2cm]{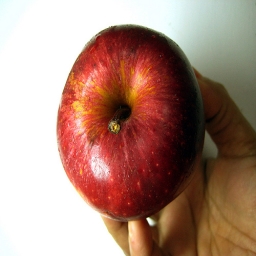} & \includegraphics[width=2cm, height=2cm]{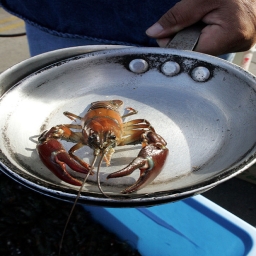} \\
\end{tabular}}
\caption{Qualitative fine-grained SBIR results of our method on the Shoe-V2 (row-1), Chair-V2 (row-2) and Sketchy (row-3) datasets. More qualitative results can be found in the supplementary.
}
\label{fig:qual_results_retrieval_results}
\end{minipage}%
\end{figure*}
}

\subsection{Qualitative Results}
\Cref{fig:qual_results_retrieval_results} depicts examples of retrieval by our model, with additional results in the supplementary. If the target photo has a significant number of discriminative local features compared to others in the gallery, it can be seen to always appear in the top-1. However, correct instances that get demoted in rank are preceded by ones that bear significant resemblance to the fine-grained local features of the query.
For instance, the photo ranked first in the last row can be seen to bear significant resemblance to the query (considering the fact that it belongs to the same class as the ground-truth, \ie, `pretzel', and the curvature pattern of the object).
These results provide evidence that our framework learns to draw fine-grained associations of the underlying concept across the two modalities.

\begin{wrapfigure}[13]{r}{0.4\textwidth}
\vspace{-8pt}
    \centering
    \includegraphics[height=3.8cm,width=\linewidth]{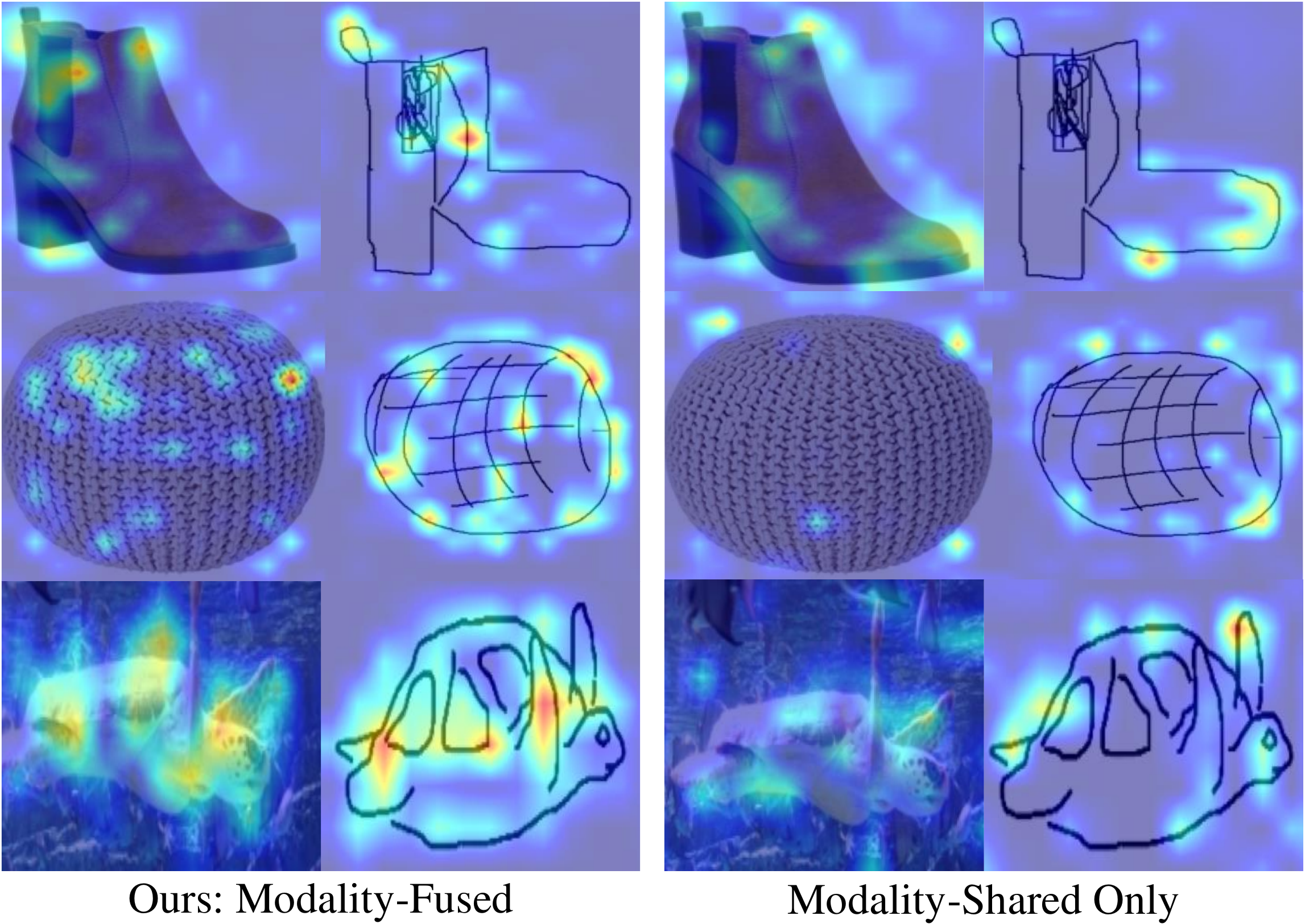}
    \captionof{figure}{Comparison of attention maps obtained from our modality-fused representations and the conventional modality-shared-only approaches.}
    \label{fig:AttentionMaps}
\end{wrapfigure}
In \Cref{fig:AttentionMaps}, we depict examples where modality-specific features uniquely identify an instance. Sketchers use techniques like scribbles to illustrate color/texture differences (Shoe)
and grid/mesh-like structures or primitive shapes like squares/triangles as approximations for complex patterns (Chair and a Turtle's shell).
As an example (row 1), while a large variety of shoes would have the same kind of front, which modality-shared models focus on (right), our model (left), by the virtue of modality fusion (XMA), is capable of attending to more discriminative features like contrasting color/texture depicted by scribbles, or partition denoted by a line, while also capturing shared features like local geometry.
\section{Conclusion}
\label{sec:concl}
We approached the problem of cross-modal representation learning for the task of fine-grained sketch-based image retrieval (FG-SBIR) by taking a detour from the conventional objective for the task. We posited that sketches and photos are instances of an underlying singular abstract concept, and both modalities contain complementary information for constructing a representation of that abstraction. This motivated us to frame our representation learning objective so as to fuse information from local correspondences across both the modalities via the cross-modal attention operation.
We then formulated a technique to decouple the modality-fused representations into independent modality-specific encoders via a contrastive learning objective that would direct the modality specific encoders to converge towards the unified, fused representations, while preserving the geometry of the cross-modal attention space.
We empirically validated the capability of our method to learn expressive representations by achieving state-of-the-art results on FG-SBIR benchmark datasets.


\section*{Acknowledgements}
This work has been partially supported by the ERC 853489--DEXIM, by the DFG--EXC number 2064/1--Project number 390727645, and as part of the Excellence Strategy of the German Federal and State Governments.

\clearpage

\bibliography{egbib}

\clearpage


\end{document}


\graphicspath{{figures/}}
\DeclareGraphicsExtensions{.pdf,.png}

\maketitle

\section*{Supplementary Material}

\section{Further Experimental Details and Findings}
\subsection{Platform Details}
We implement our XModalViT model using the PyTorch \cite{Paszke2017PyTorch} deep learning framework, on an Ubuntu 20.04 workstation with a single Nvidia GeForce RTX 3090 GPU, an 8-core Intel Xeon processor and 32 GBs of RAM. Since we are using a fixed-size queue to store XMA representations, we do not have a dependency on batch-size for the purpose of negative sampling as part of our contrastive learning phase, which enables us to train both the teacher and the students on a single GPU. 

\subsection{Dataset Statistics}

\begin{table}[!ht]
\centering
\resizebox{\columnwidth}{!}{
\begin{tabular}{l|c|c|c|c|c}
Dataset & \#Classes & \#Photos & \#Sketches & \#Sketches/Photos & Test Fraction \\
\hline
QMUL-Shoe-V2 & -- & 2000 & 6648 & 2 to 4 & 0.1 \\
QMUL-Chair-V2 & -- & 400 & 1275 & 2 to 4 & 0.1 \\
\hline
Sketchy & 125 & 12,500 & 75,471 & 5 to 9 & 0.1
\end{tabular}
}
\caption{Details of the datasets used for experimental evaluation.}
\label{tab:Datasets}
\end{table}

\subsection{Effects of Varying the Queue Sizes in XAQC}
\begin{figure}[!ht]
    \centering
    \includegraphics[width=4.5cm, height=3.5cm]{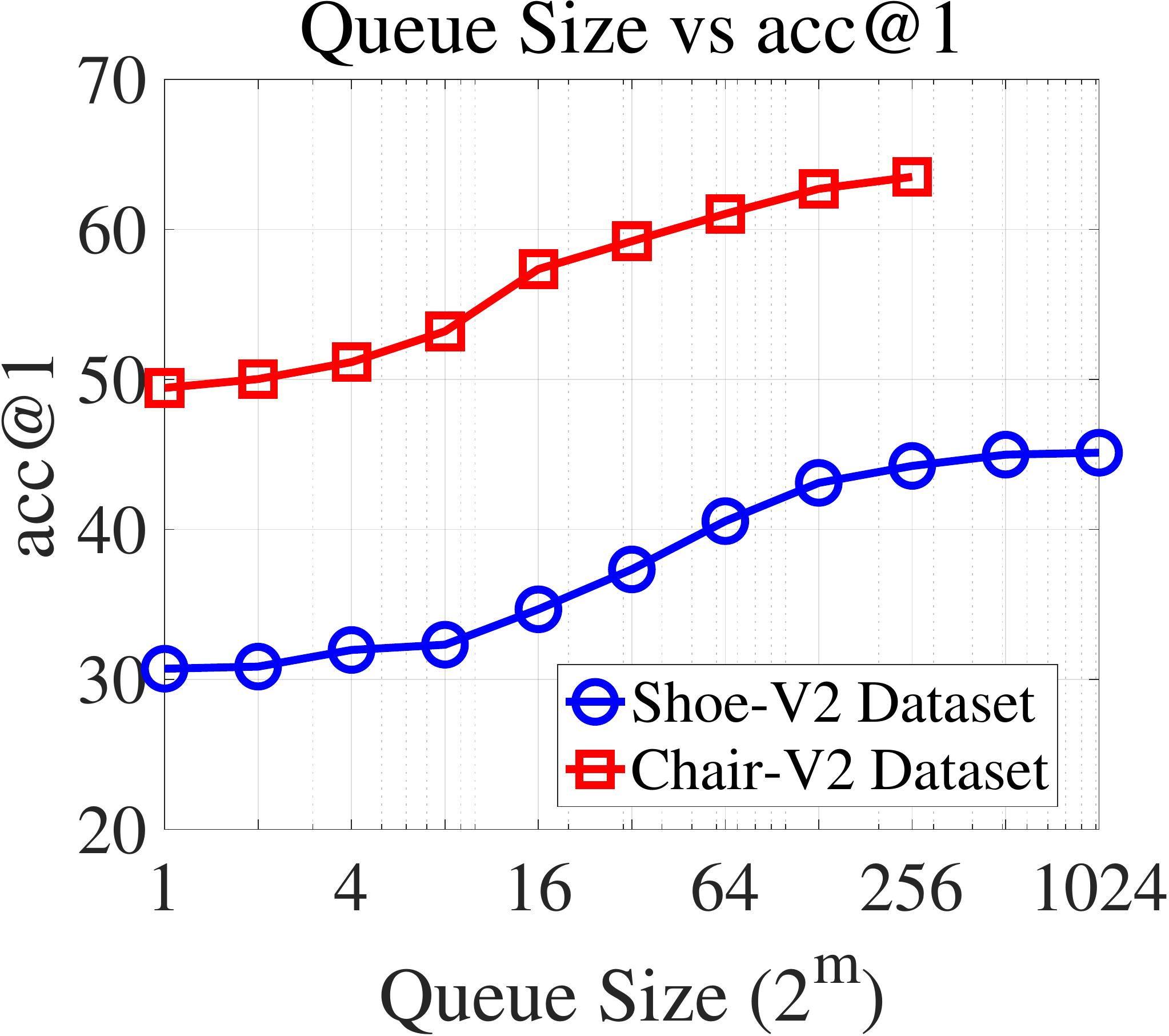}
    \caption{Acc@1 with varying XAQC queue sizes $2^m$, where m $\in \{1, 2, ...,\lfloor\log_2 N\rfloor\}$, and $N$ is the total number of datapoints.}
\label{fig:queue_size_ablation}
\end{figure}

We vary the XAQC queue sizes as $2^m$, where m $\in \{1,2,... \lfloor\log_2 N\rfloor\}$, and $N$ is the total number of datapoints. The observed trend has been graphically depicted in \cref{fig:queue_size_ablation}, where it can be seen that the accuracy is minimum at $m=1$, when the objective is equivalent to InfoNCE, which only performs reasonably with larger batch-sizes. However, the accuracy begins to saturate near its maximum at $m = \lfloor \log_2 N \rfloor$.

\subsection{ViT+CNN Backbones}
While keeping the ViT-B teacher backbones, we switched the student backbones to ResNet18 \cite{He2016DeepRL} and obtained acc@1 of 44.9\% and 63.21\% on the Shoe-V2 and Chair-V2 datasets respectively. However, with the InceptionV3 \cite{Szegedy2016Inceptionv3} network as the student backbones, the acc@1 on the same datasets drop to 42.5\% and 62.07\% respectively. This goes on to show that, once our ViT-Base networks as the teacher encoders learn the fused cross-modal attention representations, the correct inductive biases in CNNs can be used to approximate them. However, for the teacher network, there is no straightforward way to formulate the XMA operator that would make sense over the space of CNN feature maps. The cross-modal interaction between the global class-token of one modality and the local patch embeddings of the other is something that can be more naturally defined for Vision Transformers.

\section{Pseudocodes}
This section provides algorithmic pseudocodes for the core components of the XModalViT framework.

Notations like tuple assignment:
\begin{equation}
    (a, b) = (b, a)
\end{equation}
are inspired by Python's syntax.
The construct ``with no gradient'' signifies that the output of the operation performed within the block is treated like a constant, and its computation does not affect the gradient of any of the learnable parameters that might have been used for that purpose.

\subsection{Modality Fusion Network}
The objective of the \textsc{Modality-Fusion-Network} is to train the cross-modal attention (XMA) based teacher network, $\Gamma$, that fuses information across the photo and the sketch modalities. \Cref{alg:teacher} provides a pseudocode for the same.

\begin{algorithm}[!ht]
\SetKwInOut{Input}{Input}\SetKwInOut{Output}{Output}

\Input{A set of photos $\mathcal{P}$ and corresponding sketches $\mathcal{S}$; Cross-attention queue size $k$; Learning rate $\eta$; Number of epochs $N$}
\Output{Cross-Modal Fusion Network (Teacher): $\Gamma$}

$\mathcal{Q} \longleftarrow$ Randomly initialized queue of size $k$\\
\For{epoch $\leftarrow 1$ \KwTo $N$}{
    $p \sim \mathcal{P}, \; (s_1, s_2) \sim \mathcal{S} \; \mid \; p \xleftrightarrow{} (s_1, s_2)$\\
    
    $(\vectorname{x}^1_p, \vectorname{x}^1_s) \longleftarrow \Gamma(p, s_1)$ \\
    
    \tcp{Outputs are treated as constants}
    with no gradient: \\
     \Indp $(\vectorname{x}^2_p, \vectorname{x}^2_s) \longleftarrow \Gamma(p, s_2)$ \\
     \Indm
    
    $\mathcal{L}_{\text{teacher}} \longleftarrow \textbf{\texttt{XAQC}}(\vectorname{x}^1_p, \vectorname{x}^2_s, \mathcal{Q})$ \\
    
    $\mathcal{Q} \longleftarrow \mathcal{Q}$.enqueue$(\vectorname{x}^2_s)$\\
    $\Gamma \longleftarrow \Gamma - \eta \nabla_\Gamma \mathcal{L}_{\text{teacher}}$
}
\caption{\textsc{Modality-Fusion-Network}: Train the cross-modal attention (XMA) based teacher network, $\Gamma$, that fuses information across the photo and the sketch modalities.}
\label{alg:teacher}
\end{algorithm}

$\mathcal{Q}$ is initialized as a queue of size $k$ containing random vectors, which is later used for storing XMA-sketch representations.
We sample a photo $p$ and 2 of its corresponding sketches $s_1$ and $s_2$. In line-5, $(p, s_1)$ is propagated through the teacher network for obtaining its XMA-photo embedding, $\vectorname{x}^1_p$. The XMA-sketch embedding, $\vectorname{x}^2_s$, for $(p, s_1)$ is computed in this manner. We then compute the XAQC loss for the teacher by treating $\vectorname{x}^2_s$ as a soft-target for $\vectorname{x}^1_p$ and all the representations in $\vectorname{x}^1_s$ as negatives. We then enqueue $\vectorname{x}^1_s$ into $\mathcal{Q}$ in line-9. We finally update the teacher network with the gradient of the XAQC loss with respect to its parameters.

\subsection{Cross-Modal Knowledge Distillation}

\begin{algorithm}[!ht]
\SetKwInOut{Input}{Input}\SetKwInOut{Output}{Output}

\Input{A set of photos $\mathcal{P}$ and corresponding sketches $\mathcal{S}$; Teacher network $\Gamma$; Cross-attention queue size $k$; Learning rate $\eta$; Number of epochs $N$}
\Output{Student Networks - Photo encoder $\xi_{\text{photo}}$ and sketch encoder $\xi_{\text{sketch}}$}

$\mathcal{Q}_\mathcal{P}, \mathcal{Q}_\mathcal{S} \longleftarrow$ Randomly initialized queues of size $k$\\
\For{epoch $\leftarrow 1$ \KwTo $N$}{
    $p_1 \sim \mathcal{P}, \; s_1 \sim \mathcal{S} \; \mid \; p_1 \xleftrightarrow{} s_1$ \\
    
    $p_2, p_3 \sim \mathcal{P}, \; s_2, s_3 \sim \mathcal{S} \; \mid \; p_2 \xleftrightarrow{} s_2, p_3 \xleftrightarrow{} s_3$ \\
    
    \tcp{Outputs are treated as constants}
    with no gradient: \\
        \For{$i \leftarrow 1$ \KwTo $3$}{
        $(\vectorname{x}_{p_i}, \vectorname{x}_{s_i}) \longleftarrow \Gamma(p_i, s_i)$ \\
        }
    
    \For{$i \leftarrow 1$ \KwTo $3$}{
        $(\vectorname{z}_{p_i}, \vectorname{z}_{s_i}) \longleftarrow \xi_\text{photo}(p_i), \xi_\text{sketch}(s_i)$ \\
    }
    $\mathcal{L}^\text{photo-student}_\text{XAQC} \longleftarrow \textbf{\texttt{XAQC}}(\vectorname{x}_{p_1}, \vectorname{z}_{p_1}, \mathcal{Q}_\mathcal{P})$ \\
    
    $\mathcal{L}^\text{sketch-student}_\text{XAQC} \longleftarrow \textbf{\texttt{XAQC}}(\vectorname{x}_{s_1}, \vectorname{z}_{s_1}, \mathcal{Q}_\mathcal{S})$ \\
    
    $\mathcal{Q}_\mathcal{P} \longleftarrow \mathcal{Q}_\mathcal{P}$.enqueue$(\vectorname{x}_{p_1})$ \\
    $\mathcal{Q}_\mathcal{S} \longleftarrow \mathcal{Q}_\mathcal{S}$.enqueue$(\vectorname{x}_{s_1})$ \\
    
    \tcp{$m \in \{p, m\}$}
    $\pi^\text{teacher}_m \longleftarrow \psi_1(\vectorname{x}_{m_1}, \vectorname{x}_{m_2}) + \psi_2(\vectorname{x}_{m_1}, \vectorname{x}_{m_2}, \vectorname{x}_{m_3})$ \\
    $\pi^\text{student}_m \longleftarrow \psi_1(\vectorname{z}_{m_1}, \vectorname{z}_{m_2}) + \psi_2(\vectorname{z}_{m_1}, \vectorname{z}_{m_2}, \vectorname{z}_{m_3})$
    
    $\mathcal{L}_\text{XMRD} \longleftarrow
    \delta(\pi^\text{teacher}_p, \pi^\text{student}_p) + 
    \delta(\pi^\text{teacher}_s, \pi^\text{student}_s)$
    
    $\mathcal{L}_\text{student} \longleftarrow
    \mathcal{L}^\text{sketch-student}_\text{XAQC} + \mathcal{L}^\text{photo-student}_\text{XAQC} + \lambda \cdot \mathcal{L}_\text{XMRD}$

    $\xi_\text{photo} \longleftarrow \xi_\text{photo} - \nabla_{\xi_\text{photo}} \mathcal{L}_\text{student}$
    
    $\xi_\text{sketch} \longleftarrow \xi_\text{sketch} - \nabla_{\xi_\text{sketch}} \mathcal{L}_\text{student}$
}
\caption{\textsc{Cross-Modal-Knowledge-Distillation}: Decouple the domain of $\Gamma$ by transferring its representations to independent encoders, $\xi_\text{photo}$ and $\xi_\text{sketch}$.}
\label{alg:student}
\end{algorithm}

The objective of cross-modal knowledge distillation is to decouple the input-space of the modality fusion network (teacher), $\Gamma$, into independent, modality-specific encoders, $\xi_\text{photo}$ and $\xi_\text{sketch}$. \Cref{alg:student} describes the process in the form of a pseudocode.

We start by sampling 3 corresponding photo-sketch pairs $(p_1, s_1), (p_2, s_2)$ and $(p_3, s_3).$ We propagate these pairs through the modality fusion network (teacher), $\Gamma$, to obtain their XMA-photo and XMA-sketch embeddings. The photos and the sketches are then separately encoded via the photo and the sketch students respectively, to obtain the approximate versions of their XMA representations, \ie, $\vectorname{z}_{p_i}$ and $\vectorname{z}_{s_i}$. With the objective of aligning these $z_\mathlarger{*}$ representations with the true XMA representations, $x_\mathlarger{*}$, obtained from the teacher, we minimize the contrastive XAQC loss between the two (lines 11 and 12). We also aim to preserve the geometry of the teacher's representation space in that of the students' by distilling distance and angular relationships between arbitrary $k$-tuples of datapoints. This process of cross-modal relation distillation (XMRD) is depicted in lines 16-18. $m$, here, has been introduced for the purpose of conciseness, serving as an abstract notation for modality, \ie, standing for both photos and sketches. $\psi_1$ and $\psi_2$ are the distance and angle relation functions respectively, and $\delta$ is the Huber loss, as described in the main text. The total student loss is computed as the sum of the losses from the individual students and the XMRD loss (weighted by a balancing factor of $\lambda$). We finally update the individual students by the gradient of the total student loss with respect to the weights of the corresponding student encoders.

\section{Embedding Visualizations}
\begin{figure}[!ht]
\centering
\resizebox{0.8\textwidth}{!}{
\begin{tabular}{c}
\includegraphics[width=0.71\textwidth]{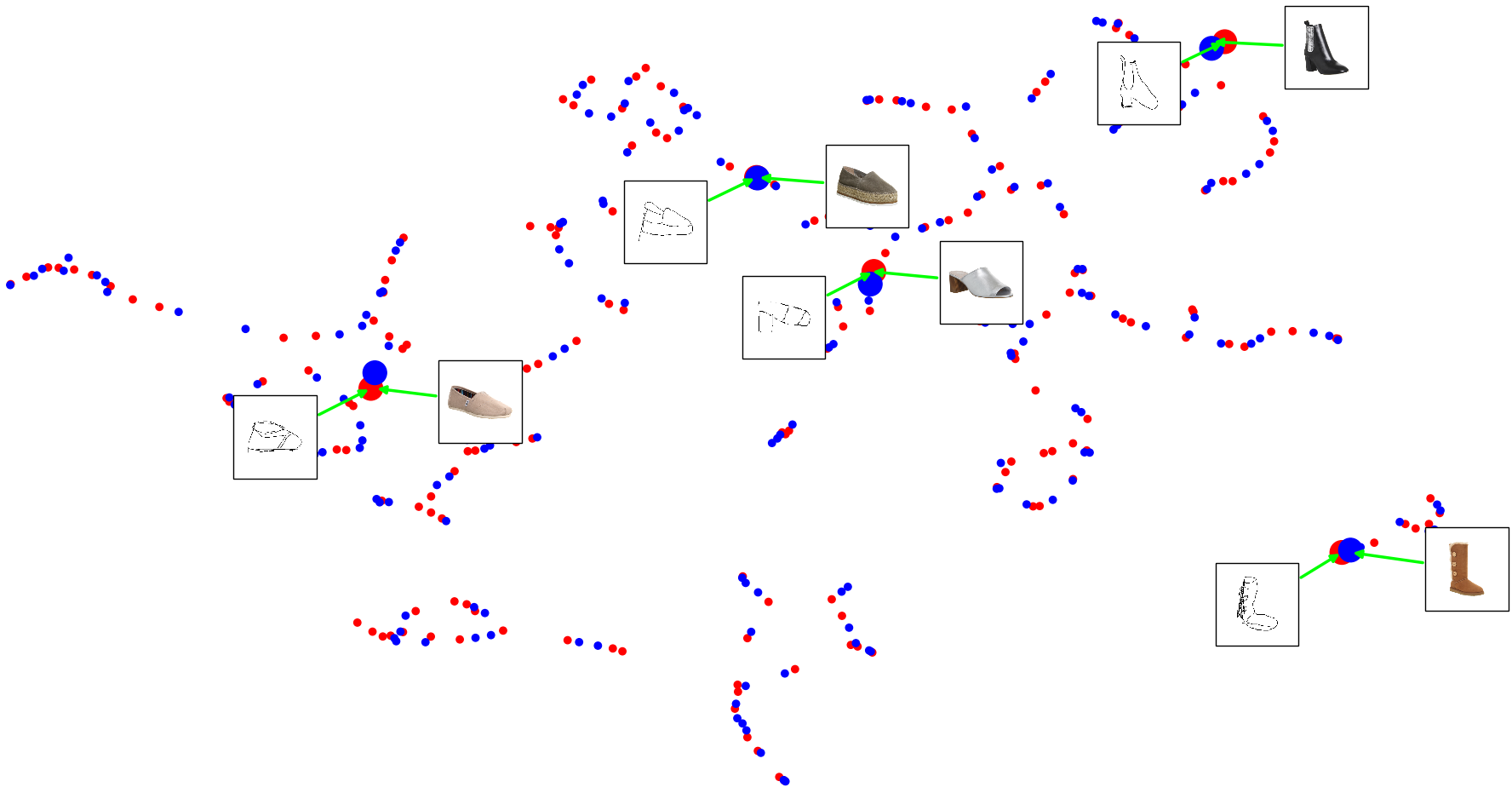} \\
(a) \\
\includegraphics[width=0.71\textwidth]{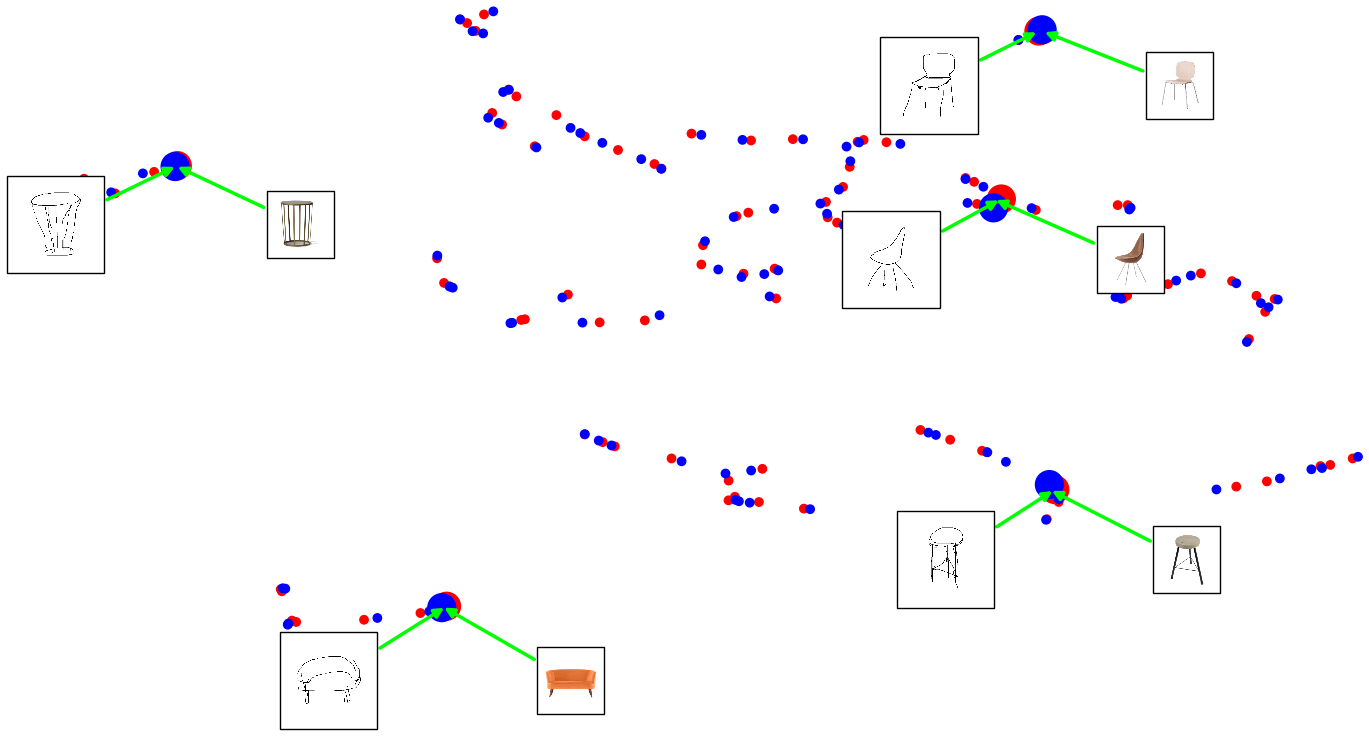} \\
(b) \\
\includegraphics[width=0.71\textwidth]{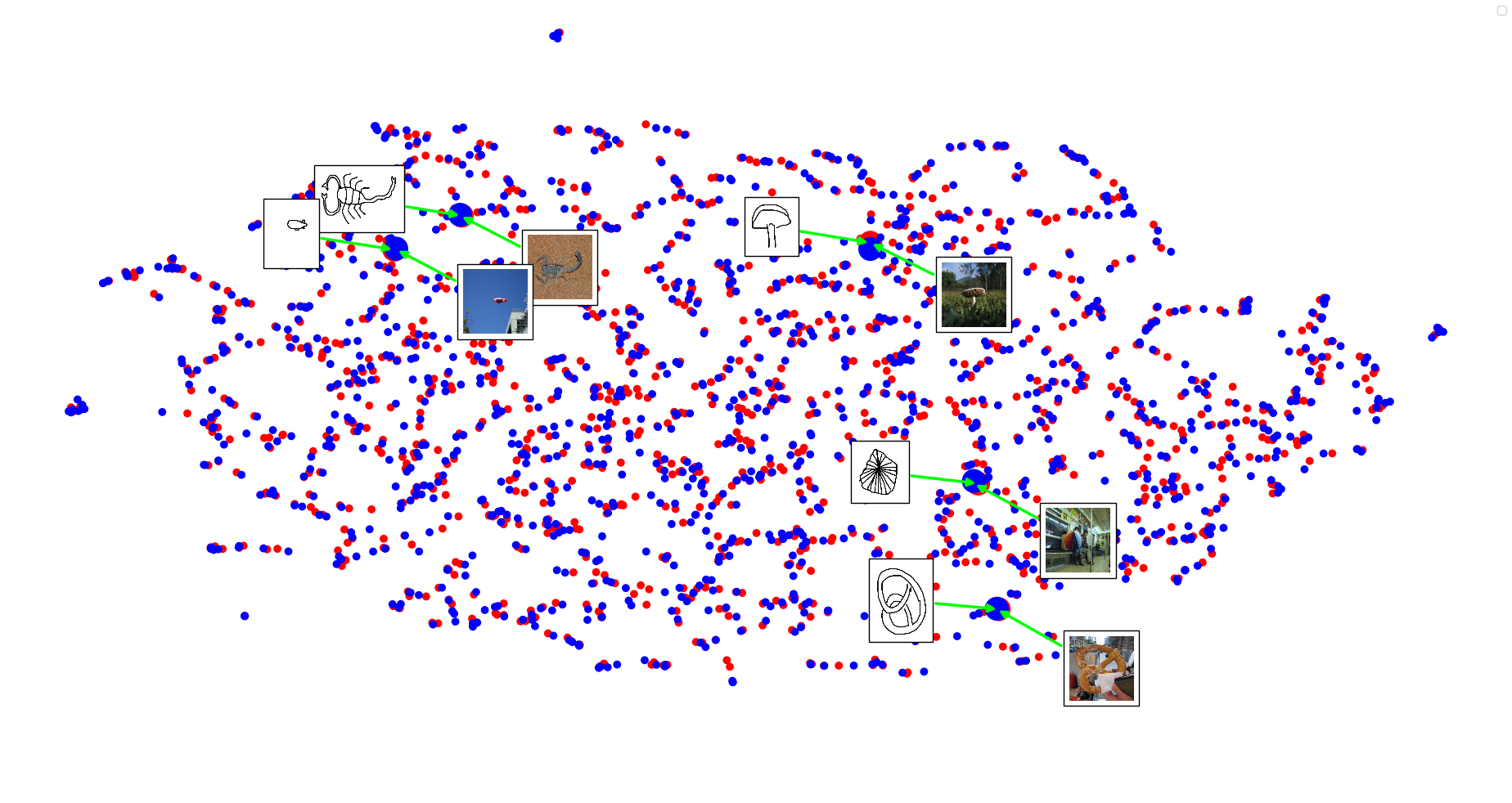} \\
(c)
\end{tabular}
}
\caption{UMAP \cite{mcinnes2018umap-software} visualizations of photo-student (red) and sketch-student (blue) embeddings on the (a) QMUL-Shoe-V2 (b) QMUL-Chair-V2 and (c) Sketchy datasets.} \label{fig:umap_qmul}
\end{figure}

\Cref{fig:umap_qmul} shows the test-set embeddings of the photo (red) and sketch (blue) student encoders, projected onto a 2-dimensional space via UMAP \cite{mcinnes2018umap-software}. The layout of the datapoints across the two modalities can be seen as being very similar, indicating that both the encoders have learned to model the distribution of the underlying shared abstract concept. As a result of this, and by the virtue of the instance-discriminative XAQC loss, photo-sketch pairs of the same instance get mapped close to each other, a phenomenon that has also been depicted in the visualizations.

\newpage

\section{Retrieval Results}

\begin{figure}[!ht]
\begin{center}
\resizebox{\textwidth}{!}{
\begin{tabular}{@{}c@{}c@{}c@{}c@{}c@{}c@{}c@{}c@{}c@{}c@{}c@{}c}
& & {\huge 1 \hspace{6pt}} & {\huge 2 \hspace{6pt}} & {\huge 3 \hspace{6pt}} & {\huge 4 \hspace{6pt}} & {\huge 5 \hspace{6pt}} & {\huge 6 \hspace{6pt}} & {\huge 7 \hspace{6pt}} & {\huge 8 \hspace{6pt}} & {\huge 9 \hspace{6pt}} & {\huge 10 \hspace{6pt}} \\
\raisebox{1.5cm}{\huge 1 \hspace{6pt}} & \includegraphics[width=3cm, height=3cm]{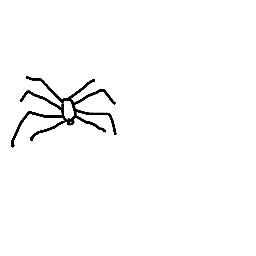} & \includegraphics[width=3cm, height=3cm]{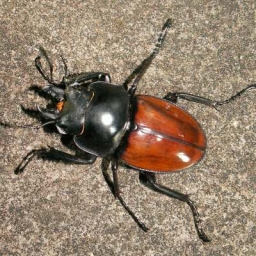} & \includegraphics[width=3cm, height=3cm, cfbox=ForestGreen 4pt 4pt]{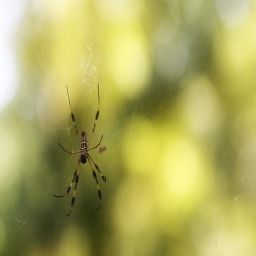} & \includegraphics[width=3cm, height=3cm]{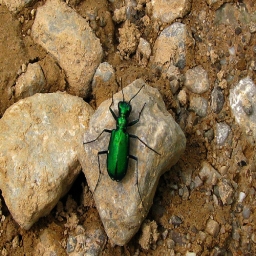} & \includegraphics[width=3cm, height=3cm]{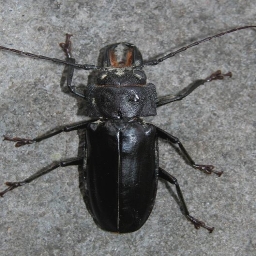} & \includegraphics[width=3cm, height=3cm]{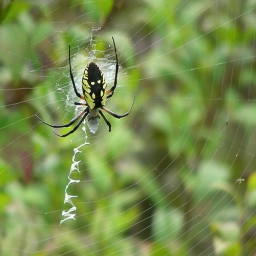} & \includegraphics[width=3cm, height=3cm]{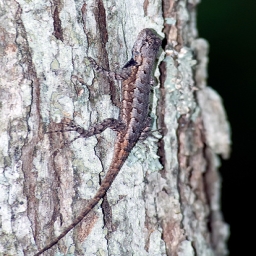} & \includegraphics[width=3cm, height=3cm]{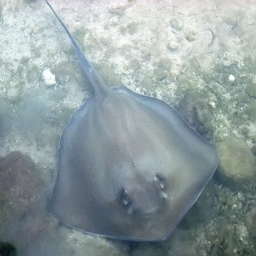} & \includegraphics[width=3cm, height=3cm]{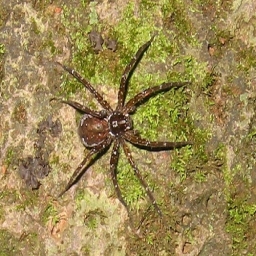} & \includegraphics[width=3cm, height=3cm]{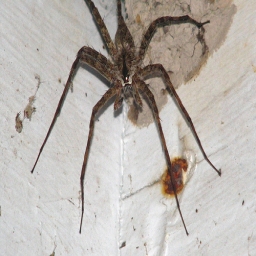} & \includegraphics[width=3cm, height=3cm]{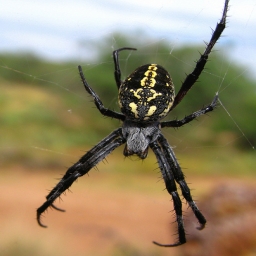} \\ 
\raisebox{1.5cm}{\huge 2 \hspace{6pt}} & \includegraphics[width=3cm, height=3cm]{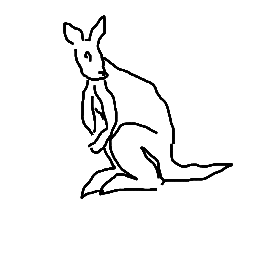} & \includegraphics[width=3cm, height=3cm]{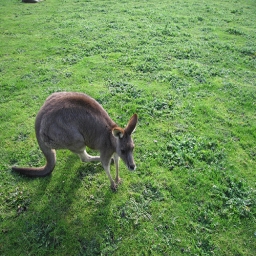} & \includegraphics[width=3cm, height=3cm, cfbox=ForestGreen 4pt 4pt]{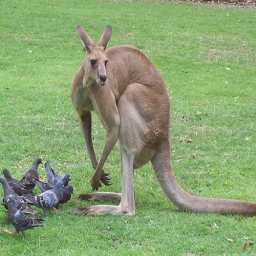} & \includegraphics[width=3cm, height=3cm]{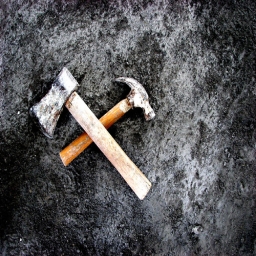} & \includegraphics[width=3cm, height=3cm]{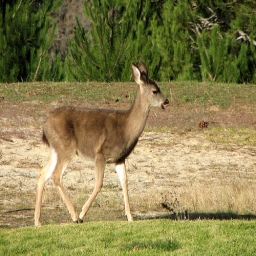} & \includegraphics[width=3cm, height=3cm]{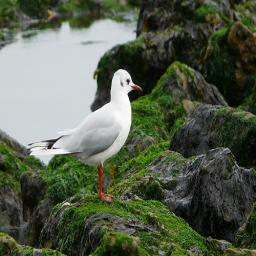} & \includegraphics[width=3cm, height=3cm]{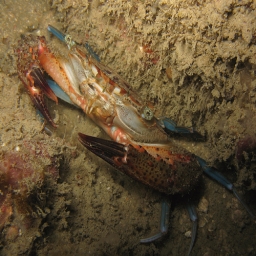} & \includegraphics[width=3cm, height=3cm]{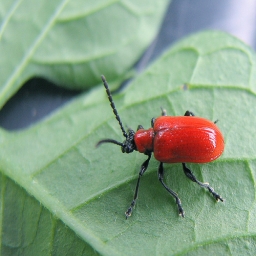} & \includegraphics[width=3cm, height=3cm]{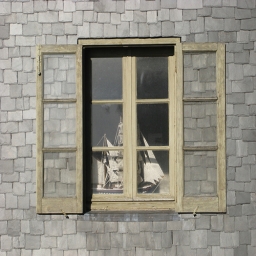} & \includegraphics[width=3cm, height=3cm]{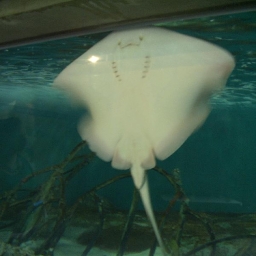} & \includegraphics[width=3cm, height=3cm]{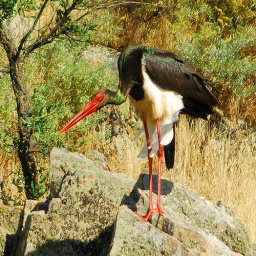} \\ 
\raisebox{1.5cm}{\huge 3 \hspace{6pt}} & \includegraphics[width=3cm, height=3cm]{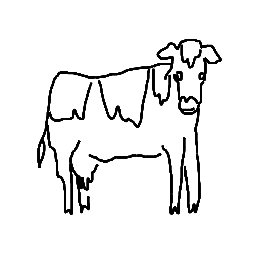} & \includegraphics[width=3cm, height=3cm]{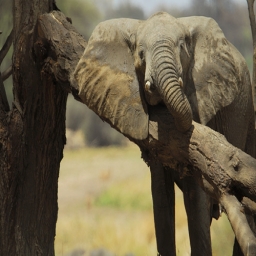} & \includegraphics[width=3cm, height=3cm, cfbox=ForestGreen 4pt 4pt]{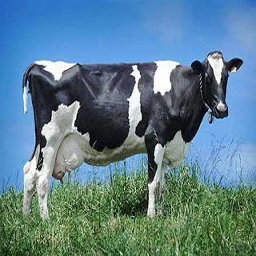} & \includegraphics[width=3cm, height=3cm]{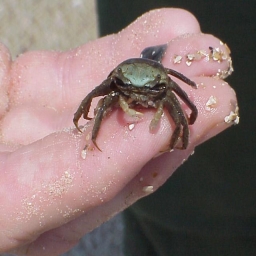} & \includegraphics[width=3cm, height=3cm]{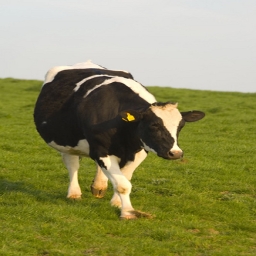} & \includegraphics[width=3cm, height=3cm]{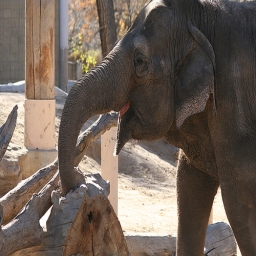} & \includegraphics[width=3cm, height=3cm]{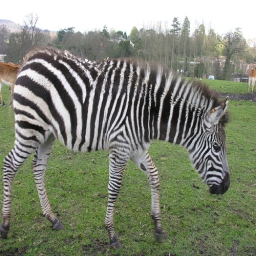} & \includegraphics[width=3cm, height=3cm]{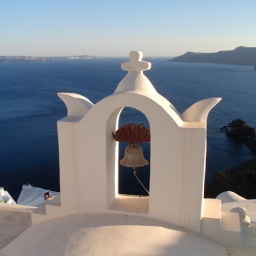} & \includegraphics[width=3cm, height=3cm]{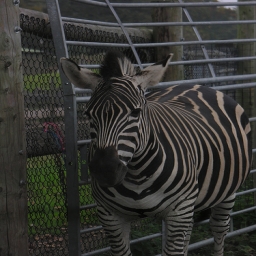} & \includegraphics[width=3cm, height=3cm]{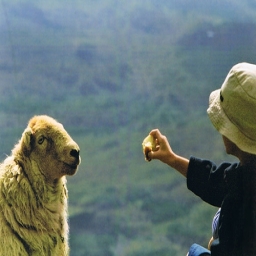} & \includegraphics[width=3cm, height=3cm]{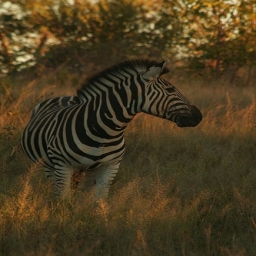} \\ 
\raisebox{1.5cm}{\huge 4 \hspace{6pt}} & \includegraphics[width=3cm, height=3cm]{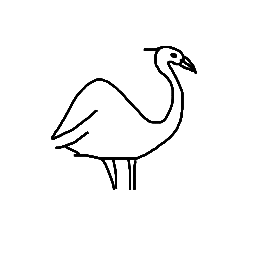} & \includegraphics[width=3cm, height=3cm, cfbox=ForestGreen 4pt 4pt]{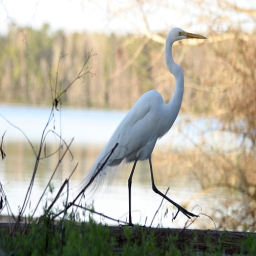} & \includegraphics[width=3cm, height=3cm]{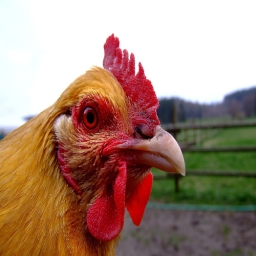} & \includegraphics[width=3cm, height=3cm]{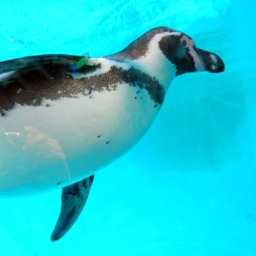} & \includegraphics[width=3cm, height=3cm]{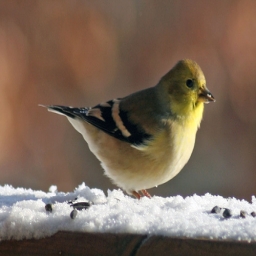} & \includegraphics[width=3cm, height=3cm]{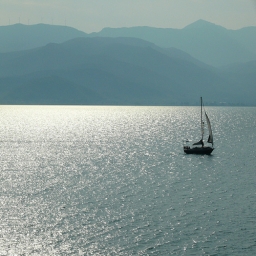} & \includegraphics[width=3cm, height=3cm]{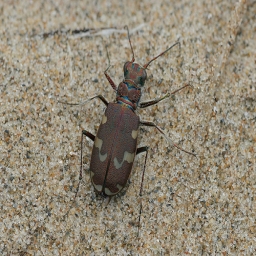} & \includegraphics[width=3cm, height=3cm]{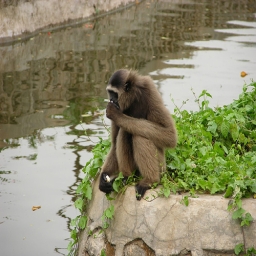} & \includegraphics[width=3cm, height=3cm]{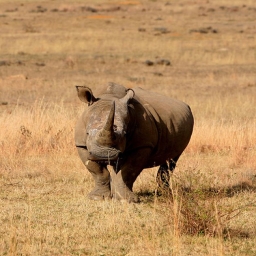} & \includegraphics[width=3cm, height=3cm]{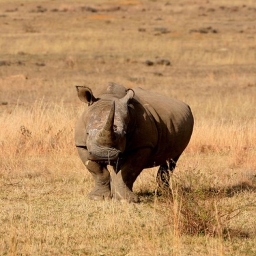} & \includegraphics[width=3cm, height=3cm]{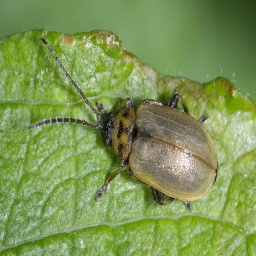} \\ 
\raisebox{1.5cm}{\huge 5 \hspace{6pt}} & \includegraphics[width=3cm, height=3cm]{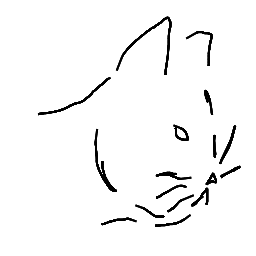} & \includegraphics[width=3cm, height=3cm, cfbox=ForestGreen 4pt 4pt]{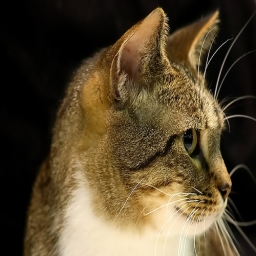} & \includegraphics[width=3cm, height=3cm]{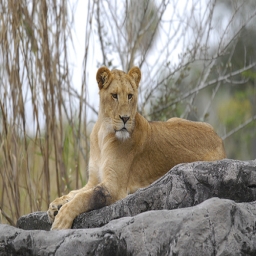} & \includegraphics[width=3cm, height=3cm]{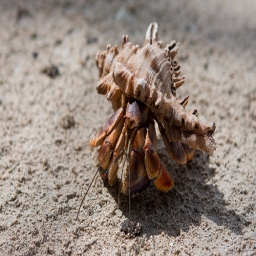} & \includegraphics[width=3cm, height=3cm]{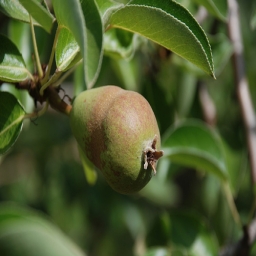} & \includegraphics[width=3cm, height=3cm]{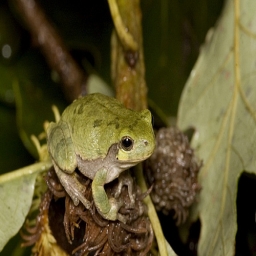} & \includegraphics[width=3cm, height=3cm]{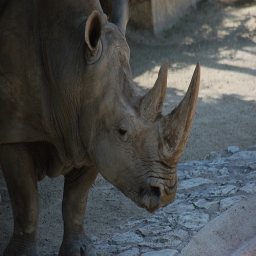} & \includegraphics[width=3cm, height=3cm]{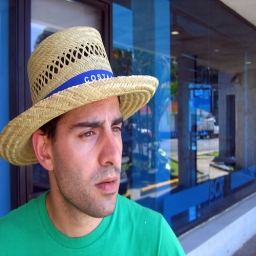} & \includegraphics[width=3cm, height=3cm]{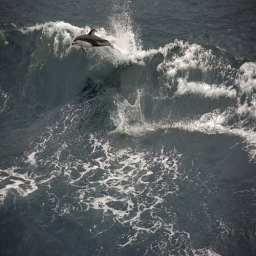} & \includegraphics[width=3cm, height=3cm]{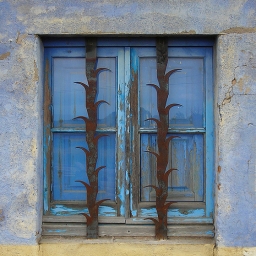} & \includegraphics[width=3cm, height=3cm]{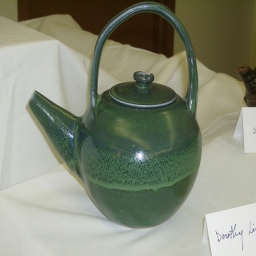} \\ 
\raisebox{1.5cm}{\huge 6 \hspace{6pt}} & \includegraphics[width=3cm, height=3cm]{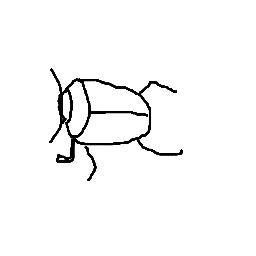} & \includegraphics[width=3cm, height=3cm]{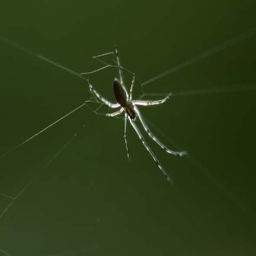} & \includegraphics[width=3cm, height=3cm]{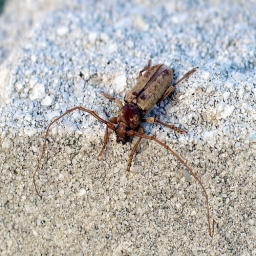} & \includegraphics[width=3cm, height=3cm]{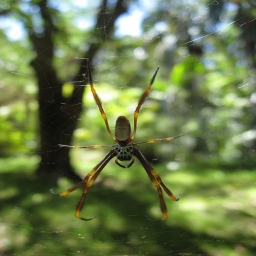} & \includegraphics[width=3cm, height=3cm, cfbox=ForestGreen 4pt 4pt]{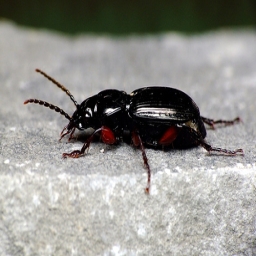} & \includegraphics[width=3cm, height=3cm]{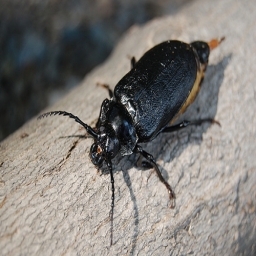} & \includegraphics[width=3cm, height=3cm]{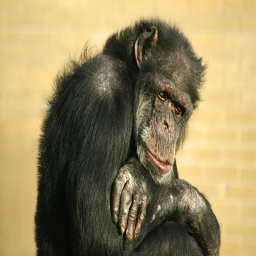} & \includegraphics[width=3cm, height=3cm]{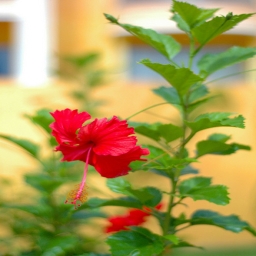} & \includegraphics[width=3cm, height=3cm]{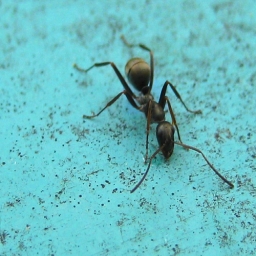} & \includegraphics[width=3cm, height=3cm]{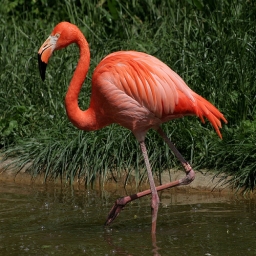} & \includegraphics[width=3cm, height=3cm]{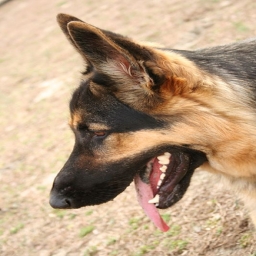} \\ 
\raisebox{1.5cm}{\huge 7 \hspace{6pt}} & \includegraphics[width=3cm, height=3cm]{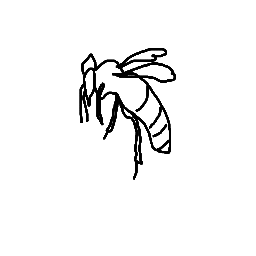} & \includegraphics[width=3cm, height=3cm, cfbox=ForestGreen 4pt 4pt]{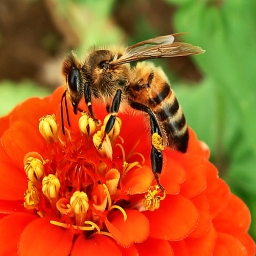} & \includegraphics[width=3cm, height=3cm]{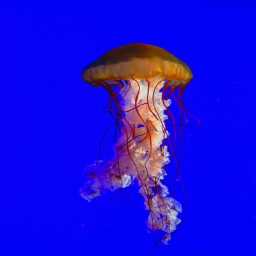} & \includegraphics[width=3cm, height=3cm]{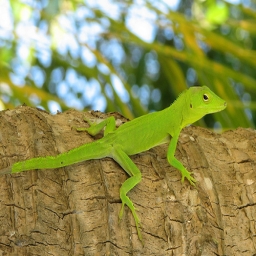} & \includegraphics[width=3cm, height=3cm]{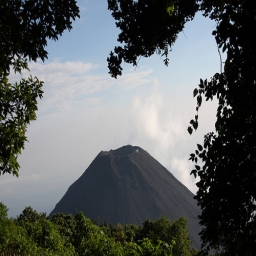} & \includegraphics[width=3cm, height=3cm]{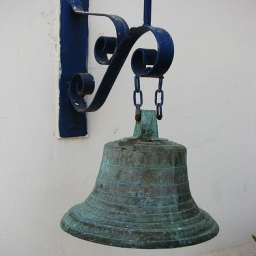} & \includegraphics[width=3cm, height=3cm]{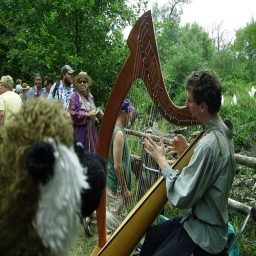} & \includegraphics[width=3cm, height=3cm]{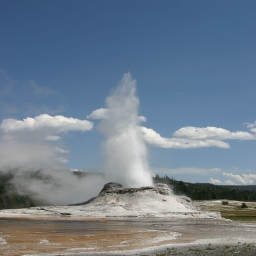} & \includegraphics[width=3cm, height=3cm]{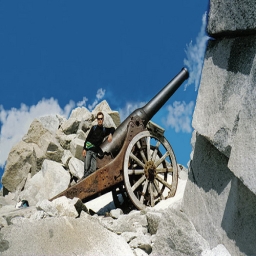} & \includegraphics[width=3cm, height=3cm]{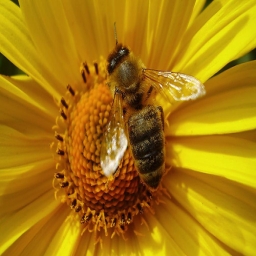} & \includegraphics[width=3cm, height=3cm]{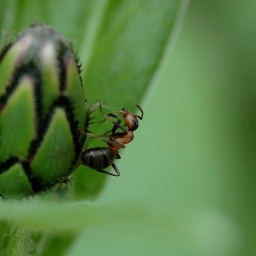} \\ 
\raisebox{1.5cm}{\huge 8 \hspace{6pt}} & \includegraphics[width=3cm, height=3cm]{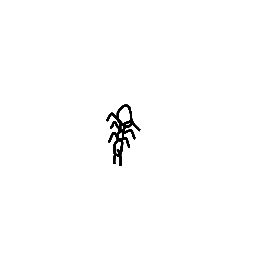} & \includegraphics[width=3cm, height=3cm, cfbox=ForestGreen 4pt 4pt]{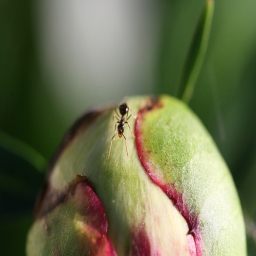} & \includegraphics[width=3cm, height=3cm]{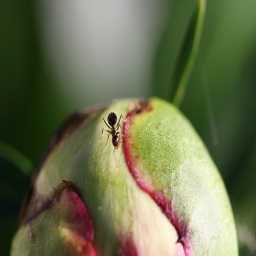} & \includegraphics[width=3cm, height=3cm]{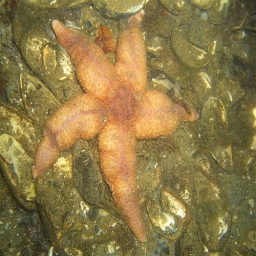} & \includegraphics[width=3cm, height=3cm]{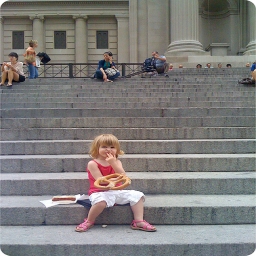} & \includegraphics[width=3cm, height=3cm]{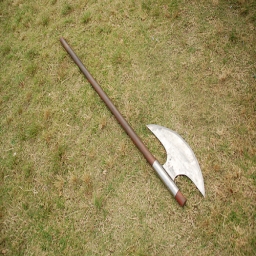} & \includegraphics[width=3cm, height=3cm]{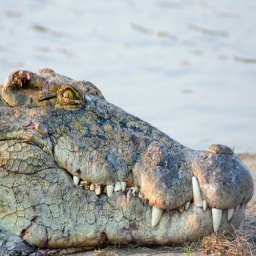} & \includegraphics[width=3cm, height=3cm]{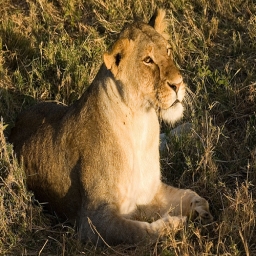} & \includegraphics[width=3cm, height=3cm]{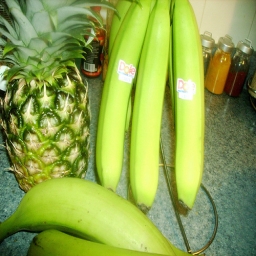} & \includegraphics[width=3cm, height=3cm]{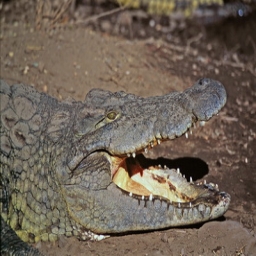} & \includegraphics[width=3cm, height=3cm]{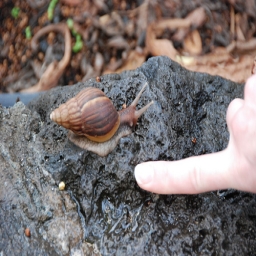} \\ 
\raisebox{1.5cm}{\huge 9 \hspace{6pt}} & \includegraphics[width=3cm, height=3cm]{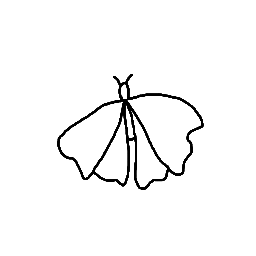} & \includegraphics[width=3cm, height=3cm, cfbox=ForestGreen 4pt 4pt]{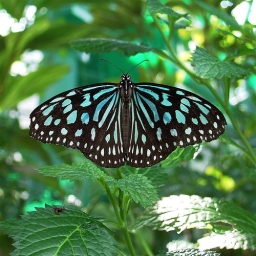} & \includegraphics[width=3cm, height=3cm]{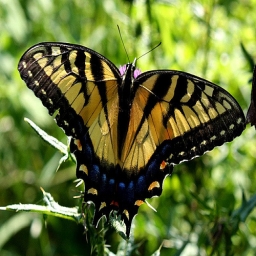} & \includegraphics[width=3cm, height=3cm]{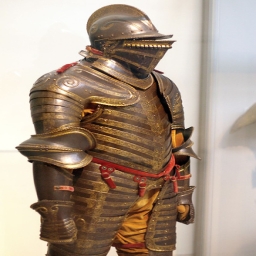} & \includegraphics[width=3cm, height=3cm]{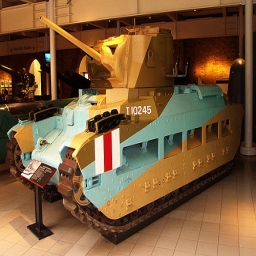} & \includegraphics[width=3cm, height=3cm]{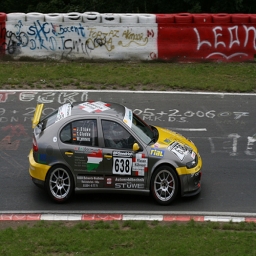} & \includegraphics[width=3cm, height=3cm]{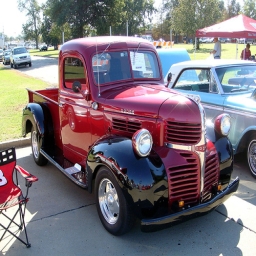} & \includegraphics[width=3cm, height=3cm]{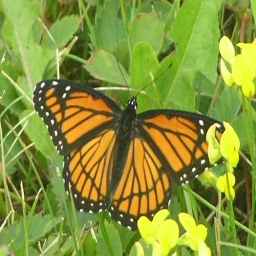} & \includegraphics[width=3cm, height=3cm]{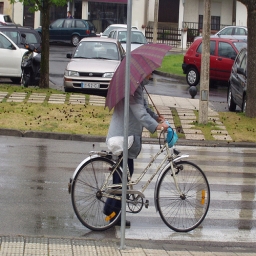} & \includegraphics[width=3cm, height=3cm]{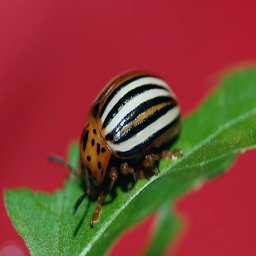} & \includegraphics[width=3cm, height=3cm]{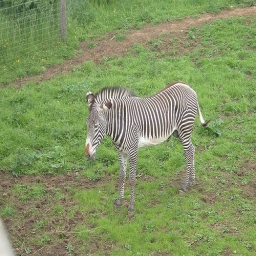} \\ 
\raisebox{1.5cm}{\huge 10 \hspace{6pt}} & \includegraphics[width=3cm, height=3cm]{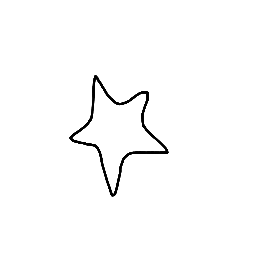} & \includegraphics[width=3cm, height=3cm, cfbox=ForestGreen 4pt 4pt]{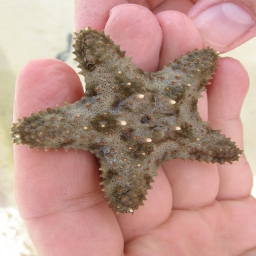} & \includegraphics[width=3cm, height=3cm]{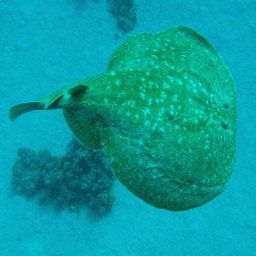} & \includegraphics[width=3cm, height=3cm]{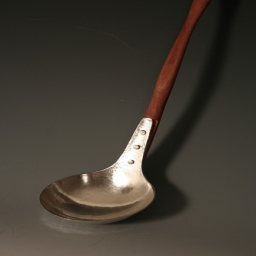} & \includegraphics[width=3cm, height=3cm]{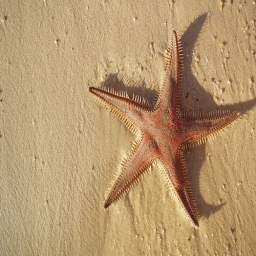} & \includegraphics[width=3cm, height=3cm]{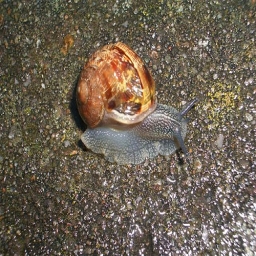} & \includegraphics[width=3cm, height=3cm]{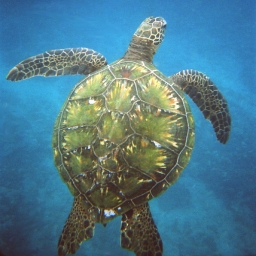} & \includegraphics[width=3cm, height=3cm]{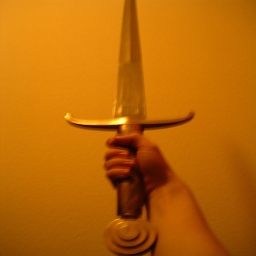} & \includegraphics[width=3cm, height=3cm]{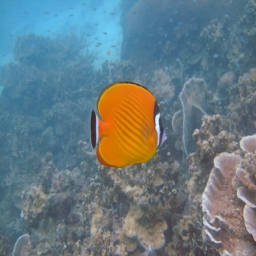} & \includegraphics[width=3cm, height=3cm]{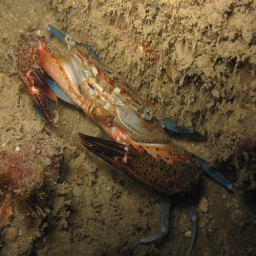} & \includegraphics[width=3cm, height=3cm]{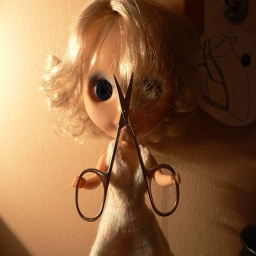} \\ 
\raisebox{1.5cm}{\huge 11 \hspace{6pt}} & \includegraphics[width=3cm, height=3cm]{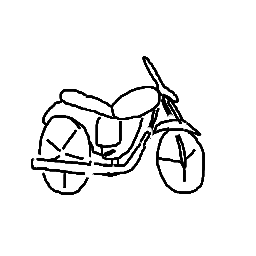} & \includegraphics[width=3cm, height=3cm]{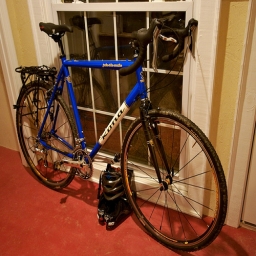} & \includegraphics[width=3cm, height=3cm, cfbox=ForestGreen 4pt 4pt]{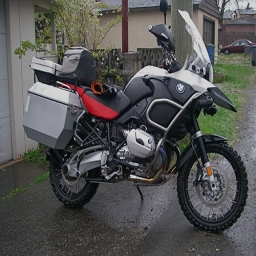} & \includegraphics[width=3cm, height=3cm]{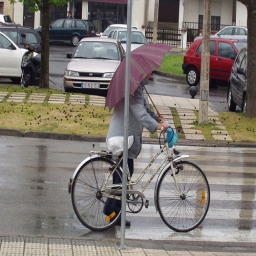} & \includegraphics[width=3cm, height=3cm]{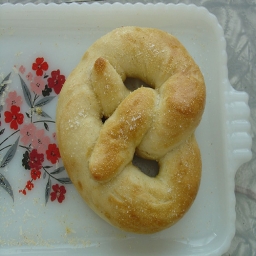} & \includegraphics[width=3cm, height=3cm]{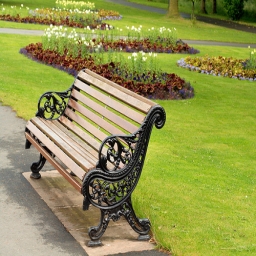} & \includegraphics[width=3cm, height=3cm]{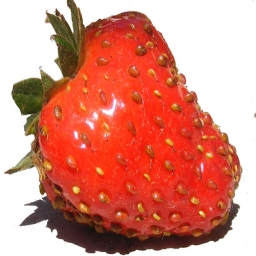} & \includegraphics[width=3cm, height=3cm]{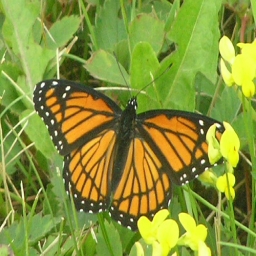} & \includegraphics[width=3cm, height=3cm]{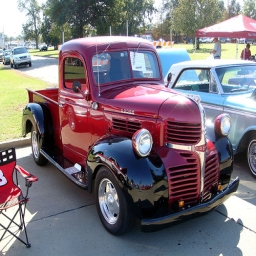} & \includegraphics[width=3cm, height=3cm]{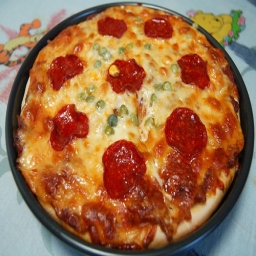} & \includegraphics[width=3cm, height=3cm]{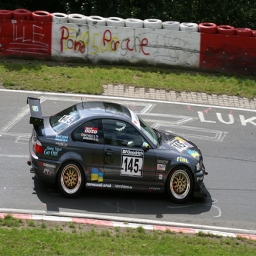} \\ 
\raisebox{1.5cm}{\huge 12 \hspace{6pt}} & \includegraphics[width=3cm, height=3cm]{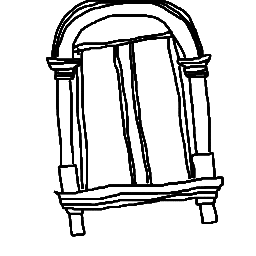} & \includegraphics[width=3cm, height=3cm, cfbox=ForestGreen 4pt 4pt]{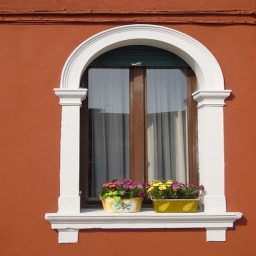} & \includegraphics[width=3cm, height=3cm]{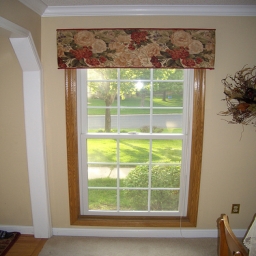} & \includegraphics[width=3cm, height=3cm]{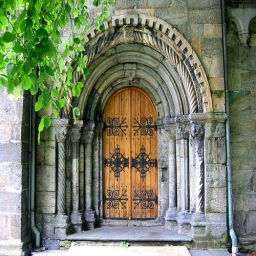} & \includegraphics[width=3cm, height=3cm]{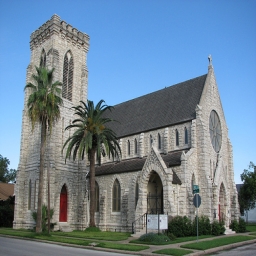} & \includegraphics[width=3cm, height=3cm]{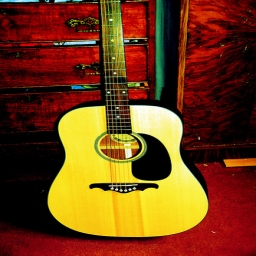} & \includegraphics[width=3cm, height=3cm]{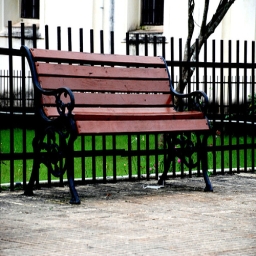} & \includegraphics[width=3cm, height=3cm]{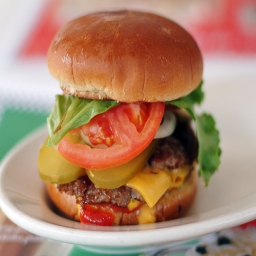} & \includegraphics[width=3cm, height=3cm]{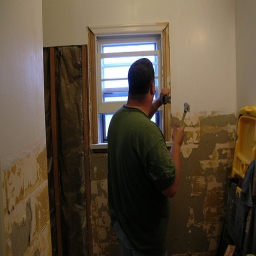} & \includegraphics[width=3cm, height=3cm]{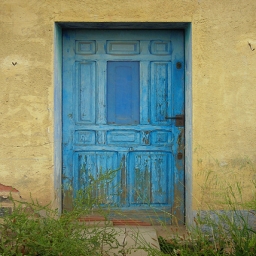} & \includegraphics[width=3cm, height=3cm]{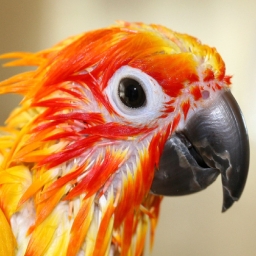}
\end{tabular}}
\end{center}
\caption{Qualitative fine-grained SBIR results on Sketchy dataset.}
\label{fig:sketchy_long}
\label{fig:qual_results_retrieval_results_app/Sketchy/}
\end{figure}

\begin{figure}[!ht]
\begin{center}
\resizebox{\textwidth}{!}{
\begin{tabular}{@{}c@{}c@{}c@{}c@{}c@{}c@{}c@{}c@{}c@{}c@{}c@{}c}
\raisebox{1.5cm}{\huge 1 \hspace{6pt}} & \includegraphics[width=3cm, height=3cm]{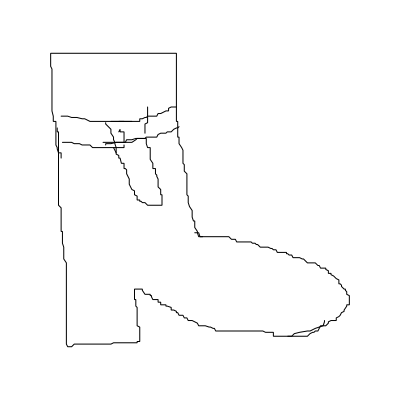} & \includegraphics[width=3cm, height=3cm, cfbox=ForestGreen 2pt 2pt]{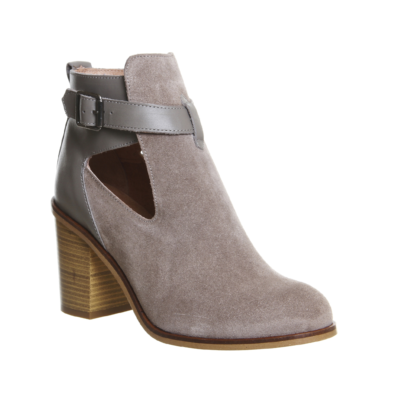} & \includegraphics[width=3cm, height=3cm]{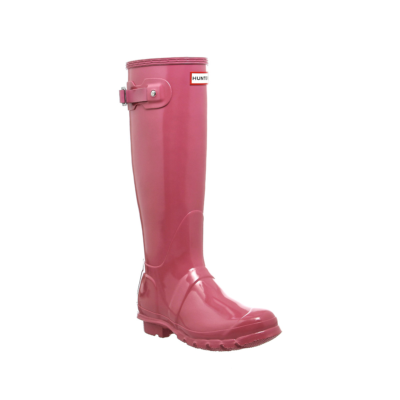} & \includegraphics[width=3cm, height=3cm]{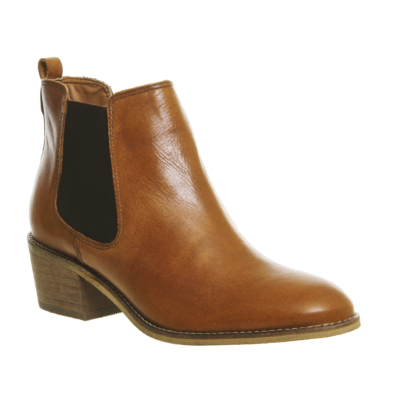} & \includegraphics[width=3cm, height=3cm]{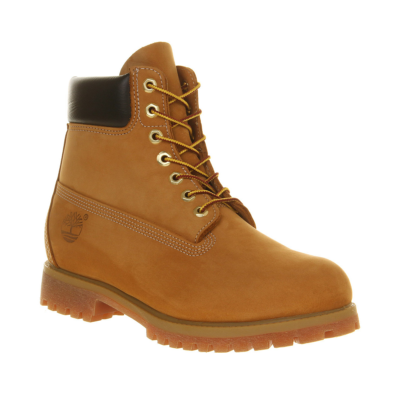} & \includegraphics[width=3cm, height=3cm]{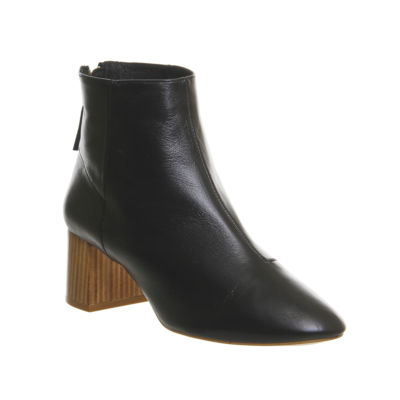} & \includegraphics[width=3cm, height=3cm]{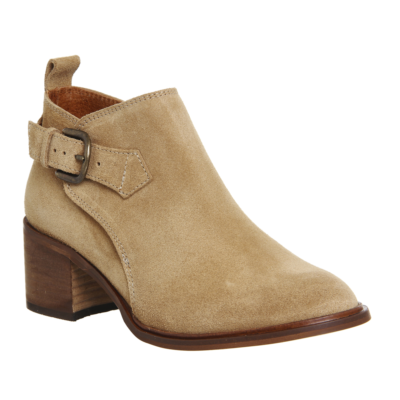} & \includegraphics[width=3cm, height=3cm]{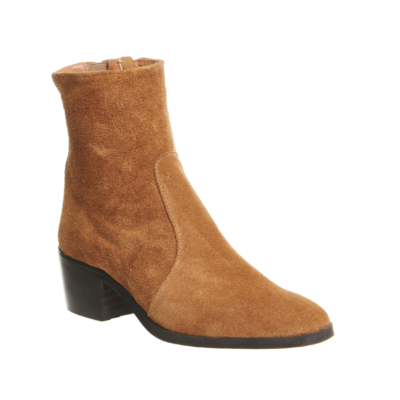} & \includegraphics[width=3cm, height=3cm]{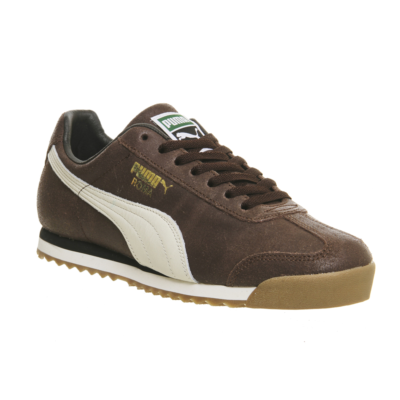} & \includegraphics[width=3cm, height=3cm]{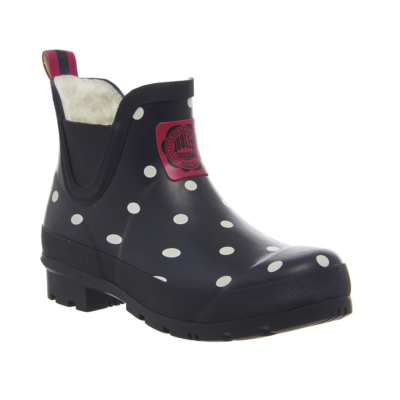} & \includegraphics[width=3cm, height=3cm]{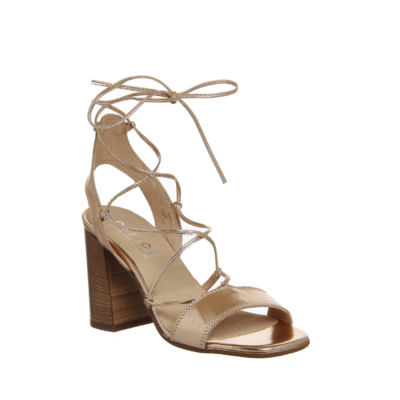} \\ 
\raisebox{1.5cm}{\huge 2 \hspace{6pt}} & \includegraphics[width=3cm, height=3cm]{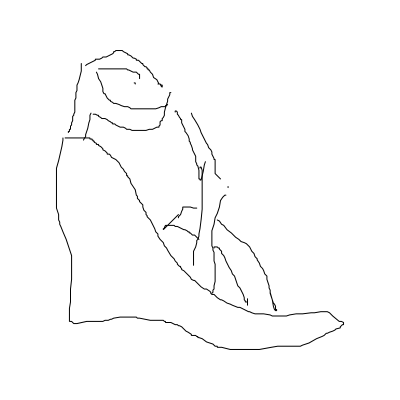} & \includegraphics[width=3cm, height=3cm, cfbox=ForestGreen 2pt 2pt]{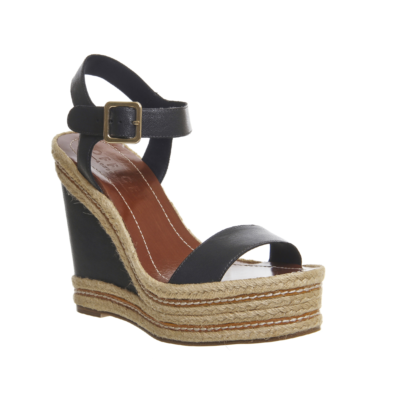} & \includegraphics[width=3cm, height=3cm]{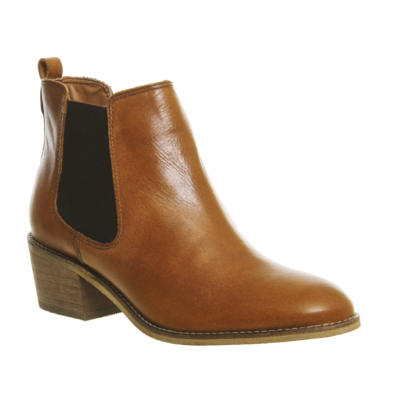} & \includegraphics[width=3cm, height=3cm]{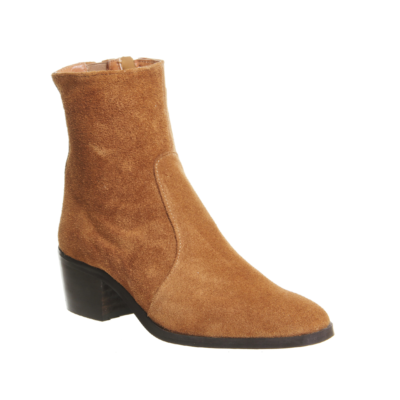} & \includegraphics[width=3cm, height=3cm]{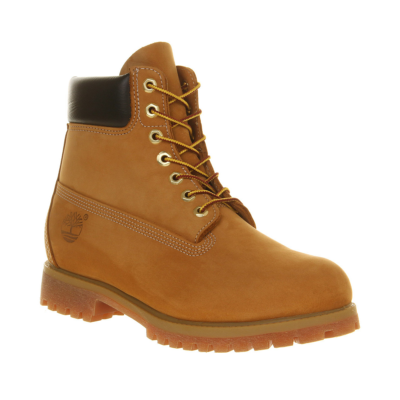} & \includegraphics[width=3cm, height=3cm]{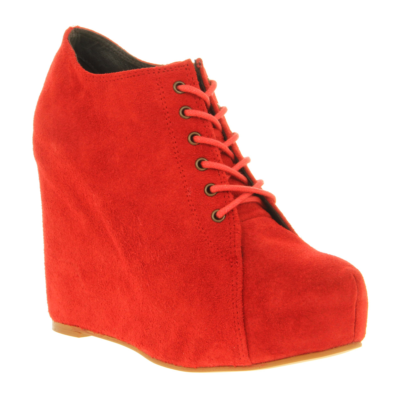} & \includegraphics[width=3cm, height=3cm]{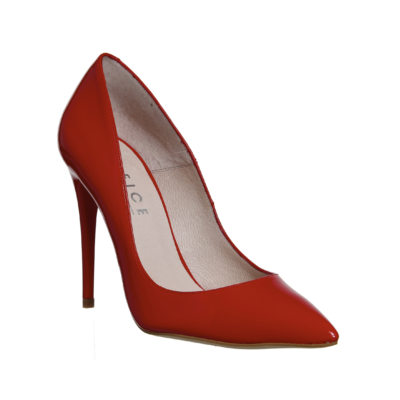} & \includegraphics[width=3cm, height=3cm]{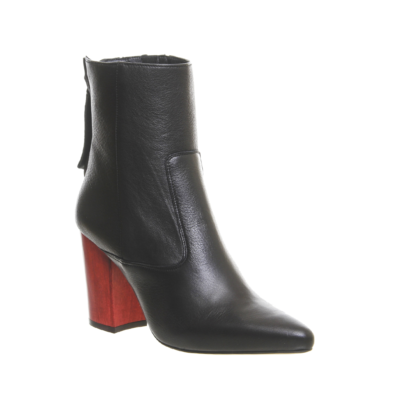} & \includegraphics[width=3cm, height=3cm]{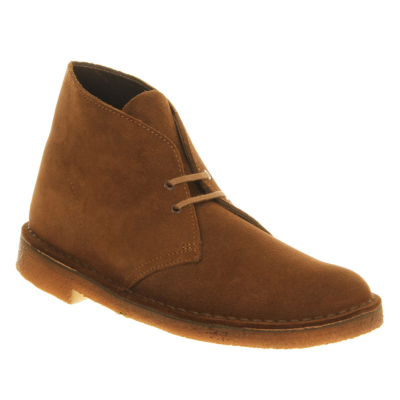} & \includegraphics[width=3cm, height=3cm]{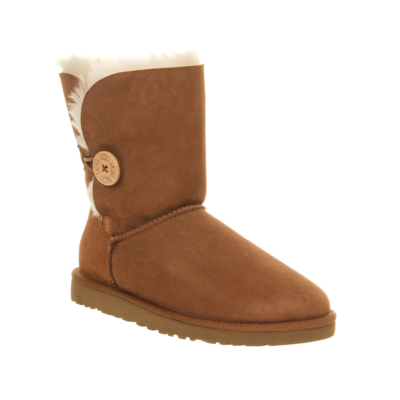} & \includegraphics[width=3cm, height=3cm]{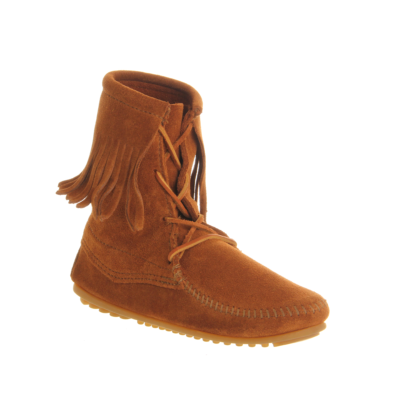} \\ 
\raisebox{1.5cm}{\huge 3 \hspace{6pt}} & \includegraphics[width=3cm, height=3cm]{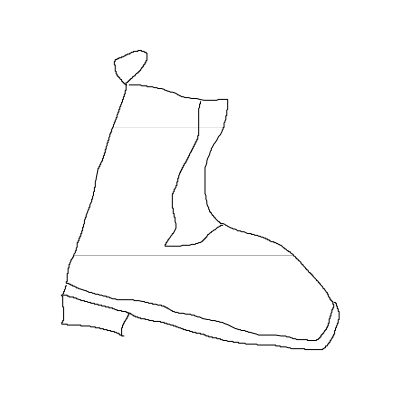} & \includegraphics[width=3cm, height=3cm, cfbox=ForestGreen 2pt 2pt]{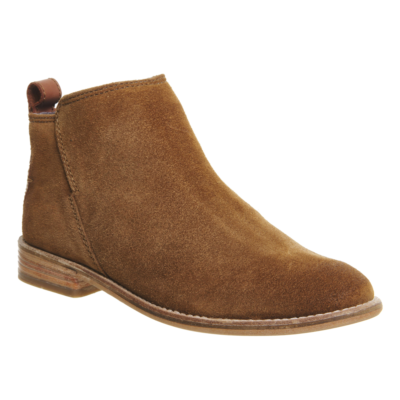} & \includegraphics[width=3cm, height=3cm]{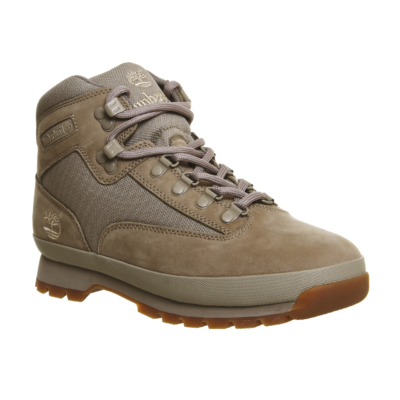} & \includegraphics[width=3cm, height=3cm]{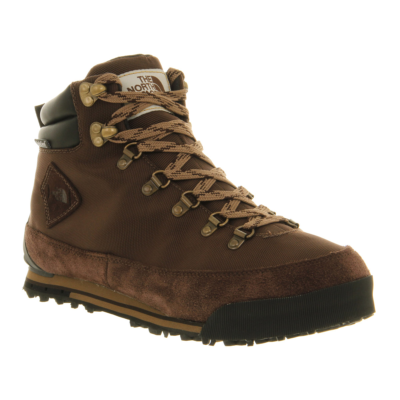} & \includegraphics[width=3cm, height=3cm]{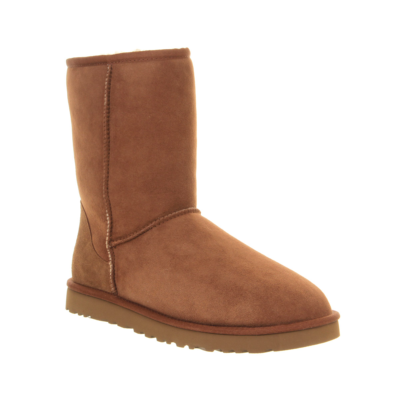} & \includegraphics[width=3cm, height=3cm]{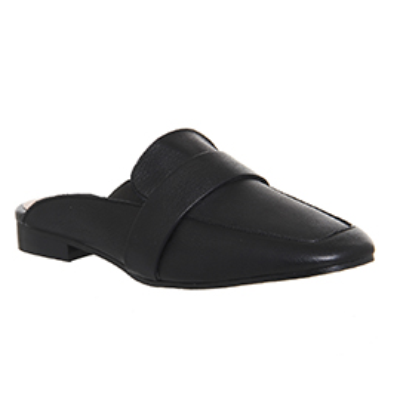} & \includegraphics[width=3cm, height=3cm]{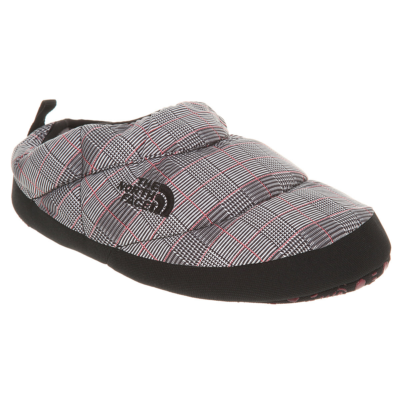} & \includegraphics[width=3cm, height=3cm]{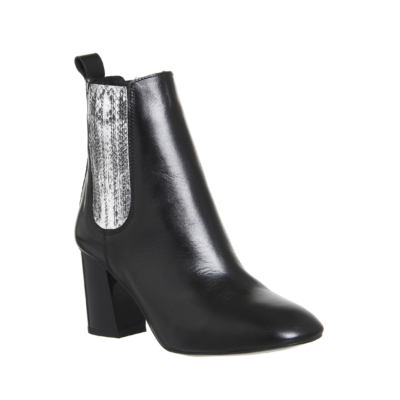} & \includegraphics[width=3cm, height=3cm]{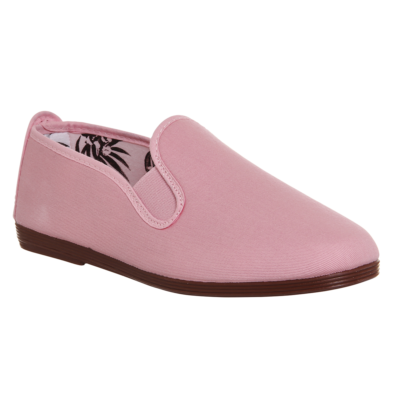} & \includegraphics[width=3cm, height=3cm]{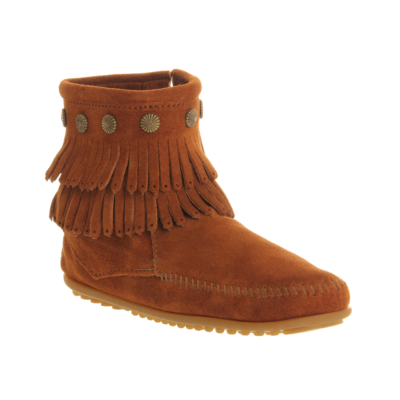} & \includegraphics[width=3cm, height=3cm]{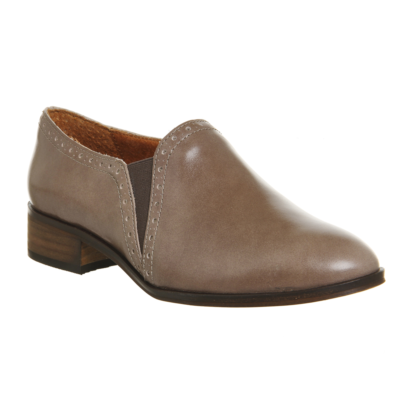} \\ 
\raisebox{1.5cm}{\huge 4 \hspace{6pt}} & \includegraphics[width=3cm, height=3cm]{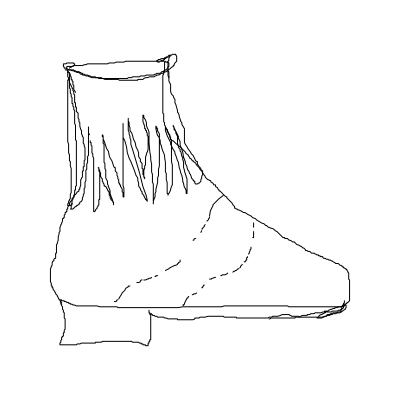} & \includegraphics[width=3cm, height=3cm]{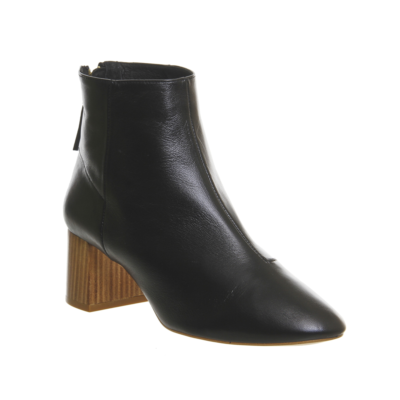} & \includegraphics[width=3cm, height=3cm, cfbox=ForestGreen 2pt 2pt]{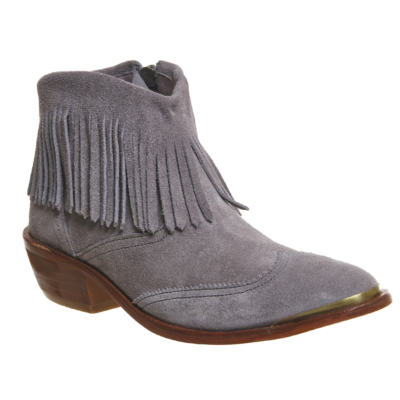} & \includegraphics[width=3cm, height=3cm]{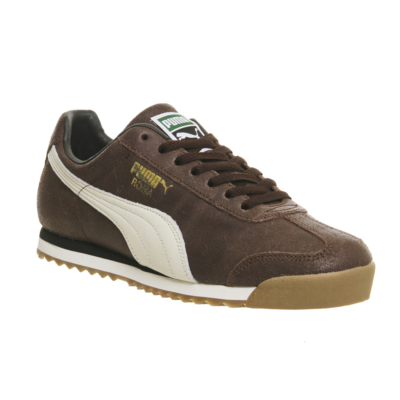} & \includegraphics[width=3cm, height=3cm]{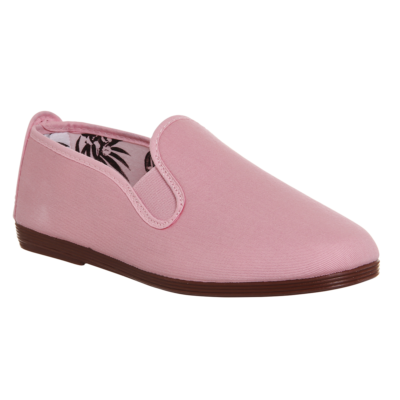} & \includegraphics[width=3cm, height=3cm]{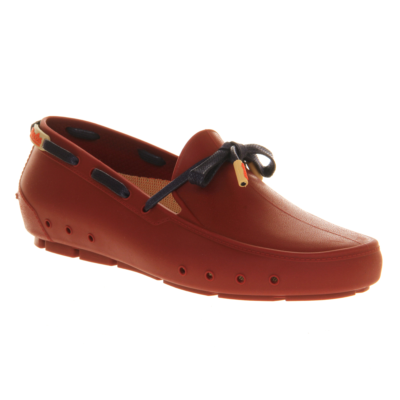} & \includegraphics[width=3cm, height=3cm]{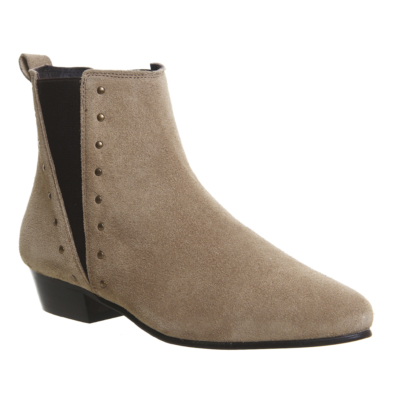} & \includegraphics[width=3cm, height=3cm]{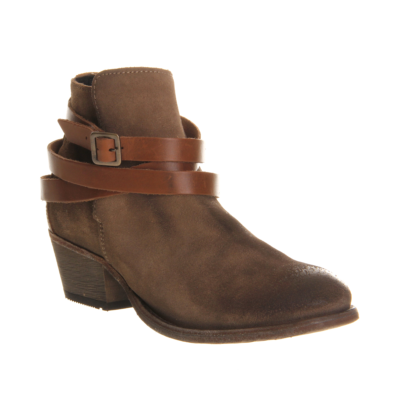} & \includegraphics[width=3cm, height=3cm]{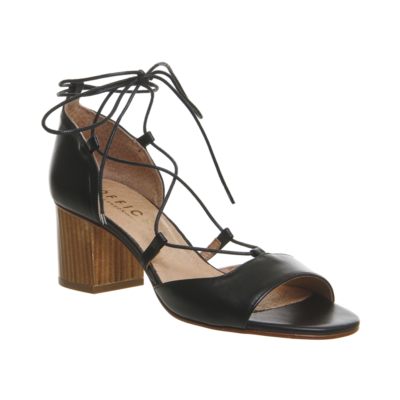} & \includegraphics[width=3cm, height=3cm]{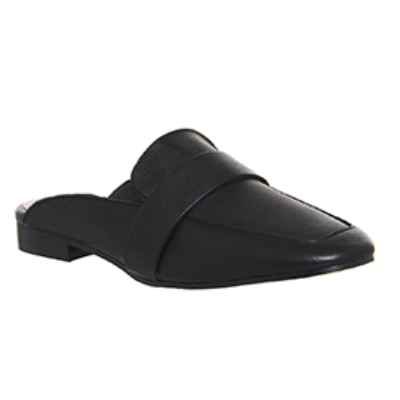} & \includegraphics[width=3cm, height=3cm]{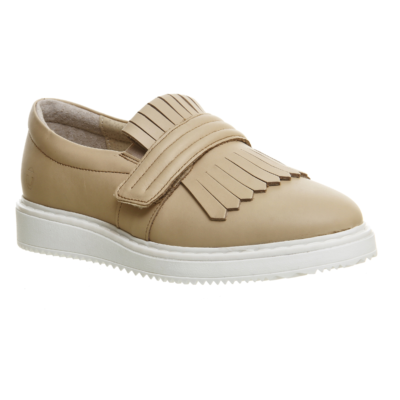} \\ 
\raisebox{1.5cm}{\huge 5 \hspace{6pt}} & \includegraphics[width=3cm, height=3cm]{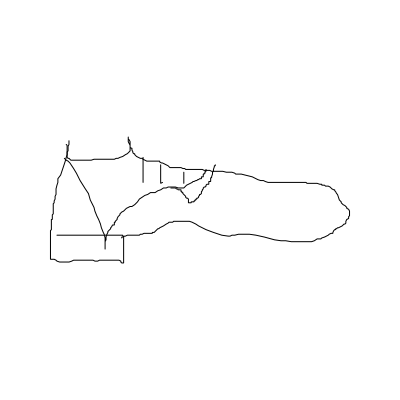} & \includegraphics[width=3cm, height=3cm, cfbox=ForestGreen 2pt 2pt]{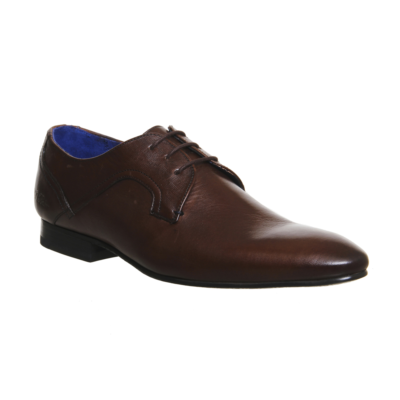} & \includegraphics[width=3cm, height=3cm]{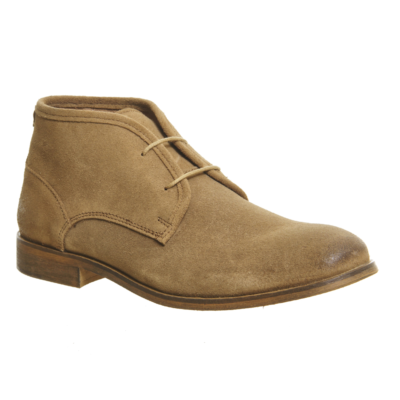} & \includegraphics[width=3cm, height=3cm]{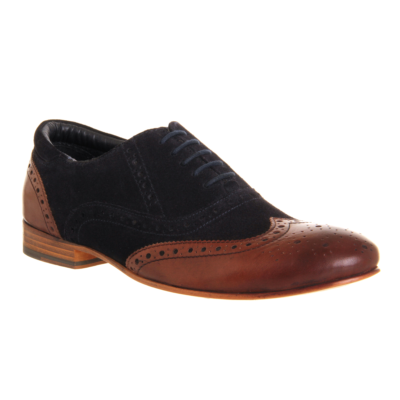} & \includegraphics[width=3cm, height=3cm]{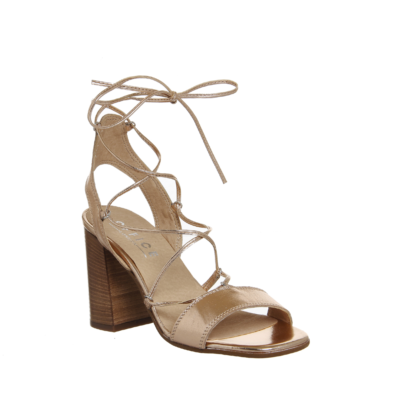} & \includegraphics[width=3cm, height=3cm]{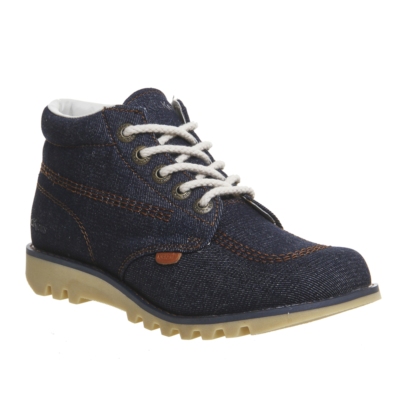} & \includegraphics[width=3cm, height=3cm]{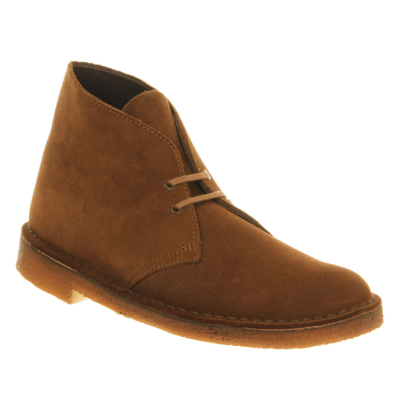} & \includegraphics[width=3cm, height=3cm]{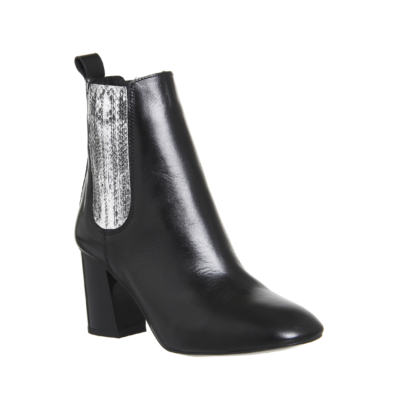} & \includegraphics[width=3cm, height=3cm]{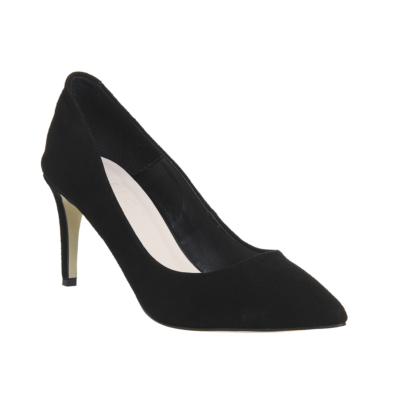} & \includegraphics[width=3cm, height=3cm]{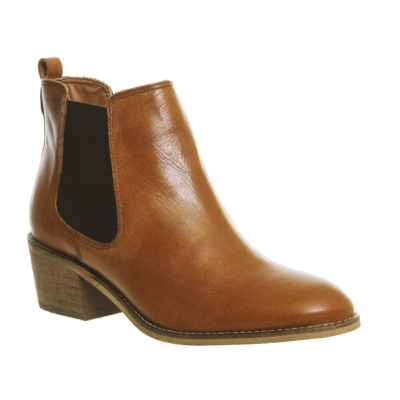} & \includegraphics[width=3cm, height=3cm]{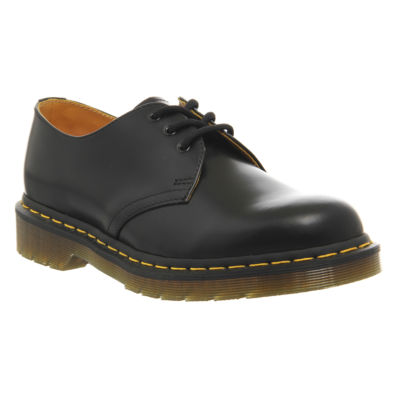} \\ 
\raisebox{1.5cm}{\huge 6 \hspace{6pt}} & \includegraphics[width=3cm, height=3cm]{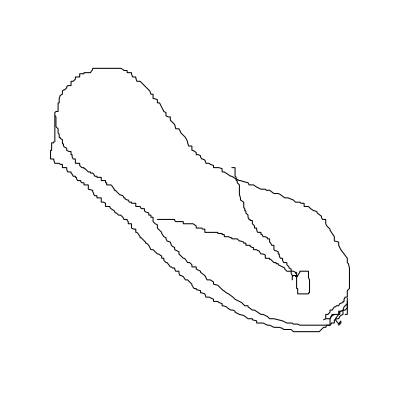} & \includegraphics[width=3cm, height=3cm, cfbox=ForestGreen 2pt 2pt]{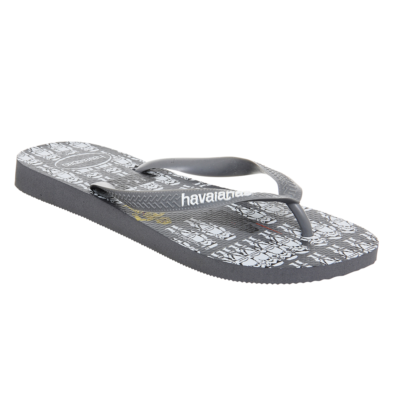} & \includegraphics[width=3cm, height=3cm]{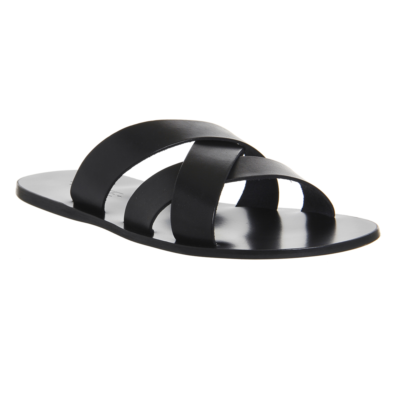} & \includegraphics[width=3cm, height=3cm]{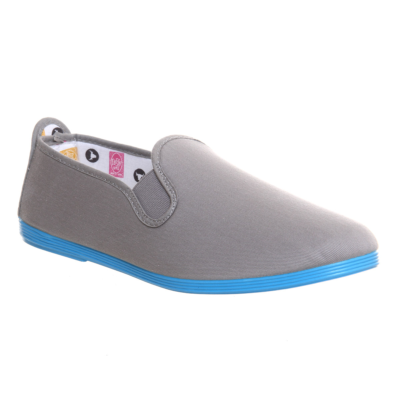} & \includegraphics[width=3cm, height=3cm]{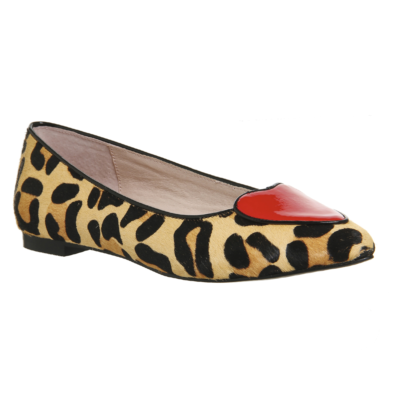} & \includegraphics[width=3cm, height=3cm]{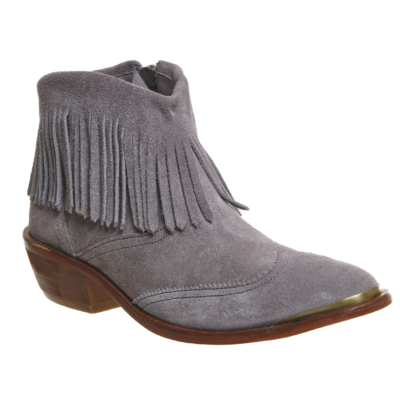} & \includegraphics[width=3cm, height=3cm]{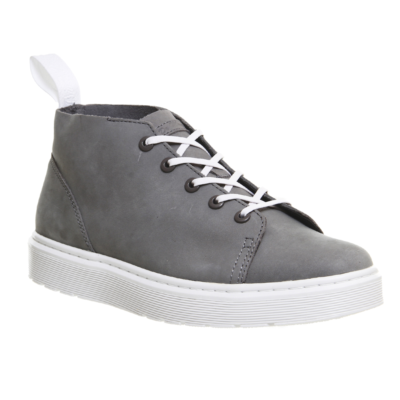} & \includegraphics[width=3cm, height=3cm]{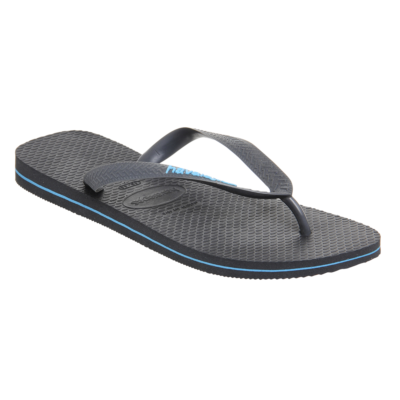} & \includegraphics[width=3cm, height=3cm]{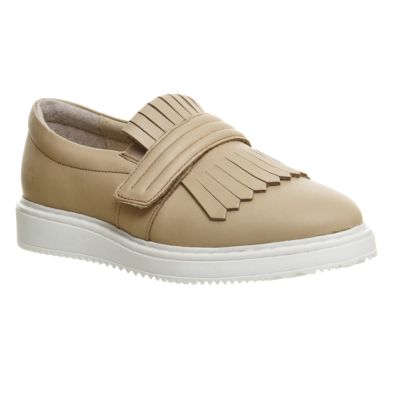} & \includegraphics[width=3cm, height=3cm]{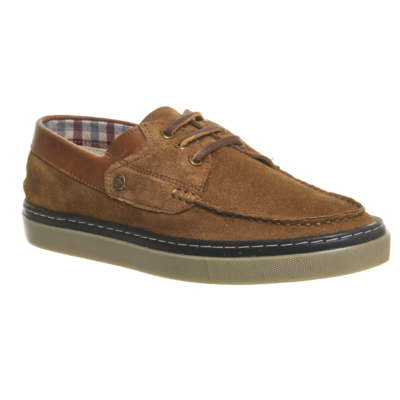} & \includegraphics[width=3cm, height=3cm]{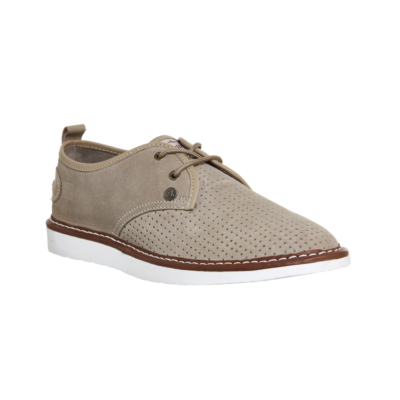} \\ 
\raisebox{1.5cm}{\huge 7 \hspace{6pt}} & \includegraphics[width=3cm, height=3cm]{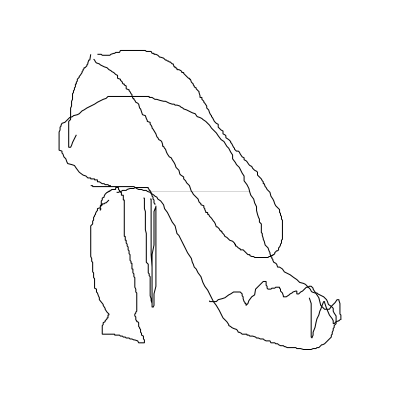} & \includegraphics[width=3cm, height=3cm, cfbox=ForestGreen 2pt 2pt]{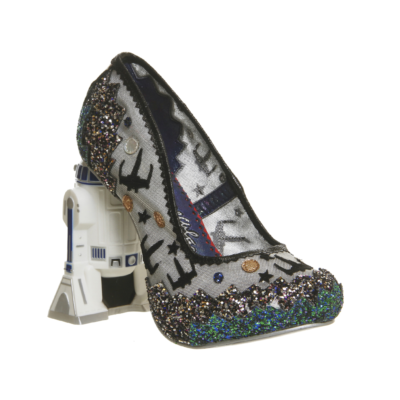} & \includegraphics[width=3cm, height=3cm]{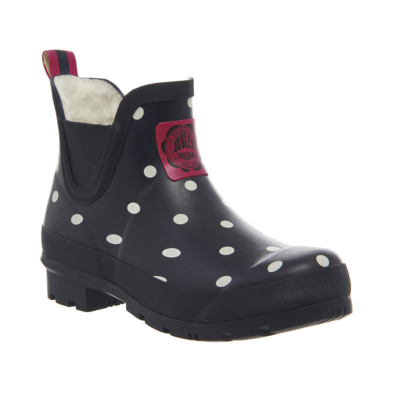} & \includegraphics[width=3cm, height=3cm]{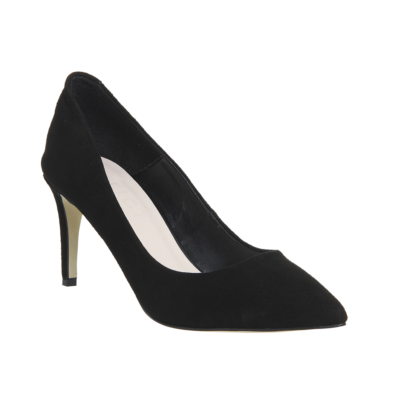} & \includegraphics[width=3cm, height=3cm]{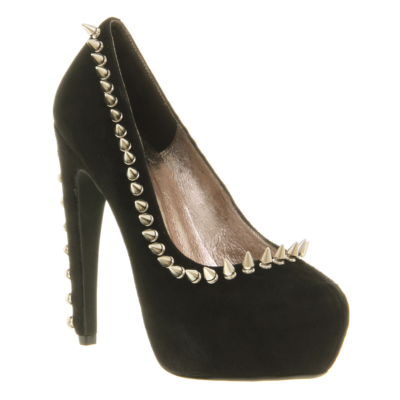} & \includegraphics[width=3cm, height=3cm]{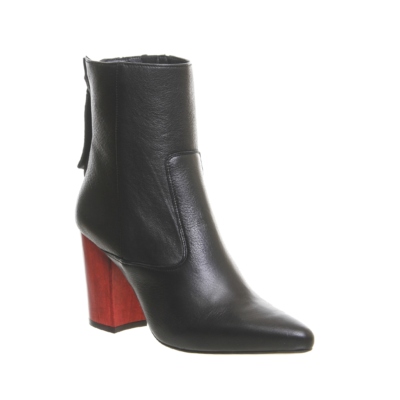} & \includegraphics[width=3cm, height=3cm]{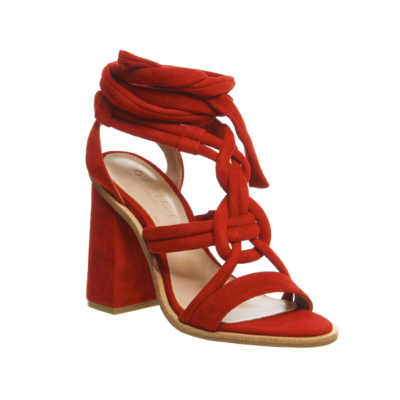} & \includegraphics[width=3cm, height=3cm]{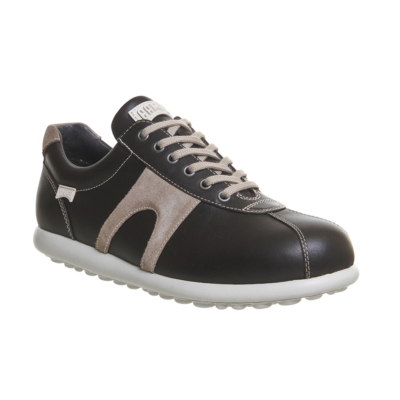} & \includegraphics[width=3cm, height=3cm]{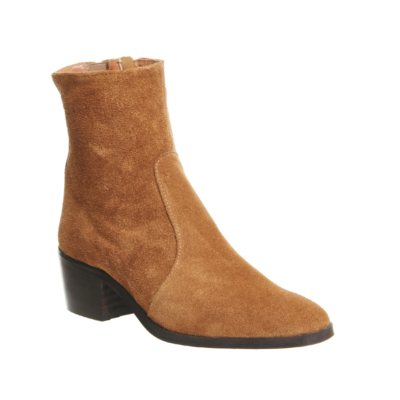} & \includegraphics[width=3cm, height=3cm]{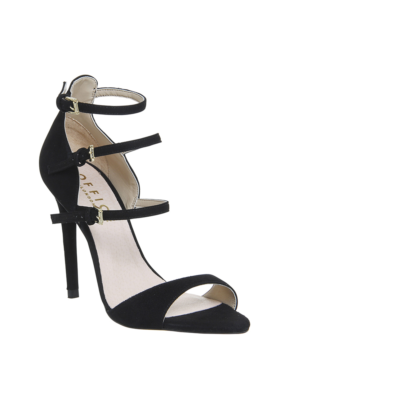} & \includegraphics[width=3cm, height=3cm]{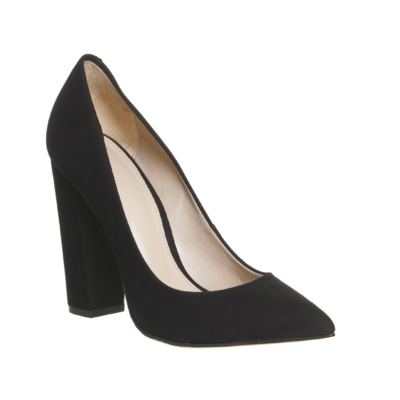} \\ 
\raisebox{1.5cm}{\huge 8 \hspace{6pt}} & \includegraphics[width=3cm, height=3cm]{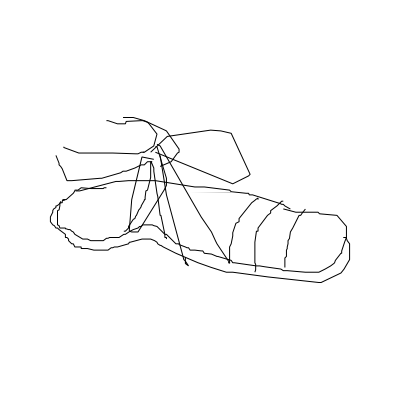} & \includegraphics[width=3cm, height=3cm, cfbox=ForestGreen 2pt 2pt]{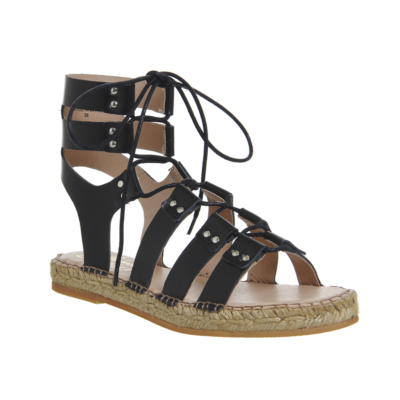} & \includegraphics[width=3cm, height=3cm]{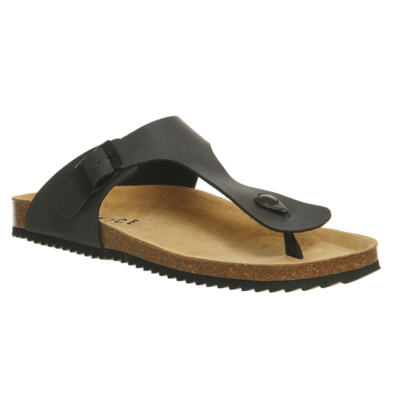} & \includegraphics[width=3cm, height=3cm]{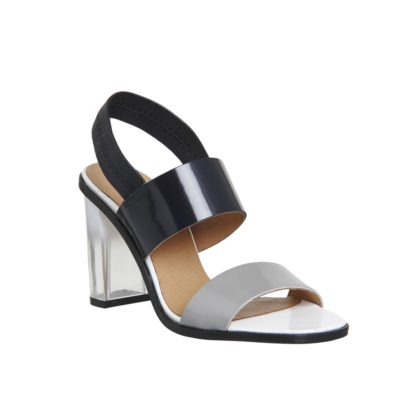} & \includegraphics[width=3cm, height=3cm]{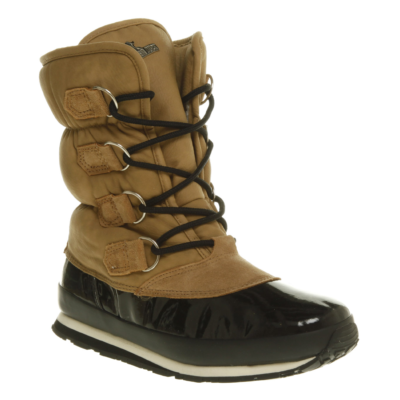} & \includegraphics[width=3cm, height=3cm]{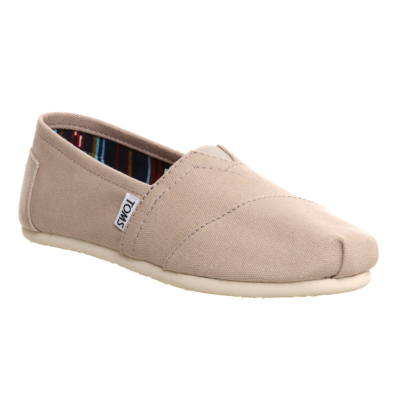} & \includegraphics[width=3cm, height=3cm]{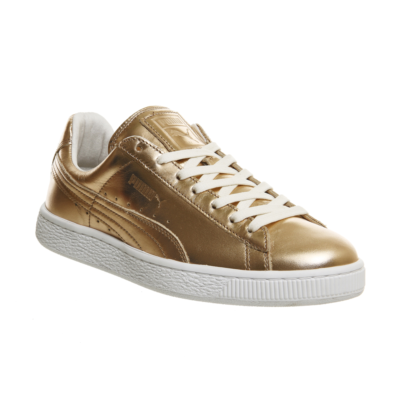} & \includegraphics[width=3cm, height=3cm]{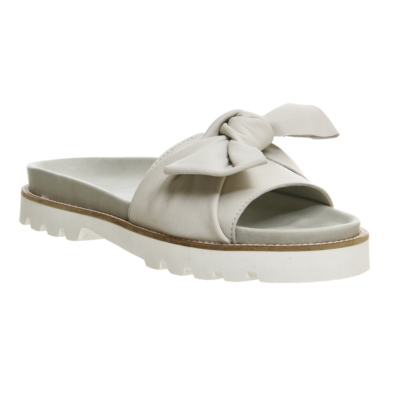} & \includegraphics[width=3cm, height=3cm]{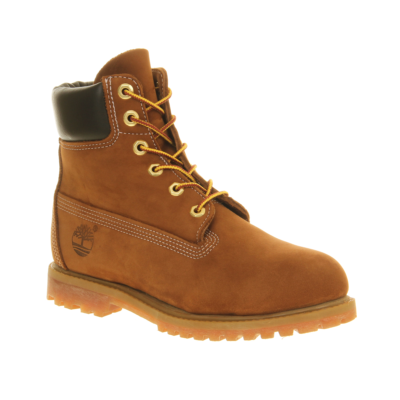} & \includegraphics[width=3cm, height=3cm]{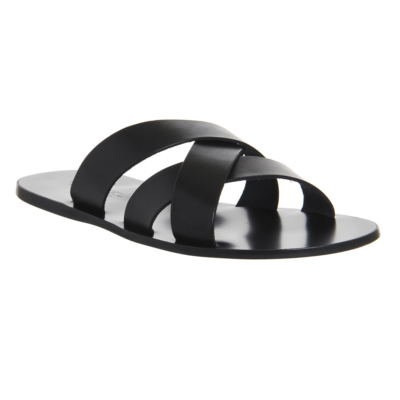} & \includegraphics[width=3cm, height=3cm]{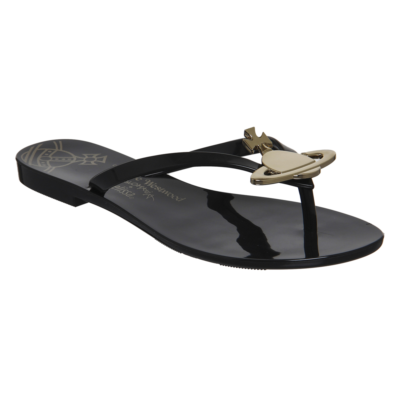} \\ 
\raisebox{1.5cm}{\huge 9 \hspace{6pt}} & \includegraphics[width=3cm, height=3cm]{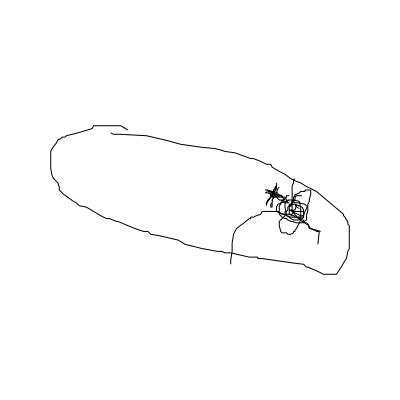} & \includegraphics[width=3cm, height=3cm]{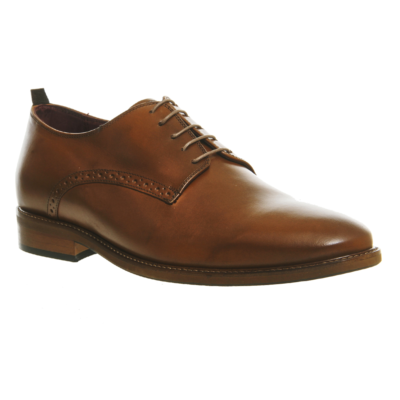} & \includegraphics[width=3cm, height=3cm]{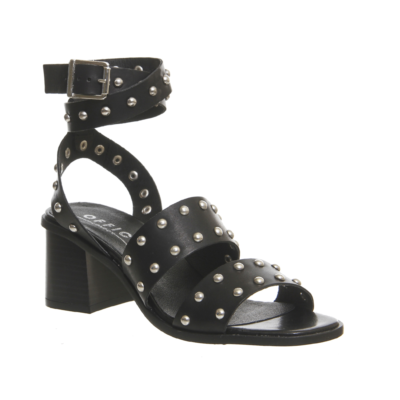} & \includegraphics[width=3cm, height=3cm]{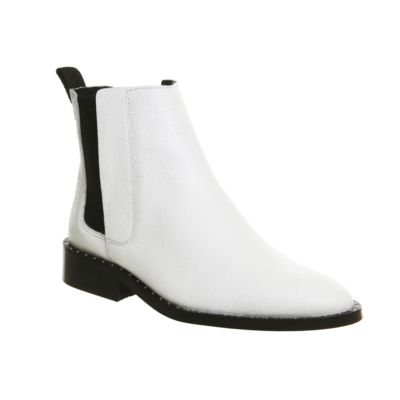} & \includegraphics[width=3cm, height=3cm]{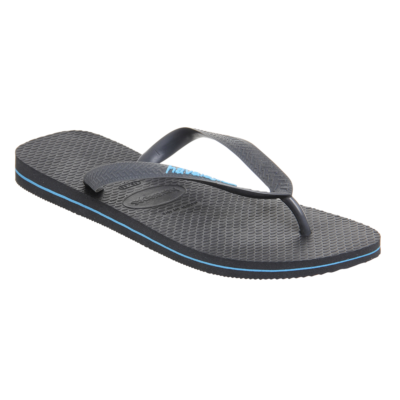} & \includegraphics[width=3cm, height=3cm]{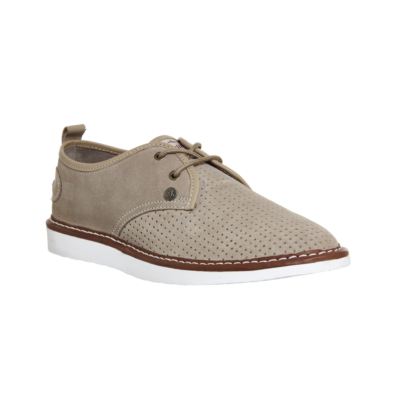} & \includegraphics[width=3cm, height=3cm]{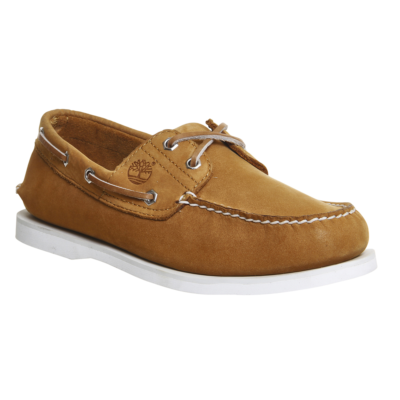} & \includegraphics[width=3cm, height=3cm]{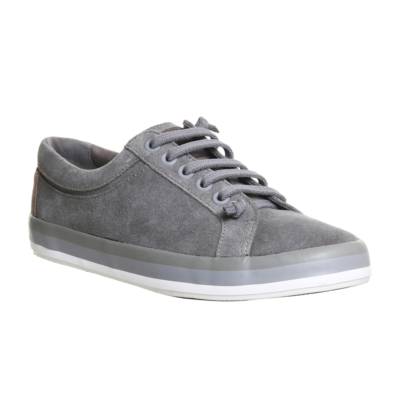} & \includegraphics[width=3cm, height=3cm]{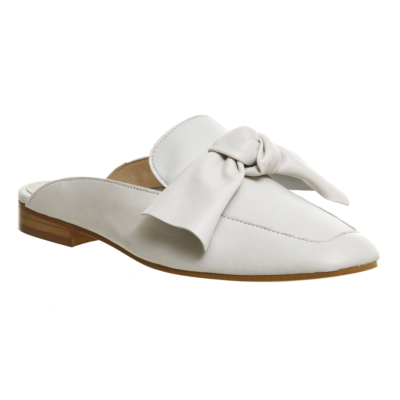} & \includegraphics[width=3cm, height=3cm, cfbox=ForestGreen 2pt 2pt]{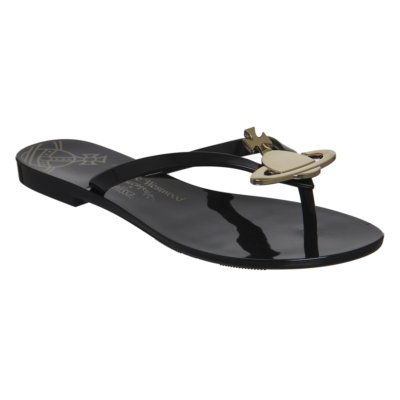} & \includegraphics[width=3cm, height=3cm]{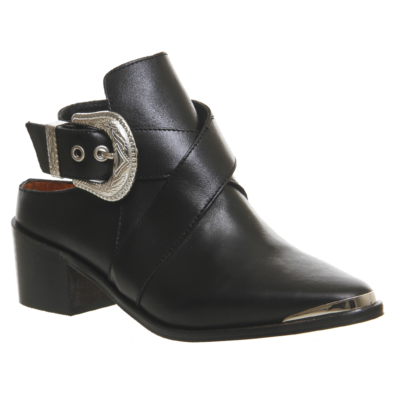} \\ 
\raisebox{1.5cm}{\huge 10 \hspace{6pt}} & \includegraphics[width=3cm, height=3cm]{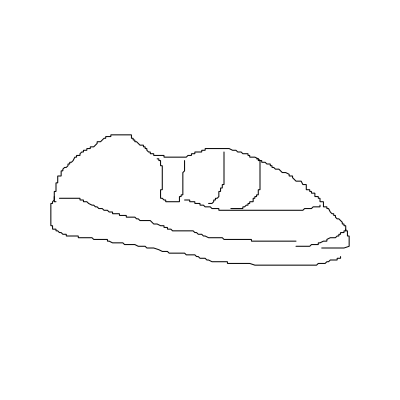} & \includegraphics[width=3cm, height=3cm, cfbox=ForestGreen 2pt 2pt]{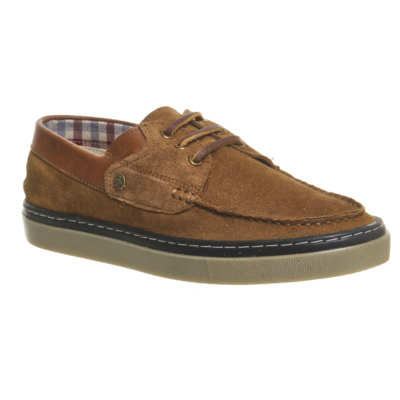} & \includegraphics[width=3cm, height=3cm]{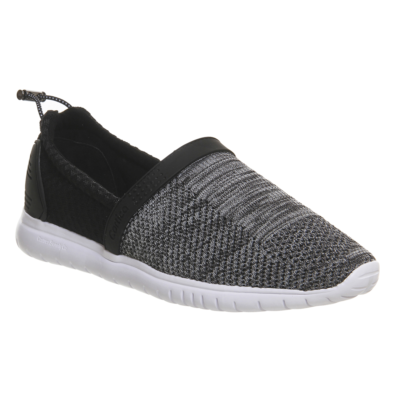} & \includegraphics[width=3cm, height=3cm]{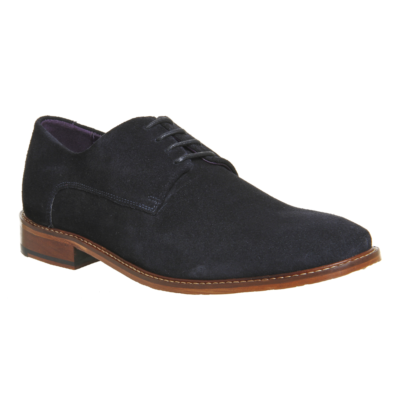} & \includegraphics[width=3cm, height=3cm]{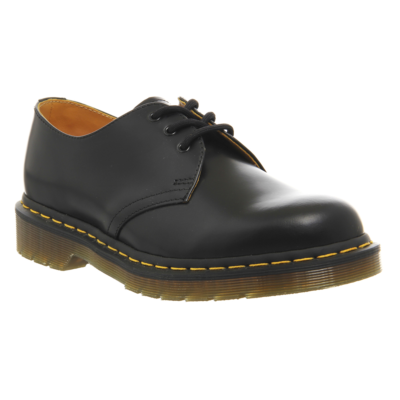} & \includegraphics[width=3cm, height=3cm]{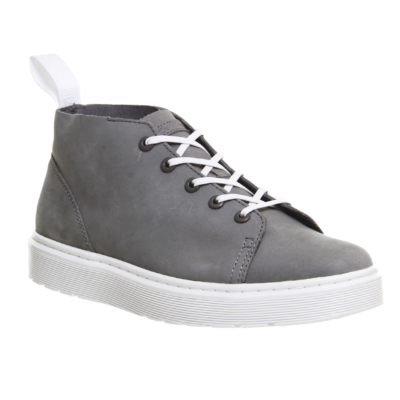} & \includegraphics[width=3cm, height=3cm]{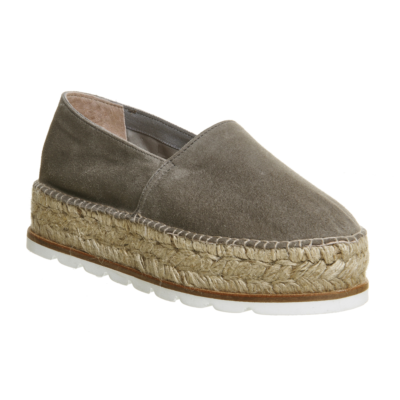} & \includegraphics[width=3cm, height=3cm]{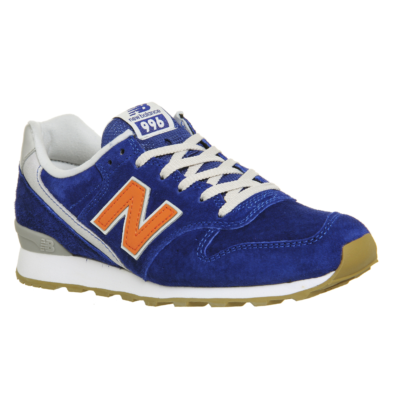} & \includegraphics[width=3cm, height=3cm]{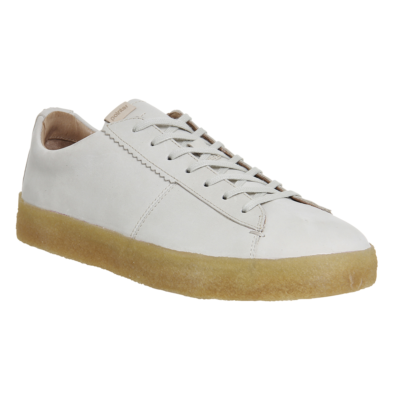} & \includegraphics[width=3cm, height=3cm]{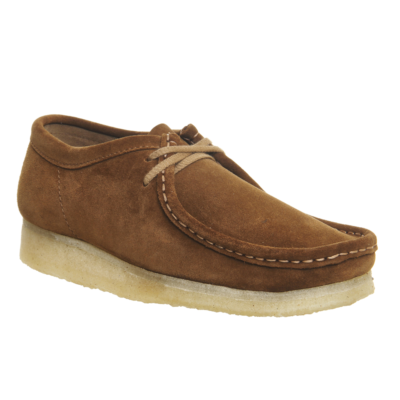} & \includegraphics[width=3cm, height=3cm]{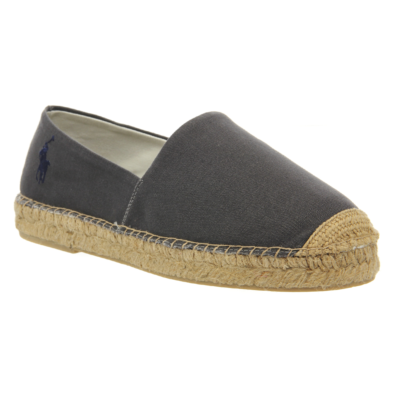} \\ 
\raisebox{1.5cm}{\huge 11 \hspace{6pt}} & \includegraphics[width=3cm, height=3cm]{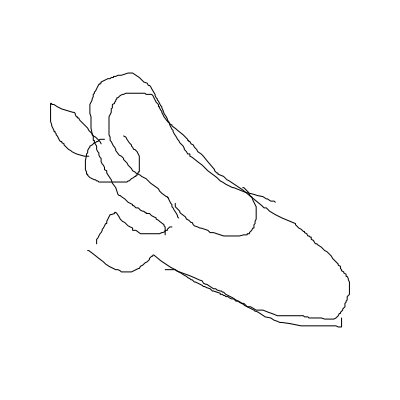} & \includegraphics[width=3cm, height=3cm, cfbox=ForestGreen 2pt 2pt]{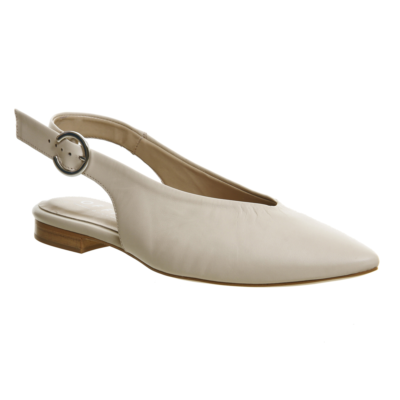} & \includegraphics[width=3cm, height=3cm]{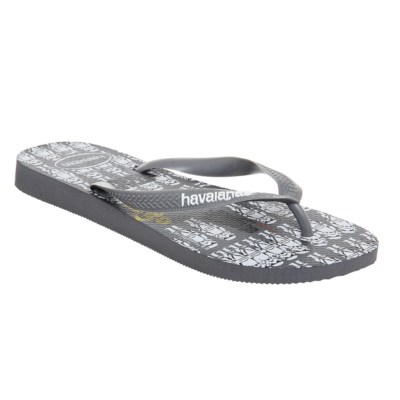} & \includegraphics[width=3cm, height=3cm]{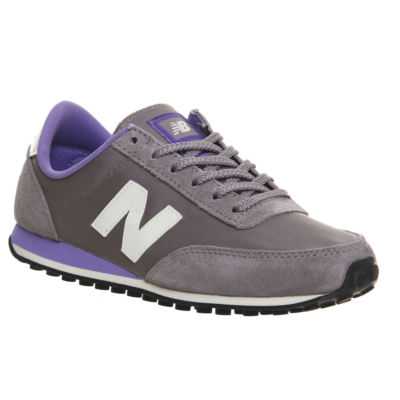} & \includegraphics[width=3cm, height=3cm]{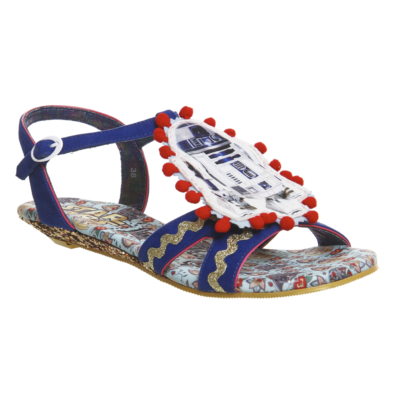} & \includegraphics[width=3cm, height=3cm]{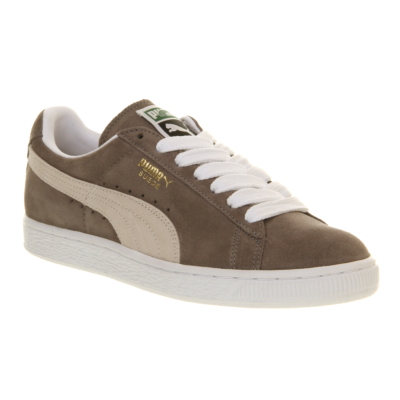} & \includegraphics[width=3cm, height=3cm]{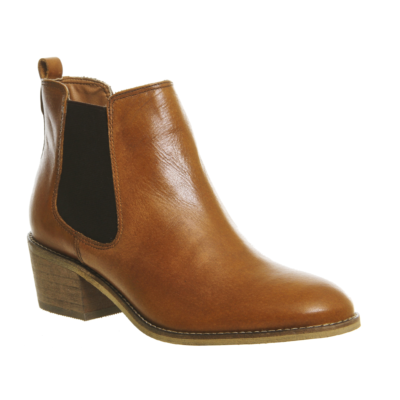} & \includegraphics[width=3cm, height=3cm]{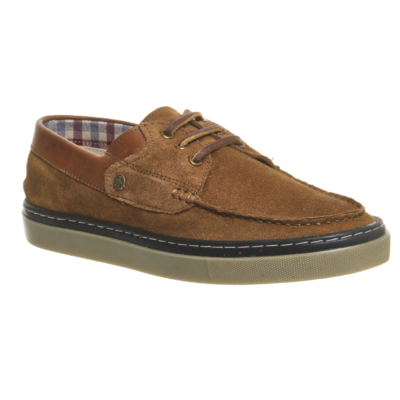} & \includegraphics[width=3cm, height=3cm]{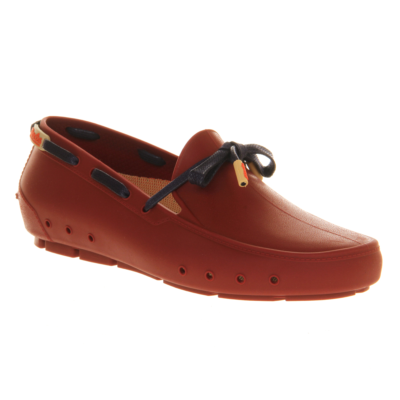} & \includegraphics[width=3cm, height=3cm]{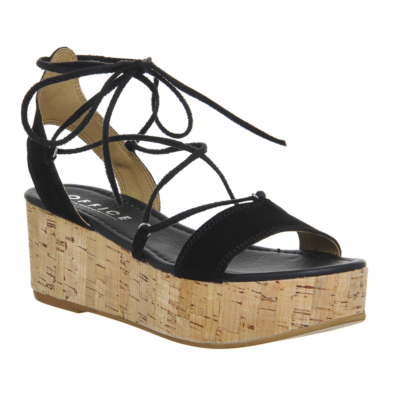} & \includegraphics[width=3cm, height=3cm]{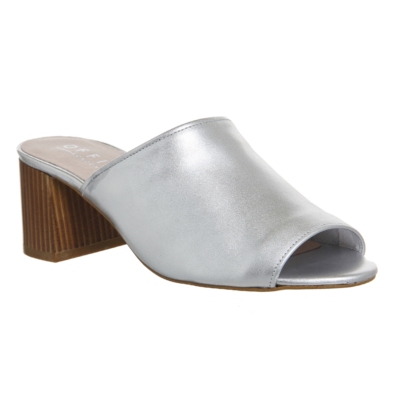} \\ 
\raisebox{1.5cm}{\huge 12 \hspace{6pt}} & \includegraphics[width=3cm, height=3cm]{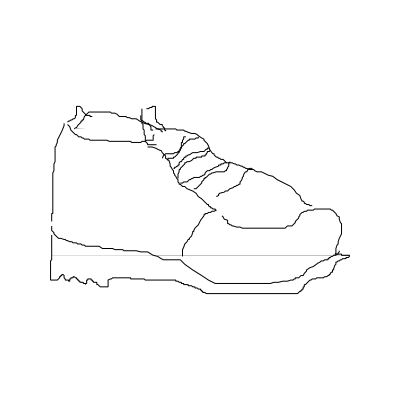} & \includegraphics[width=3cm, height=3cm, cfbox=ForestGreen 2pt 2pt]{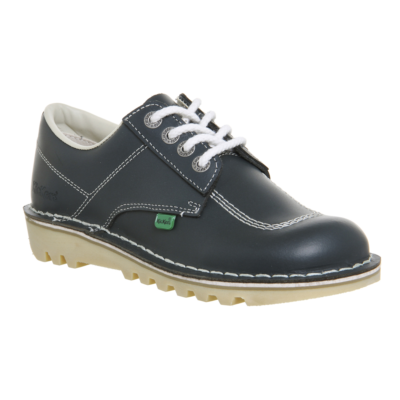} & \includegraphics[width=3cm, height=3cm]{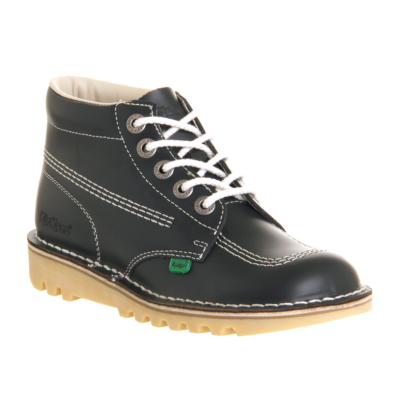} & \includegraphics[width=3cm, height=3cm]{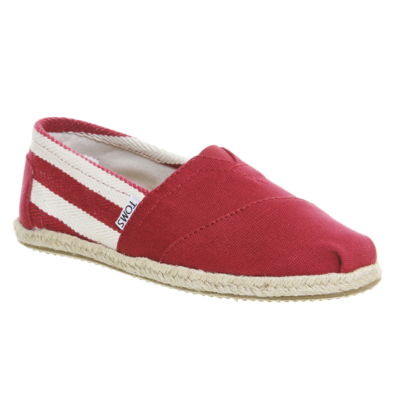} & \includegraphics[width=3cm, height=3cm]{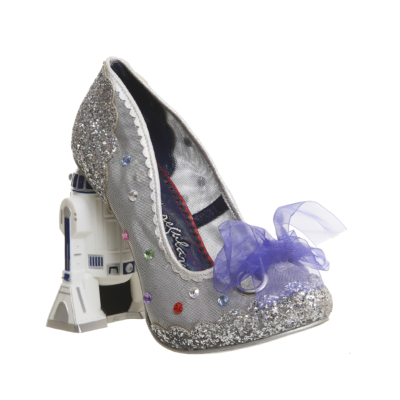} & \includegraphics[width=3cm, height=3cm]{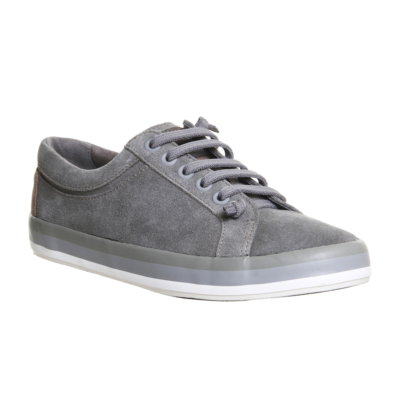} & \includegraphics[width=3cm, height=3cm]{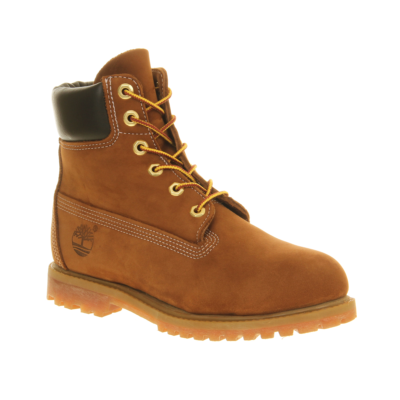} & \includegraphics[width=3cm, height=3cm]{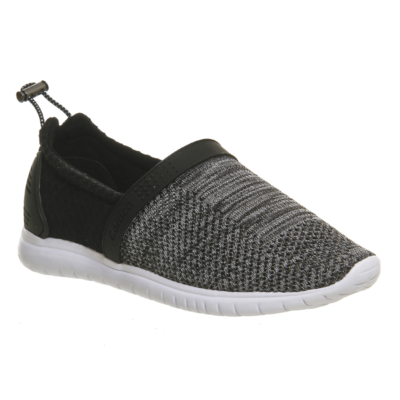} & \includegraphics[width=3cm, height=3cm]{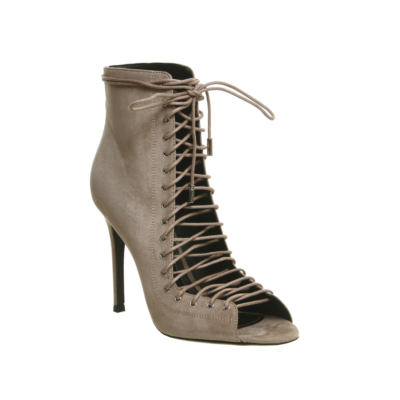} & \includegraphics[width=3cm, height=3cm]{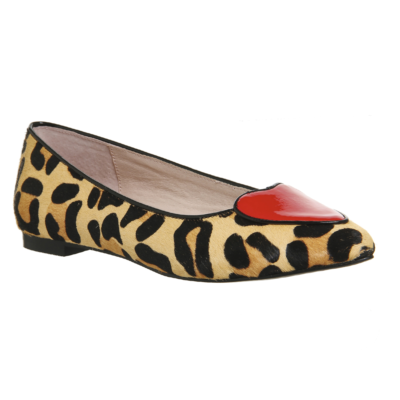} & \includegraphics[width=3cm, height=3cm]{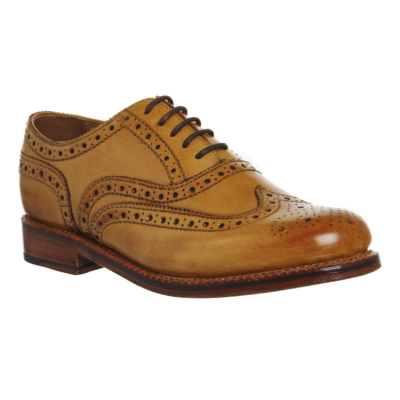}
\end{tabular}
}\end{center}
\caption{Qualitative fine-grained SBIR results on Shoe-V2 dataset.}
\label{fig:shoe_v2_long}
\label{fig:qual_results_retrieval_results_app/ShoeV2/}
\end{figure}

\begin{figure*}[!ht]
\begin{center}
\resizebox{\textwidth}{!}{
\begin{tabular}{@{}c@{}c@{}c@{}c@{}c@{}c@{}c@{}c@{}c@{}c@{}c@{}c}
\raisebox{1.5cm}{\huge 1 \hspace{6pt}} & \includegraphics[width=3cm, height=3cm]{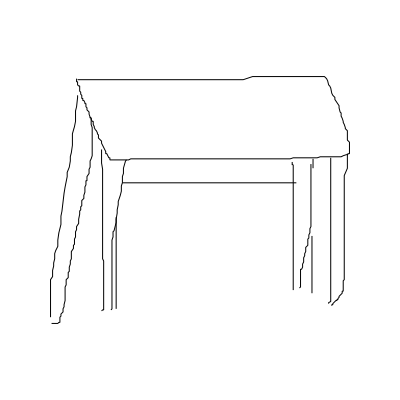} & \includegraphics[width=3cm, height=3cm, cfbox=ForestGreen 2pt 2pt]{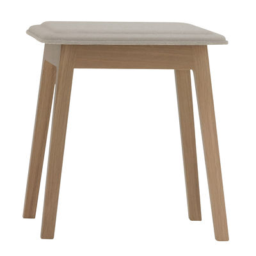} & \includegraphics[width=3cm, height=3cm]{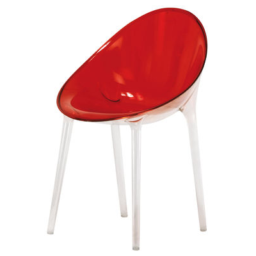} & \includegraphics[width=3cm, height=3cm]{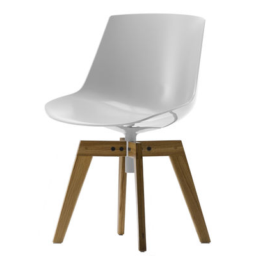} & \includegraphics[width=3cm, height=3cm]{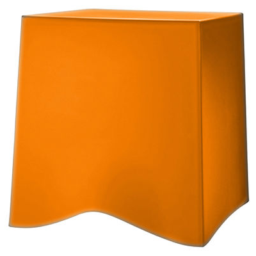} & \includegraphics[width=3cm, height=3cm]{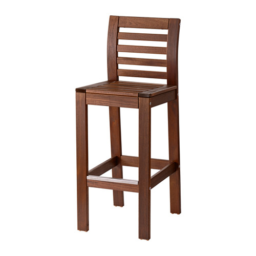} & \includegraphics[width=3cm, height=3cm]{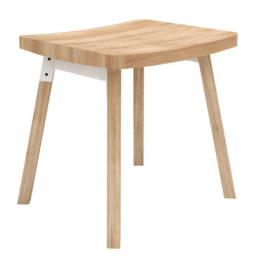} & \includegraphics[width=3cm, height=3cm]{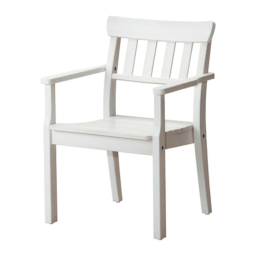} & \includegraphics[width=3cm, height=3cm]{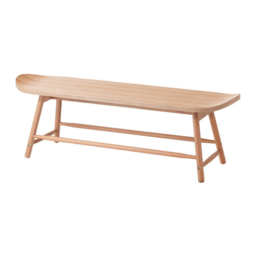} & \includegraphics[width=3cm, height=3cm]{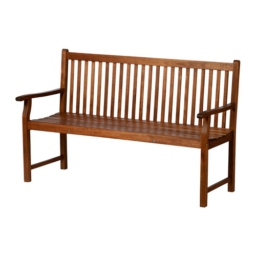} & \includegraphics[width=3cm, height=3cm]{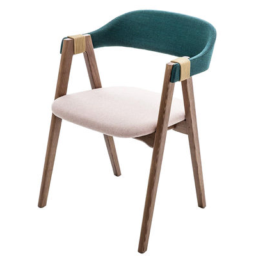} \\ 
\raisebox{1.5cm}{\huge 2 \hspace{6pt}} & \includegraphics[width=3cm, height=3cm]{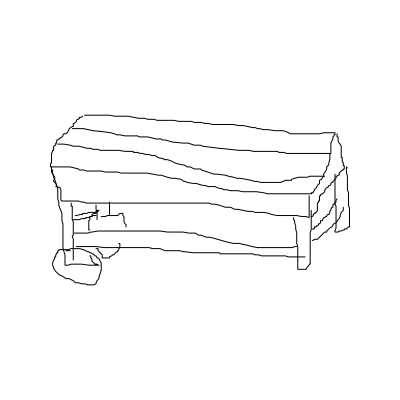} & \includegraphics[width=3cm, height=3cm, cfbox=ForestGreen 2pt 2pt]{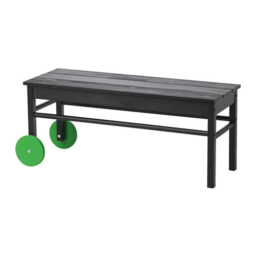} & \includegraphics[width=3cm, height=3cm]{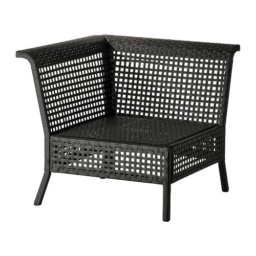} & \includegraphics[width=3cm, height=3cm]{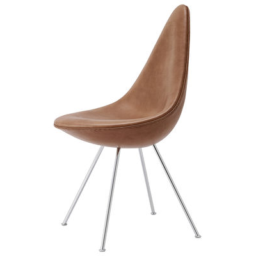} & \includegraphics[width=3cm, height=3cm]{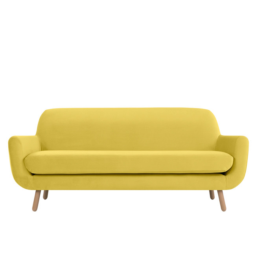} & \includegraphics[width=3cm, height=3cm]{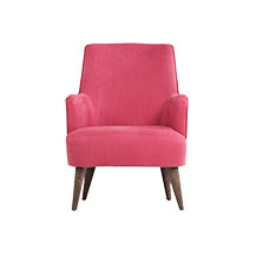} & \includegraphics[width=3cm, height=3cm]{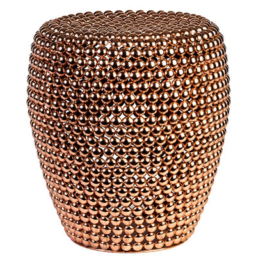} & \includegraphics[width=3cm, height=3cm]{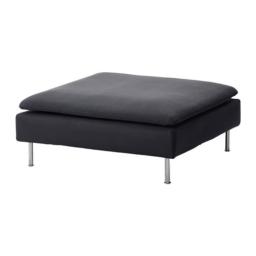} & \includegraphics[width=3cm, height=3cm]{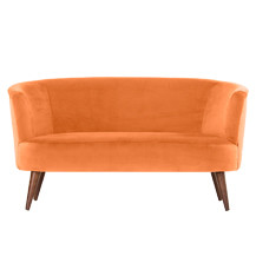} & \includegraphics[width=3cm, height=3cm]{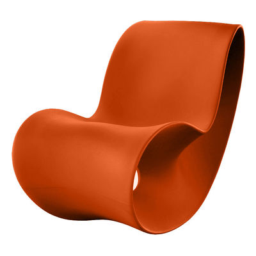} & \includegraphics[width=3cm, height=3cm]{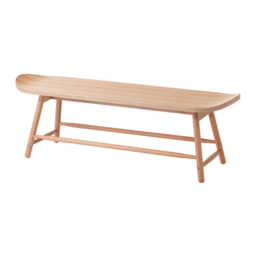} \\ 
\raisebox{1.5cm}{\huge 3 \hspace{6pt}} & \includegraphics[width=3cm, height=3cm]{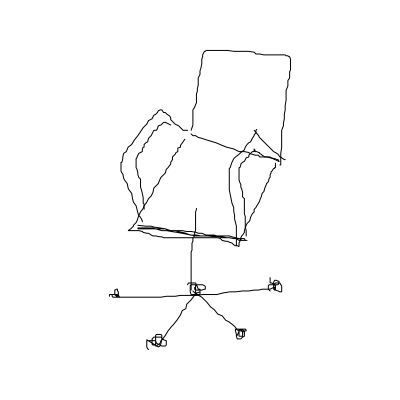} & \includegraphics[width=3cm, height=3cm, cfbox=ForestGreen 2pt 2pt]{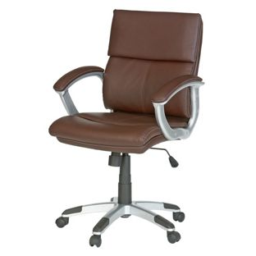} & \includegraphics[width=3cm, height=3cm]{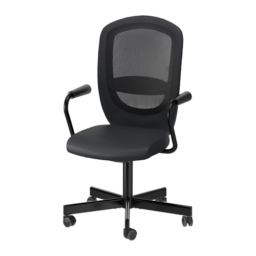} & \includegraphics[width=3cm, height=3cm]{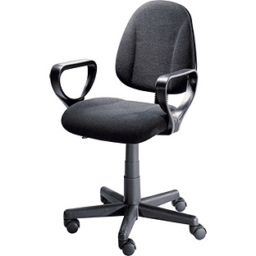} & \includegraphics[width=3cm, height=3cm]{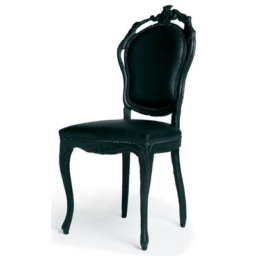} & \includegraphics[width=3cm, height=3cm]{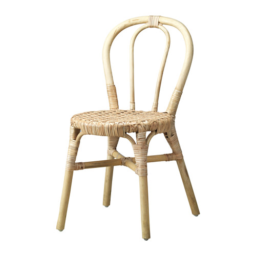} & \includegraphics[width=3cm, height=3cm]{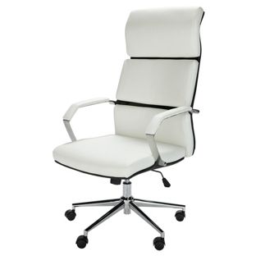} & \includegraphics[width=3cm, height=3cm]{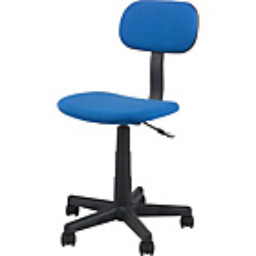} & \includegraphics[width=3cm, height=3cm]{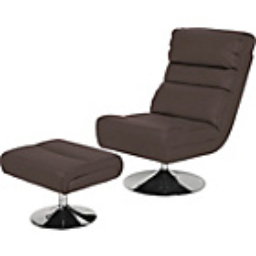} & \includegraphics[width=3cm, height=3cm]{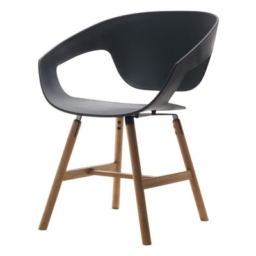} & \includegraphics[width=3cm, height=3cm]{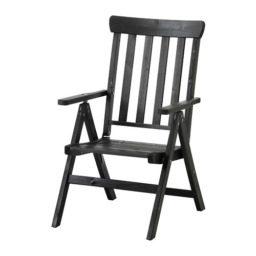} \\ 
\raisebox{1.5cm}{\huge 4 \hspace{6pt}} & \includegraphics[width=3cm, height=3cm]{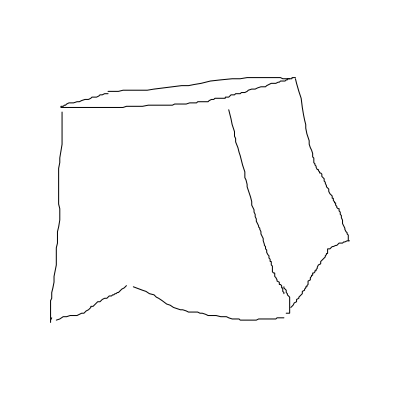} & \includegraphics[width=3cm, height=3cm, cfbox=ForestGreen 2pt 2pt]{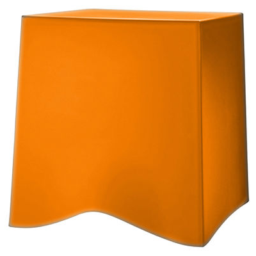} & \includegraphics[width=3cm, height=3cm]{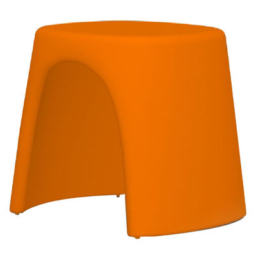} & \includegraphics[width=3cm, height=3cm]{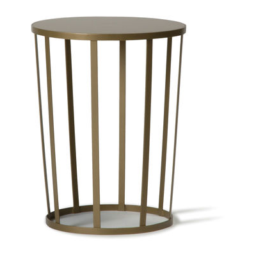} & \includegraphics[width=3cm, height=3cm]{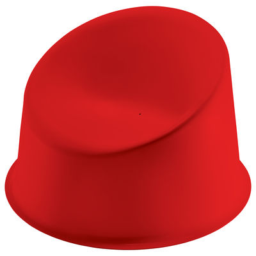} & \includegraphics[width=3cm, height=3cm]{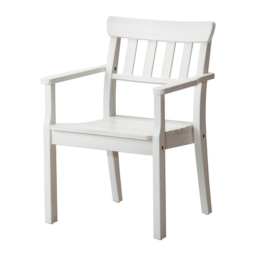} & \includegraphics[width=3cm, height=3cm]{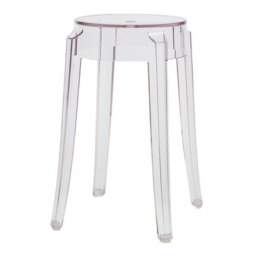} & \includegraphics[width=3cm, height=3cm]{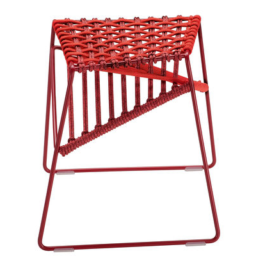} & \includegraphics[width=3cm, height=3cm]{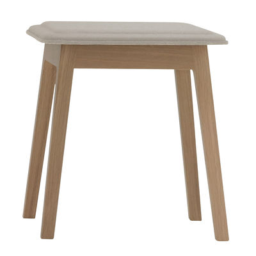} & \includegraphics[width=3cm, height=3cm]{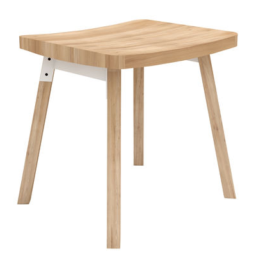} & \includegraphics[width=3cm, height=3cm]{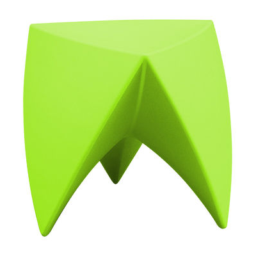} \\ 
\raisebox{1.5cm}{\huge 5 \hspace{6pt}} & \includegraphics[width=3cm, height=3cm]{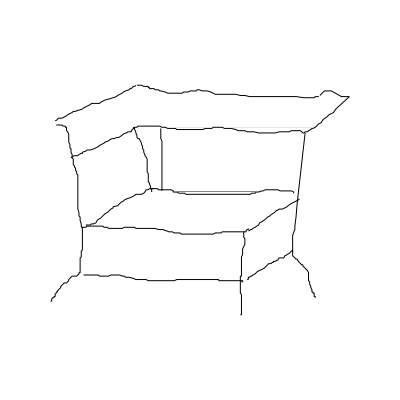} & \includegraphics[width=3cm, height=3cm, cfbox=ForestGreen 2pt 2pt]{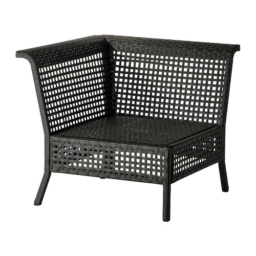} & \includegraphics[width=3cm, height=3cm]{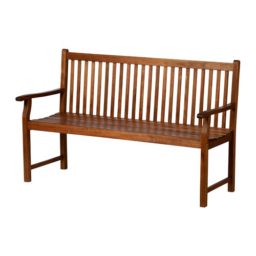} & \includegraphics[width=3cm, height=3cm]{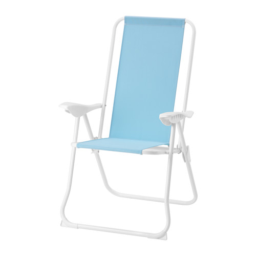} & \includegraphics[width=3cm, height=3cm]{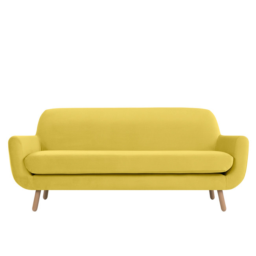} & \includegraphics[width=3cm, height=3cm]{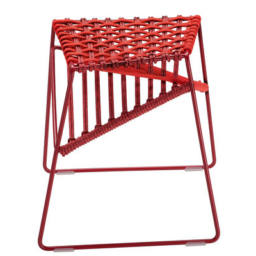} & \includegraphics[width=3cm, height=3cm]{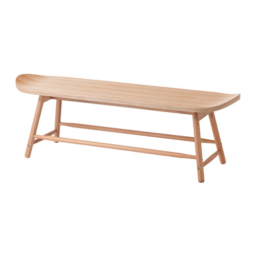} & \includegraphics[width=3cm, height=3cm]{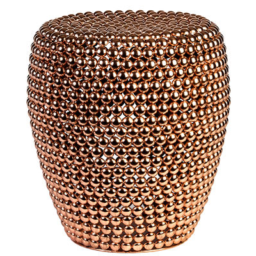} & \includegraphics[width=3cm, height=3cm]{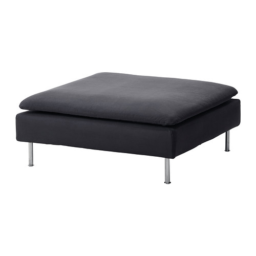} & \includegraphics[width=3cm, height=3cm]{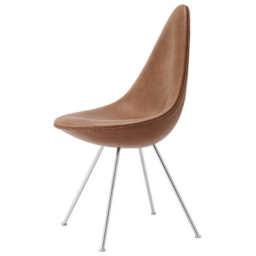} & \includegraphics[width=3cm, height=3cm]{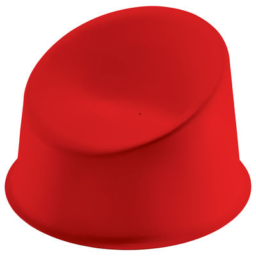} \\ 
\raisebox{1.5cm}{\huge 6 \hspace{6pt}} & \includegraphics[width=3cm, height=3cm]{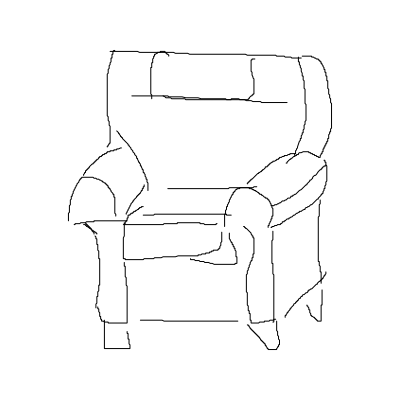} & \includegraphics[width=3cm, height=3cm]{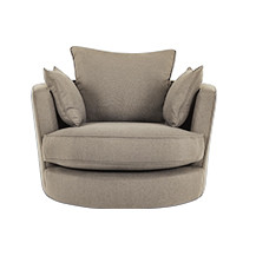} & \includegraphics[width=3cm, height=3cm]{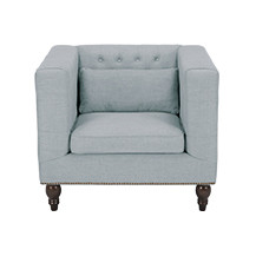} & \includegraphics[width=3cm, height=3cm]{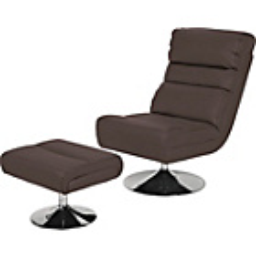} & \includegraphics[width=3cm, height=3cm, cfbox=ForestGreen 2pt 2pt]{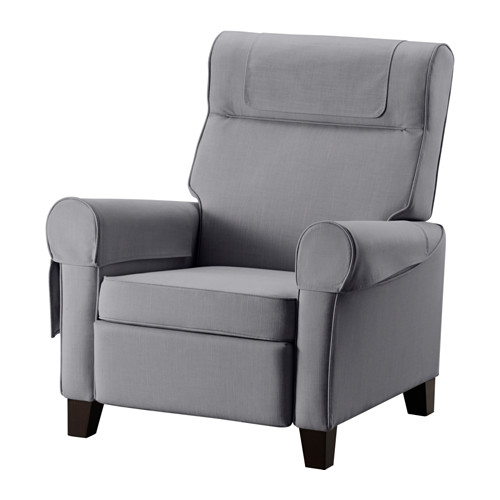} & \includegraphics[width=3cm, height=3cm]{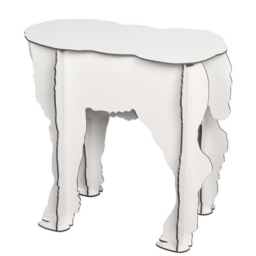} & \includegraphics[width=3cm, height=3cm]{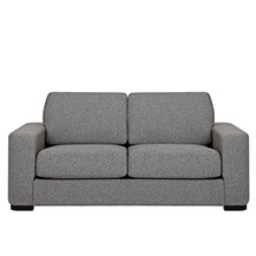} & \includegraphics[width=3cm, height=3cm]{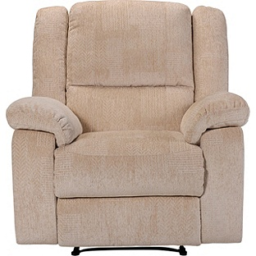} & \includegraphics[width=3cm, height=3cm]{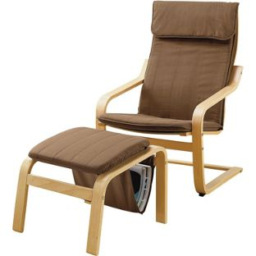} & \includegraphics[width=3cm, height=3cm]{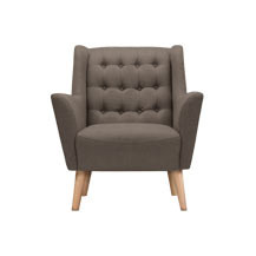} & \includegraphics[width=3cm, height=3cm]{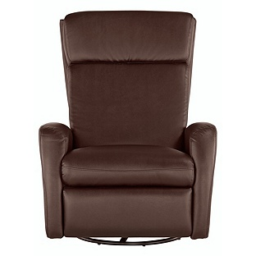} \\ 
\raisebox{1.5cm}{\huge 7 \hspace{6pt}} & \includegraphics[width=3cm, height=3cm]{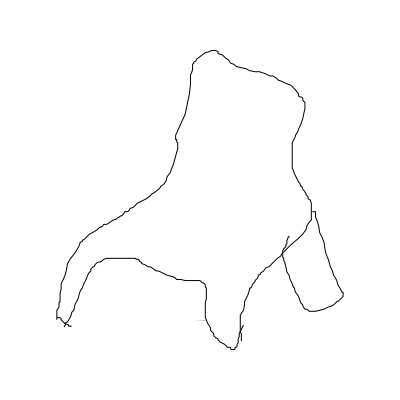} & \includegraphics[width=3cm, height=3cm]{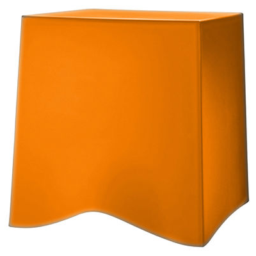} & \includegraphics[width=3cm, height=3cm, cfbox=ForestGreen 2pt 2pt]{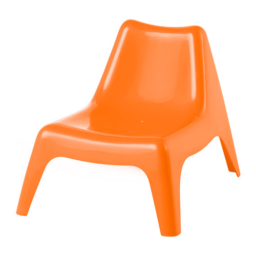} & \includegraphics[width=3cm, height=3cm]{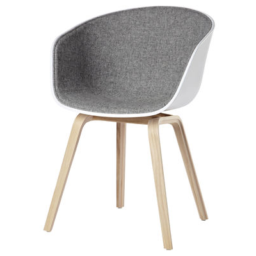} & \includegraphics[width=3cm, height=3cm]{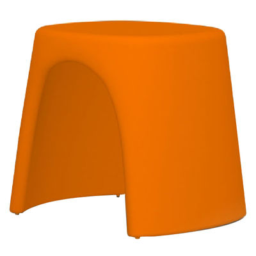} & \includegraphics[width=3cm, height=3cm]{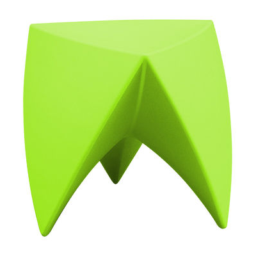} & \includegraphics[width=3cm, height=3cm]{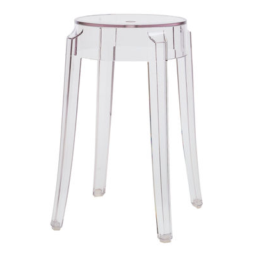} & \includegraphics[width=3cm, height=3cm]{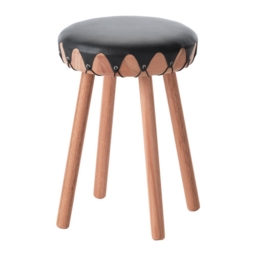} & \includegraphics[width=3cm, height=3cm]{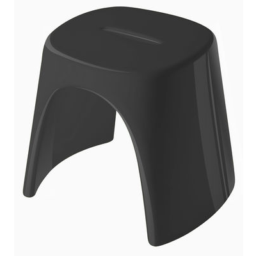} & \includegraphics[width=3cm, height=3cm]{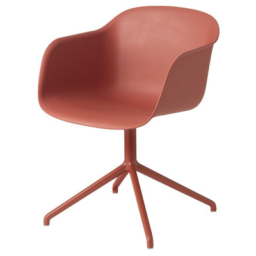} & \includegraphics[width=3cm, height=3cm]{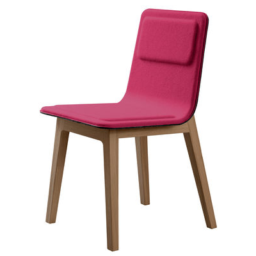} \\ 
\raisebox{1.5cm}{\huge 8 \hspace{6pt}} & \includegraphics[width=3cm, height=3cm]{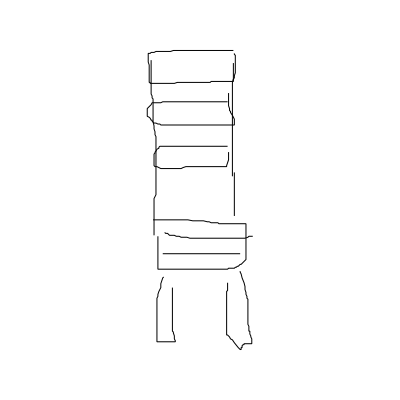} & \includegraphics[width=3cm, height=3cm, cfbox=ForestGreen 2pt 2pt]{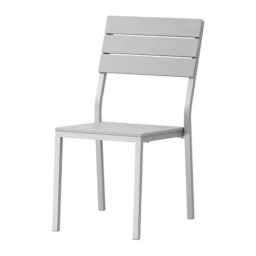} & \includegraphics[width=3cm, height=3cm]{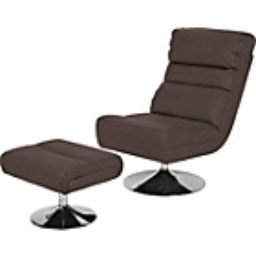} & \includegraphics[width=3cm, height=3cm]{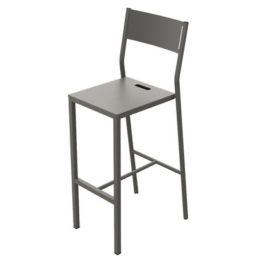} & \includegraphics[width=3cm, height=3cm]{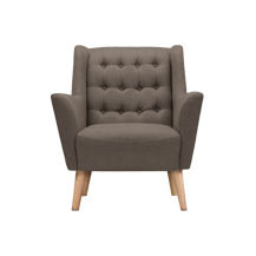} & \includegraphics[width=3cm, height=3cm]{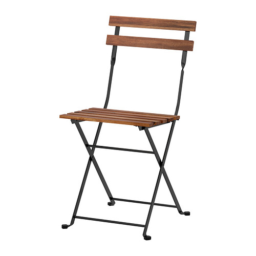} & \includegraphics[width=3cm, height=3cm]{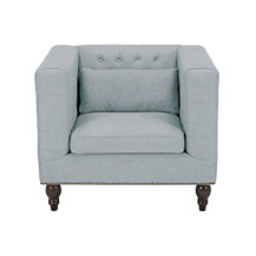} & \includegraphics[width=3cm, height=3cm]{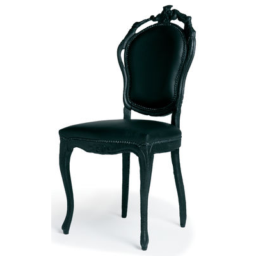} & \includegraphics[width=3cm, height=3cm]{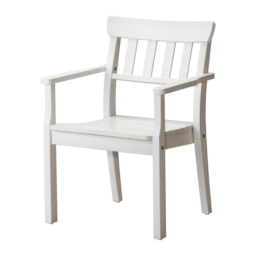} & \includegraphics[width=3cm, height=3cm]{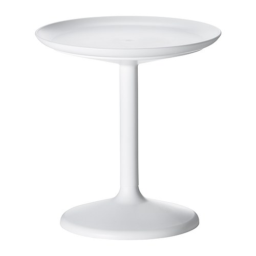} & \includegraphics[width=3cm, height=3cm]{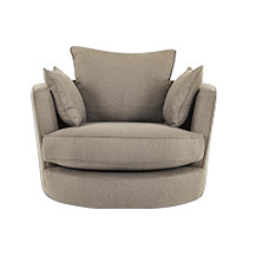} \\ 
\raisebox{1.5cm}{\huge 9 \hspace{6pt}} & \includegraphics[width=3cm, height=3cm]{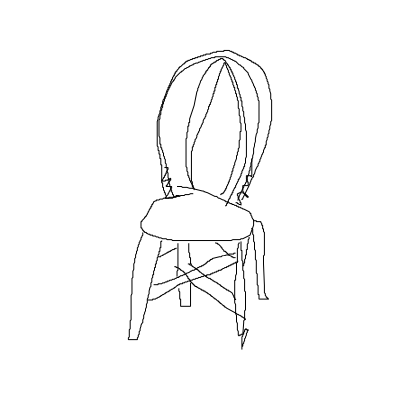} & \includegraphics[width=3cm, height=3cm, cfbox=ForestGreen 2pt 2pt]{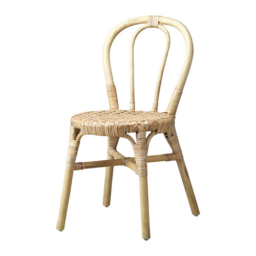} & \includegraphics[width=3cm, height=3cm]{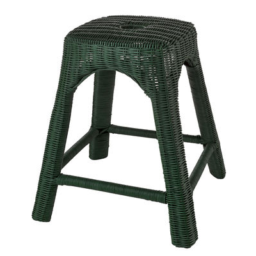} & \includegraphics[width=3cm, height=3cm]{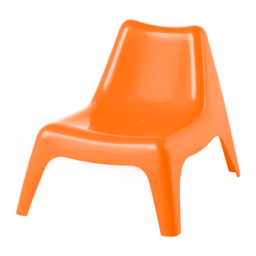} & \includegraphics[width=3cm, height=3cm]{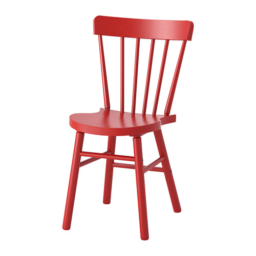} & \includegraphics[width=3cm, height=3cm]{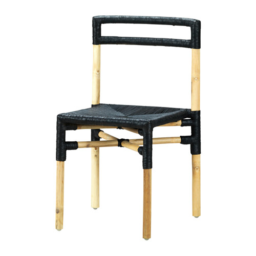} & \includegraphics[width=3cm, height=3cm]{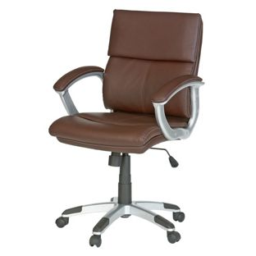} & \includegraphics[width=3cm, height=3cm]{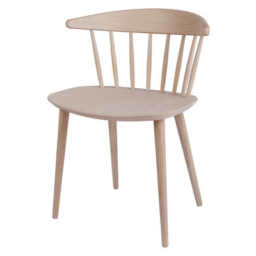} & \includegraphics[width=3cm, height=3cm]{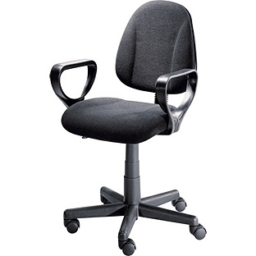} & \includegraphics[width=3cm, height=3cm]{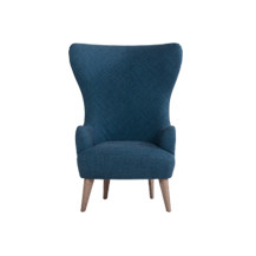} & \includegraphics[width=3cm, height=3cm]{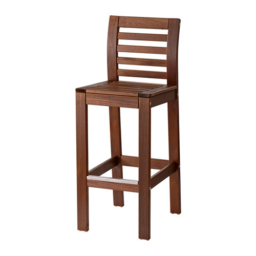} \\ 
\raisebox{1.5cm}{\huge 10 \hspace{6pt}} & \includegraphics[width=3cm, height=3cm]{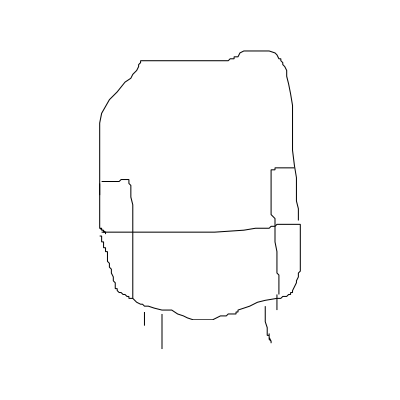} & \includegraphics[width=3cm, height=3cm]{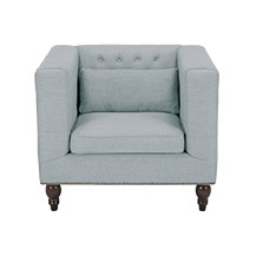} & \includegraphics[width=3cm, height=3cm, cfbox=ForestGreen 2pt 2pt]{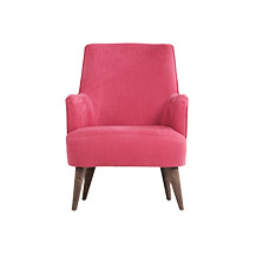} & \includegraphics[width=3cm, height=3cm]{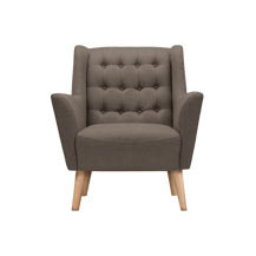} & \includegraphics[width=3cm, height=3cm]{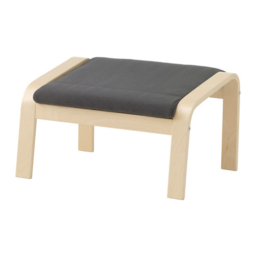} & \includegraphics[width=3cm, height=3cm]{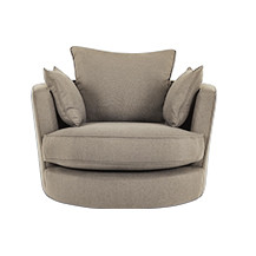} & \includegraphics[width=3cm, height=3cm]{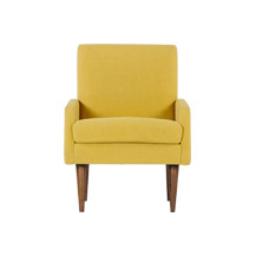} & \includegraphics[width=3cm, height=3cm]{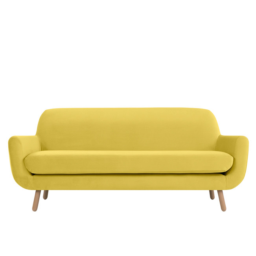} & \includegraphics[width=3cm, height=3cm]{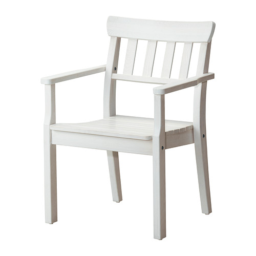} & \includegraphics[width=3cm, height=3cm]{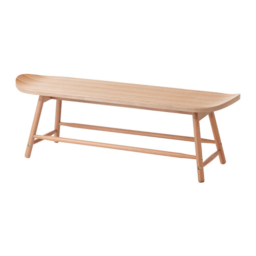} & \includegraphics[width=3cm, height=3cm]{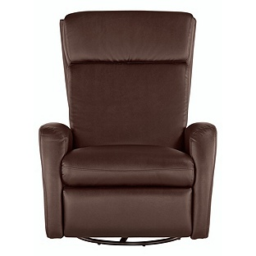} \\ 
\raisebox{1.5cm}{\huge 11 \hspace{6pt}} & \includegraphics[width=3cm, height=3cm]{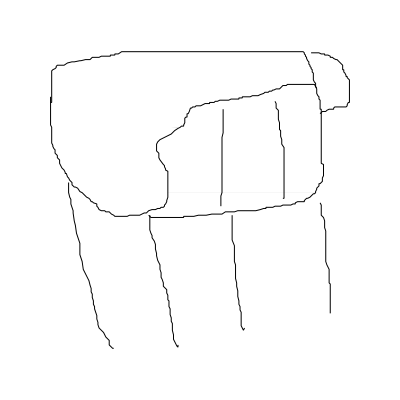} & \includegraphics[width=3cm, height=3cm, cfbox=ForestGreen 2pt 2pt]{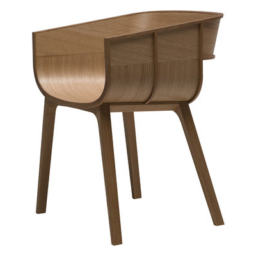} & \includegraphics[width=3cm, height=3cm]{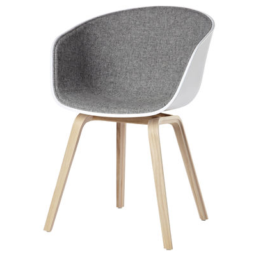} & \includegraphics[width=3cm, height=3cm]{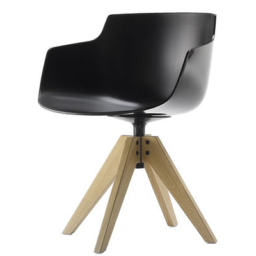} & \includegraphics[width=3cm, height=3cm]{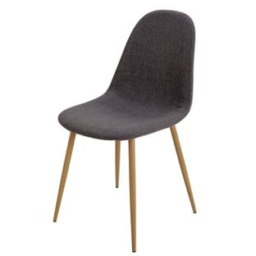} & \includegraphics[width=3cm, height=3cm]{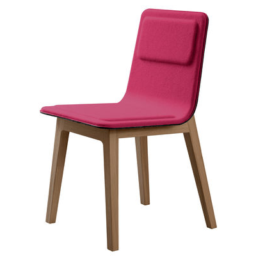} & \includegraphics[width=3cm, height=3cm]{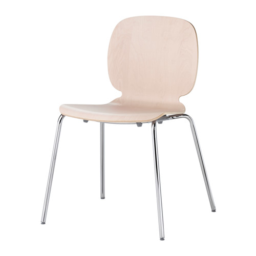} & \includegraphics[width=3cm, height=3cm]{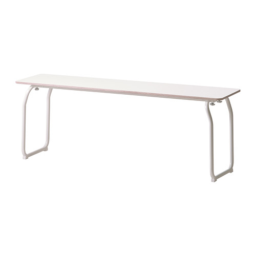} & \includegraphics[width=3cm, height=3cm]{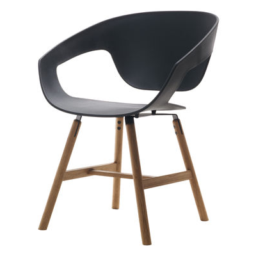} & \includegraphics[width=3cm, height=3cm]{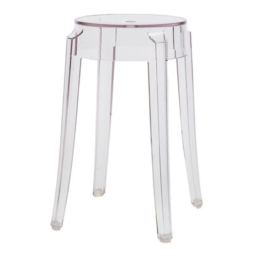} & \includegraphics[width=3cm, height=3cm]{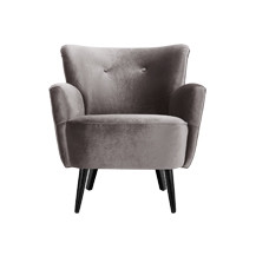} \\ 
\raisebox{1.5cm}{\huge 12 \hspace{6pt}} & \includegraphics[width=3cm, height=3cm]{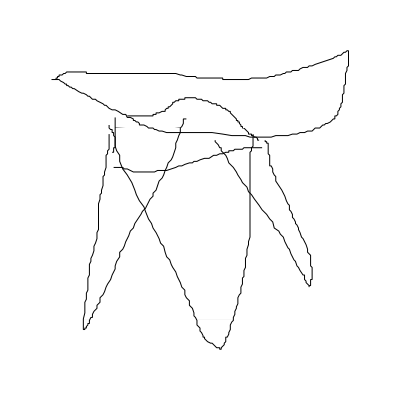} & \includegraphics[width=3cm, height=3cm]{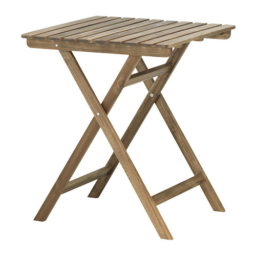} & \includegraphics[width=3cm, height=3cm]{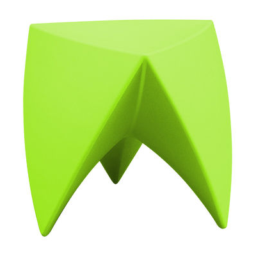} & \includegraphics[width=3cm, height=3cm, cfbox=ForestGreen 2pt 2pt]{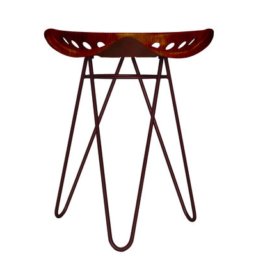} & \includegraphics[width=3cm, height=3cm]{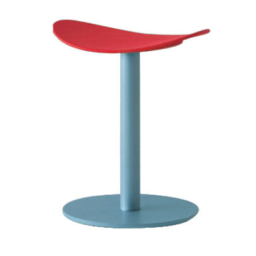} & \includegraphics[width=3cm, height=3cm]{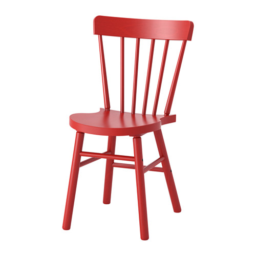} & \includegraphics[width=3cm, height=3cm]{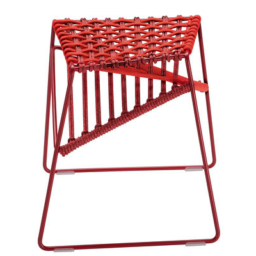} & \includegraphics[width=3cm, height=3cm]{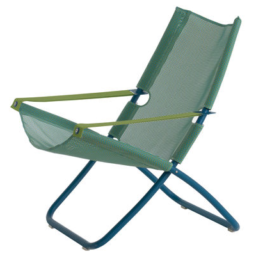} & \includegraphics[width=3cm, height=3cm]{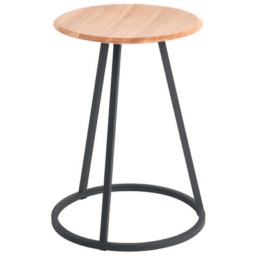} & \includegraphics[width=3cm, height=3cm]{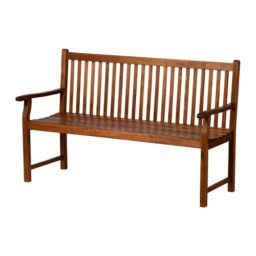} & \includegraphics[width=3cm, height=3cm]{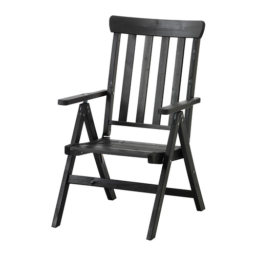}
\end{tabular}
}\end{center}
\caption{Qualitative fine-grained SBIR results on Chair-V2 dataset.}
\label{fig:chair_v2_long}
\label{fig:qual_results_retrieval_results_app/ChairV2/}
\end{figure*}

\subsection{Sketchy Dataset}
Apart from the instance-discriminative fine-grained features, the network learns attribute information such as classes with features that are closely related visually (rows 1, 3, 6, 11 of \cref{fig:sketchy_long}), object orientation and geometry (rows 8, 5, 4, 10 of \cref{fig:sketchy_long}), and even naturally occurring relationships (examples 5.1 and 5.2 of \cref{fig:sketchy_long}).

\subsection{QMUL-Shoe-V2 Dataset}
Most  instances  with  a  sufficient  number  of  fine-grained  instance-discriminative features can be seen to appear in the top 1.  However, for the ones that do get demoted in rank, (row 4 of \cref{fig:shoe_v2_long}), are preceded by an instance that have noticeable features in common that could cause confusion. Some false positive top-1 results might occur as a side-effect of modality fusion (row 10 of \cref{fig:shoe_v2_long}), where the embedding space also captures the texture information from the photo modality, which may not always be necessarily relevant (causing a shoe with a similar shiny texture as the ground-truth being returned as the first retrieval result in row 10 of \cref{fig:shoe_v2_long}).

\subsection{QMUL-Chair-V2 Dataset}
The modality fusion operator is able to capture cross-modal information such as color and texture, which are beyond the geometric structure of sketches (rows 4, 5, 7 of \cref{fig:chair_v2_long}), while being able to able to attend to the modality-native attributes, such as structural geometry and orientation (rows 3, 8, 9, 10, 11 of \cref{fig:chair_v2_long}).

\clearpage
\section{Attention Maps}
{
\setlength{\tabcolsep}{0pt}
\renewcommand{\arraystretch}{0.07}
\begin{figure*}[!h]
\centering
\resizebox{0.82\textwidth}{!}{
\begin{tabular}{@{}ccc}
\includegraphics[width=2cm,height=0.9cm]{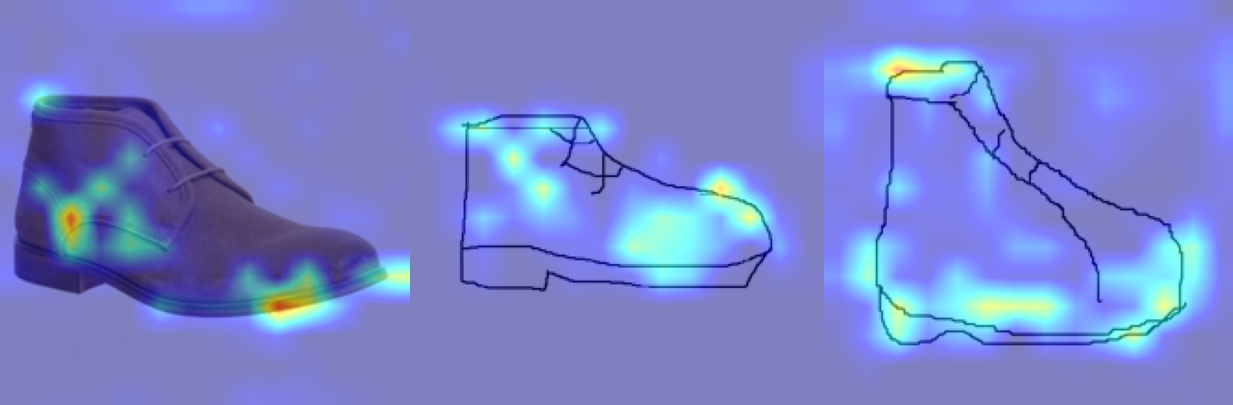}& 
\includegraphics[width=2cm,height=0.9cm]{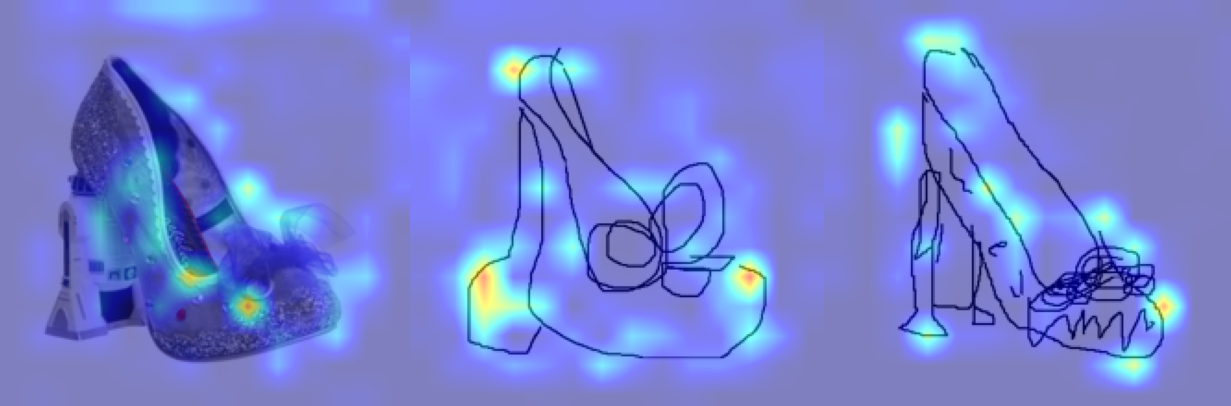}& 
\includegraphics[width=2cm,height=0.9cm]{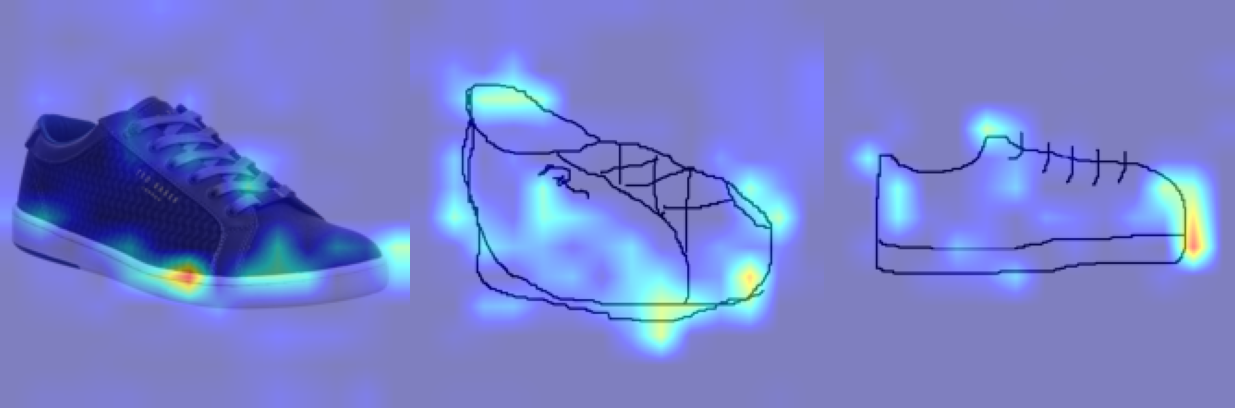} \\
\includegraphics[width=2cm,height=0.9cm]{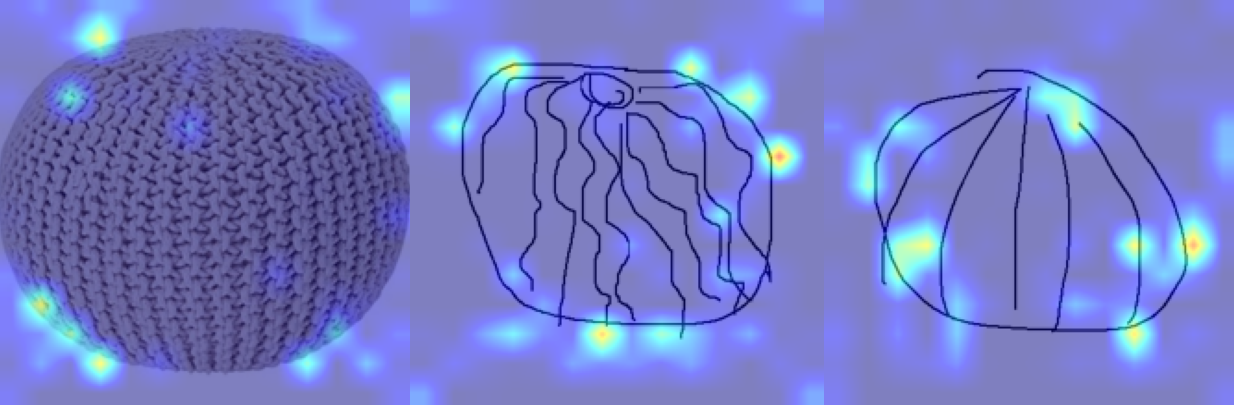} &
\includegraphics[width=2cm,height=0.9cm]{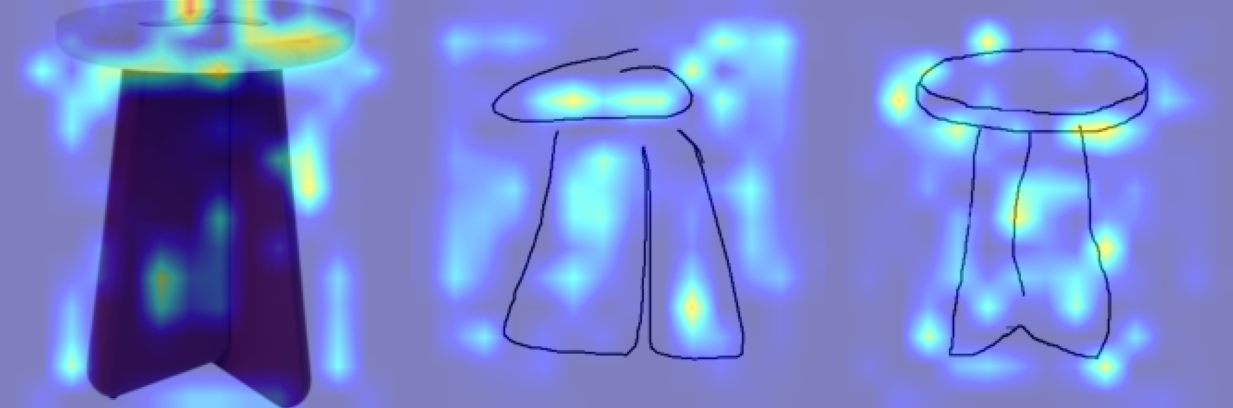}& 
\includegraphics[width=2cm,height=0.9cm]{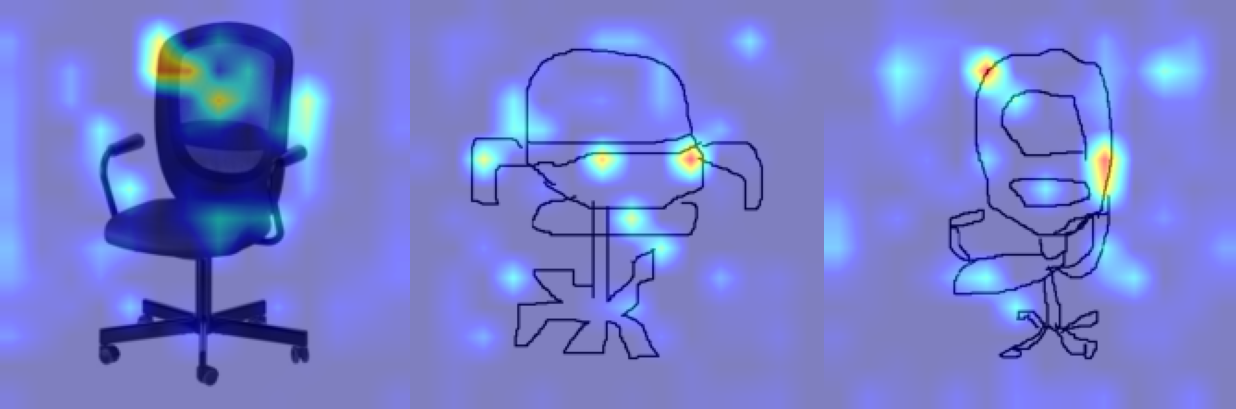} \\
\includegraphics[width=2cm,height=0.9cm]{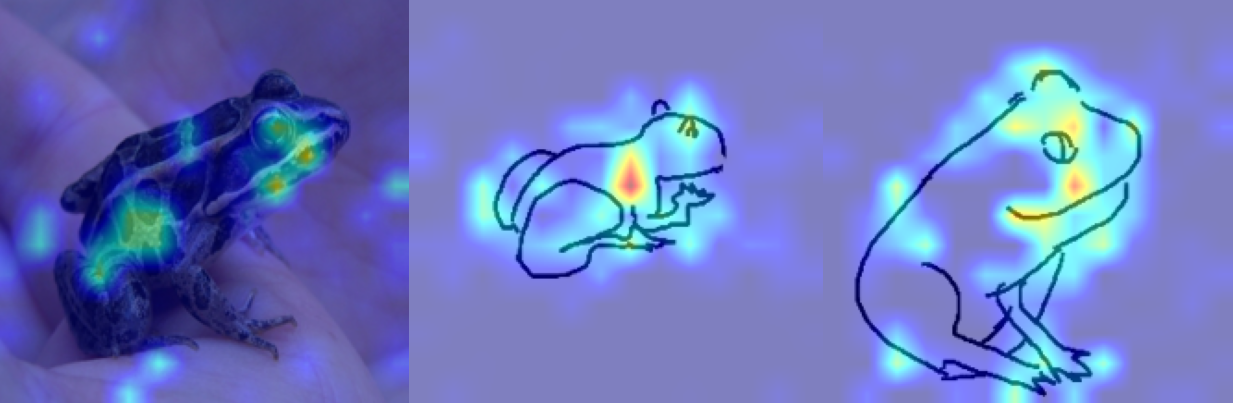} &
\includegraphics[width=2cm,height=0.9cm]{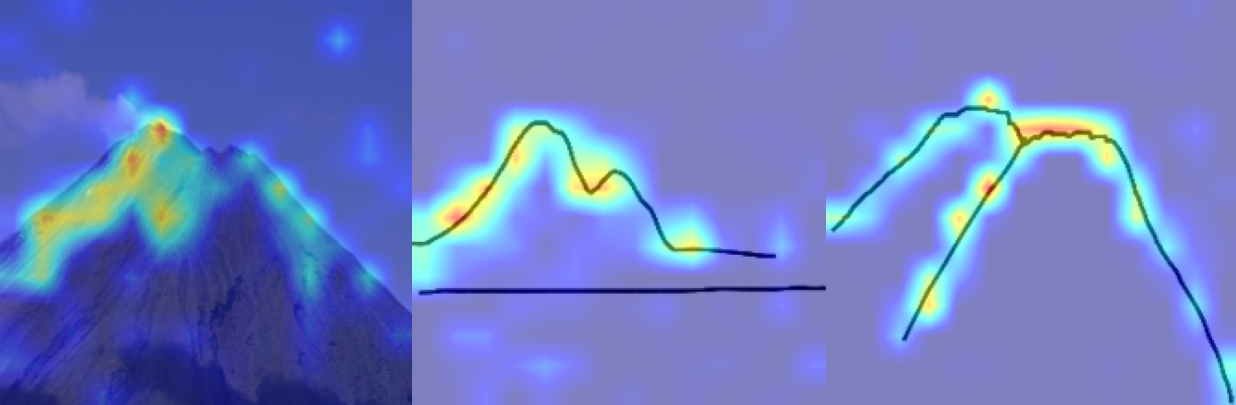}& 
\includegraphics[width=2cm,height=0.9cm]{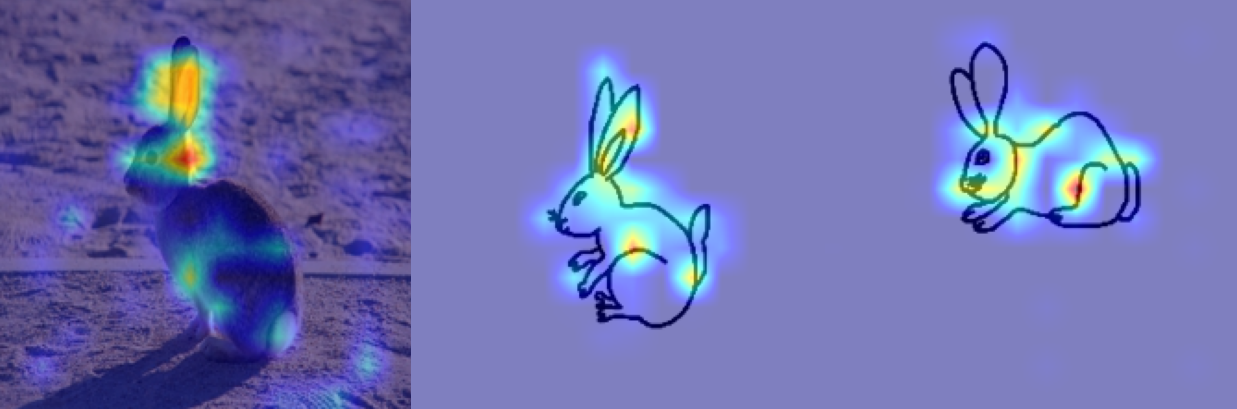}
\end{tabular}}
\caption{Attention maps on photos and corresponding sketches obtained from the student encoders (examples from the Shoe-V2, Chair-V2 and Sketchy datasets).}
\label{AttentionMaps}
\end{figure*}
}

\Cref{AttentionMaps} depicts attention maps for photos and two of their corresponding sketches obtained from the photo and the sketch students respectively.
Both the networks can be seen to generally focus on the same object localities irrespective of the modality. Also, within the sketch modality, the regions attended to by the sketch encoder are quite stable, indicating that the network has learned to focus more on the structural information and is robust to the variations in sketching style.


\bibliography{egbib}